\documentclass{article}

\usepackage{microtype}
\usepackage{graphicx}
\usepackage{booktabs} 

\usepackage{comment}
\usepackage{amsmath}
\usepackage{amssymb}
\usepackage{amsfonts}
\usepackage{dsfont}
\usepackage{subcaption}
\usepackage[ruled,lined]{algorithm2e}
\usepackage{multirow}
\usepackage{adjustbox}
\usepackage{amsthm}
\usepackage{multirow}
\usepackage{booktabs}
\usepackage{enumitem}

\usepackage{amsmath,amsfonts,bm}

\def\argmin{\mathop{\rm argmin}}%

\def\bx{\mathbf{x}} %
\def\tbx{\tilde{\mathbf{x}}} %

\def\by{\mathbf{y}} %
\def\tby{\tilde{\mathbf{y}}} %

\def\vw{\bm{w}}

\def\bv{\mathbf{v}} %
\def\tv{\tilde{v}} %
\def\tbv{\tilde{\mathbf{v}}} %

\usepackage{hyperref}

\newtheorem*{proof*}{Proof}

\SetAlgoCaptionSeparator{.}

\newcommand{\eg}{\emph{e.g.}} 
\newcommand{\ie}{\emph{i.e.}} 
\newcommand{\etc}{\emph{etc}}

\usepackage[accepted]{icml2020}

\icmltitlerunning{Submission and Formatting Instructions for ICML 2020}

\begin{document}

\twocolumn[
\icmltitlerunning{Beyond Synthetic Noise: Deep Learning on Controlled Noisy Labels}

\icmltitle{Beyond Synthetic Noise: Deep Learning on Controlled Noisy Labels}




\begin{icmlauthorlist}
\icmlauthor{Lu Jiang}{goo}
\icmlauthor{Di Huang}{cloud}
\icmlauthor{Mason Liu}{cor}
\icmlauthor{Weilong Yang}{goo}
\end{icmlauthorlist}

\icmlaffiliation{goo}{Google Research, Mountain View, United States}
\icmlaffiliation{cloud}{Google Cloud AI, Sunnyvale, United States}
\icmlaffiliation{cor}{Cornell University, Ithaca, United States}

\icmlcorrespondingauthor{Lu Jiang}{lujiang@google.com}

\icmlkeywords{Machine Learning, Robust Learning, Deep Learning, Noisy Data, ICML}
\vskip 0.3in
]



\printAffiliationsAndNotice{}  

\begin{abstract}
Performing controlled experiments on noisy data is essential in understanding deep learning across noise levels. Due to the lack of suitable datasets, previous research has only examined deep learning on controlled synthetic label noise, and real-world label noise has never been studied in a controlled setting. This paper makes three contributions. First, we establish the first benchmark of controlled real-world label noise from the web. This new benchmark enables us to study the web label noise in a controlled setting for the first time. The second contribution is a simple but effective method to overcome both synthetic and real noisy labels. We show that our method achieves the best result on our dataset as well as on two public benchmarks (CIFAR and WebVision). Third, we conduct the largest study by far into understanding deep neural networks trained on noisy labels across different noise levels, noise types, network architectures, and training settings. The data and code are released at the following link {\footnotesize \url{http://www.lujiang.info/cnlw.html}}.
\end{abstract}

\vspace{-3mm}

\section{Introduction}
Performing experiments on controlled noise is essential in understanding Deep Neural Networks (DNNs) trained on noisy labeled data. Previous work performs controlled experiments by injecting a series of synthetic label noises into a well-annotated dataset such that the dataset's noise level can vary, in a controlled manner, to reflect different magnitudes of label corruption in real applications. Through studying controlled synthetic label noise, researchers have discovered theories and methodologies that have greatly fostered the development of this field.

However, due to the lack of suitable datasets, previous work has only examined DNNs on controlled synthetic label noise, and real-world label noise has never been studied in a controlled setting. This leads to two major issues. First, as synthetic noise is generated from an artificial distribution, a tiny change in the distribution may lead to inconsistent or even contradictory findings. For example, contrary to the common understanding that DNNs trained on synthetic noisy labels generalize poorly~\citep{zhang2017understanding}, \citet{rolnick2017deep} showed that DNNs can be robust to massive label noise when the noise distribution is made slightly different. Due to the lack of datasets, these findings, unfortunately, have not yet been verified beyond synthetic noise in a controlled setting. Second, the vast majority of previous studies prefer to verify robust learning methods on a spectrum of noise levels because the goal of these methods is to overcome a wide range of noise levels. However, current evaluations are limited because they are conducted only on synthetic label noise.
Although there do exist datasets of real label noise \eg~WebVision~\citep{li2017webvision}, Clothing-1M~\citep{xiao2015learning}, \etc, they are not suitable for controlled evaluation in which a method must be systematically verified on multiple different noise levels, because the training images in these datasets are not manually labeled and hence their data noise level is fixed and unknown.

In this paper, we study a realistic type of label noise in a controlled setting called web labels. ``Webly-labeled'' images are commonly used in the literature~\cite{bootkrajang2012label,li2017webvision,krause2016unreasonable,chen2015webly}, in which both images and labels are crawled from the web and the noisy labels are automatically determined by matching the images' surrounding text to a class name during web crawling or equivalently by querying the search index afterward. Unlike synthetic labels, web labels follow a realistic label noise distribution but have not been studied in a controlled setting.

We make three contributions in this paper. First, we establish the first benchmark of controlled web label noise, where each training example is carefully annotated to indicate whether the label is correct or not. Specifically, we automatically collect images by querying Google Image Search using a set of class names, have each image annotated by 3-5 workers, and create training sets of ten controlled noise levels. As the primary goal of our annotation is to identify images with incorrect labels, to obtain a sufficient number of these images we have to collect a total of about 800,000 annotations over 212,588 images. The new benchmark enables us to go beyond synthetic label noise and study web label noise in a controlled setting. For convenience, we will refer it as web label noise (or \textbf{\emph{red noise}}) to distinguish it from synthetic label noise (or \textbf{\emph{blue noise}})\footnote{From the red and blue pill in the movie ``The Matrix‘’ (1999), where the red pill is used to refer to the truth about reality.}.

Second, this paper introduces a simple yet highly effective method to overcome both synthetic and real-world noisy labels. It is based on a new idea of minimizing the empirical vicinal risk using curriculum learning. We show that it consistently outperforms baseline methods on our datasets and achieves state-of-the-art performance on two public benchmarks of synthetic and real-world noisy labels. Notably, on the challenging benchmark WebVision 1.0~\citep{li2017webvision} that consists of 2.2 million images of real-world noisy labels, it yields a significant improvement of $3\%$ in the top-1 accuracy, achieving the best-published result under the standard training setting.

Finally, we conduct the \emph{largest study} by far into understanding DNNs trained on noisy labels across a variety of noise types (blue and red), noise levels, training settings, and network architectures. Our study confirms the existing findings of \citet{zhang2017understanding} and \citet{arpit2017closer} on synthetic labels, and brings forward new findings that may challenge our preconceptions about DNNs trained on noisy labels. See the findings in Section~\ref{sec:exp_understanding}. It is worth noting that these findings along with benchmark results are a result of conducting thousands of experiments using tremendous computation power (hundreds of thousands of V100 GPU hours). We hope our (i) benchmark, (ii) new method, and (iii) findings will facilitate future deep learning research on noisy labeled data. We will release our data and code.

\section{Related Work}

\subsection{Datasets of noisy training labels}
While many types of noises exist \eg~image corruption noise~\citep{hendrycks2019benchmarking}, image registration noise~\citep{mnih2012learning}, or noise from adversarial attacks~\citep{zhang2019theoretically}, this paper focuses on label noise, and in particular web label noise -- a common type of label noise used in the literature. To the best of our knowledge, there have been no datasets of controlled web label noise. The closest to ours is the datasets of two types of noises: controlled synthetic label noise and uncontrolled web label noise.

In the dataset of controlled synthetic label noise, a series of synthetic label noises are injected into a well-annotated dataset in a controlled manner to reflect different magnitudes of label corruption in real applications. The most common one is the symmetric label noise, in which the label of each example is independently and uniformly changed to a random class with a controlled probability. Many works studied the symmetric label in controlled settings and presented their findings, including famous ones, like~\citep{zhang2017understanding} and~\citep{arpit2017closer}. The symmetric label is also commonly used as a benchmark to evaluate robust learning methods in a noise-control setting \eg~in \citep{vahdat2017toward,shu2019meta,jiang2018mentornet,ma2018dimensionality,han2018co,van2015learning,li2019learning,arazo2019unsupervised,charoenphakdee2019symmetric}. Other types of synthetic label noise were also proposed, including class-conditional noises~\citep{patrini2017making,rolnick2017deep,reeve2019fast,han2018masking}, noises from other datasets~\citep{wang2018iterative,seo2019combinatorial}, \etc. However, these noises are still synthetically generated from artificial distributions. Moreover, different works might use different parameters to generate such synthetic noises, which may make their results incomparable. See the example of~\citep{rolnick2017deep} in the introduction. 

In the dataset of uncontrolled web label noise, both images and labels are crawled from the web and the noisy labels are automatically determined by matching the images' surrounding text to a class name. This can be achieved by querying a search index~\citep{krause2016unreasonable,li2017webvision,mahajan2018exploring}. For example, In WebVision, \citet{li2017webvision} collected web images with noisy labels by querying Google and Flickr image search using the 1,000 class names from ImageNet. \citet{mahajan2018exploring} gathered a large scale set of images with noisy labels by searching hashtags on Instagram. However, these datasets do not provide ground-truth labels for training examples. Their noise level is hence fixed and unknown. As a result, they are not suitable for controlled studies in which different noise levels must be systematically examined. Besides, to get controlled web noise, it may not be a feasible option to annotate images in these datasets due to their imbalanced class distribution.

\subsection{Robust deep learning methods}
Robust learning is experiencing a renaissance in the deep learning era. Nowadays training datasets usually contain noisy examples. The ability of DNNs to memorize all noisy training labels often leads to poor generalization on the clean test data. Recent contributions based on deep learning handle noisy data in multiple directions including dropout~\citep{arpit2017closer} and other regularization techniques~\citep{azadi2016auxiliary,noh2017regularizing}, label cleaning/correction~\citep{reed2014training,goldberger2017training,Li2017ICCV,veit2017learning,song2019selfie}, example weighting~\citep{jiang2018mentornet,ren2018learning,shu2019meta,jiang2015self,liang2016learning}, 
cross-validation~\citep{northcutt2019confident}, semi-supervised learning~\citep{hendrycks2018using,vahdat2017toward,li2020dividemix,zhang2020distilling}, data augmentation~\citep{zhang2018mixup,cheng2019robust,cheng2020advaug,liang2020simaug}, among others. Different from prior work, we introduce a simple yet effective method to overcome both synthetic and real-world noisy labels. Compared with semi-supervised learning methods, our method learns DNNs without using any clean label.

\section{Dataset}\label{sec:dataset}

\begin{table*}[ht]
\centering
\caption{\label{tab:dataset}Overview of our datasets of controlled red (web) label noise. Blue (synthetic) label noise is also included for comparison.}
\begin{tabular}{lccccc}
\toprule
{\bf Dataset} & \multicolumn{1}{l}{{\bf \#Class}} & {\bf Noise Source} & {\bf Train Size} & {\bf Val Size} & {\bf Controlled Noise Levels (\%)} \\
\midrule
Red Mini-ImageNet & \multirow{2}{*}{100} & image search label & 50,000 & \multirow{2}{*}{5,000} & 0, 5, 10, 15, 20, 30, 40, 50, 60, 80 \\
Blue Mini-ImageNet &  & symmetric label flipping & 60,000 &  & 0, 5, 10, 15, 20, 30, 40, 50, 60, 80 \\ \hline
Red Stanford Cars & \multirow{2}{*}{196} & image search label & 8,144 & \multirow{2}{*}{8,041} & 0, 5, 10, 15, 20, 30, 40, 50, 60, 80 \\
Blue Stanford Cars &  & symmetric label flipping & 8,144 &  & 0, 5, 10, 15, 20, 30, 40, 50, 60, 80 \\
\bottomrule
\end{tabular}
\vspace{-5mm}
\end{table*}

Our goal is to create a benchmark of controlled noise that resembles a realistic label noise distribution. Unlike existing datasets such as WebVision or Clothing1M, our benchmark provides \emph{controlled label noise} where every single training example is carefully annotated by several human annotators.

Our benchmark is built on top of two public datasets: Mini-ImageNet~\citep{vinyals2016matching} for coarse-grained image classification and Stanford Cars~\citep{krause2013collecting} for fine-grained image classification.
Mini-ImageNet has images of size 84x84 with 100 classes from ImageNet~\citep{deng2009imagenet}. We use all 60K images for training and the 5K images in the ILSVRC12 validation set for testing. Stanford Cars contain 16,185 high-resolution images of 196 classes of cars (Make, Model, Year) split 50-50 into training and validation set. The standard train/validation split is used.

\subsection{Dataset Construction}

We build our datasets to replace synthetic noisy labels with web noisy labels in a controlled manner. To recap, let us revisit the construction of existing datasets of noisy labels. Synthetic noisy datasets are generated beginning with a well-labeled dataset. The most common type of synthetic label noise is called \emph{symmetric label} noise in which the label of each training example is independently changed to a random incorrect class with a probability $p$ called noise level.\footnote{This is slightly different from \citep{zhang2017understanding} and is the same as~\citep{jiang2018mentornet}. We do not allow examples to be label flipped to their true labels. It makes $p$ denote the exact noise level and independent of the total number of classes.}.
The noise level indicates the percentage of training examples with incorrect labels. As the true labels for all images are known, one can enumerate $p$ to obtain training sets of different noise levels and use them in noise-control studies. On class-balanced datasets, this process is equivalent to sampling $p$\% training images from each class and then replacing their labels with the labels uniformly drawn from other classes. The drawback is that the synthetic labels are artificially created and do not follow the distribution of real-world label noise.

On the other hand, there exist a few datasets of uncontrolled web label noise~\citep{xiao2015learning,li2017webvision,krause2016unreasonable}. In these datasets, the images are crawled from the web and their labels are automatically assigned by matching the images' surrounding text to a class name. This can be achieved by querying a search index.
These datasets contain noisy web labels.
However, as their training images are not manually labeled, their label noise level is fixed and unknown, rendering existing datasets unsuitable for controlled studies.

For our benchmark, we follow the construction of synthetic datasets with one important difference -- instead of changing the labels of the sampled clean images, we replace the clean images with incorrectly labeled web images while leaving the label unchanged. The advantage of this approach is that we closely match the construction of synthetic datasets while still being able to introduce controlled web label noise.

\subsection{Noisy Web Label Acquisition}
We collect images with incorrect web labels in three steps: (1) images collection, (2) deduplication, and (3) manual annotation. In the first step, we combine images independently retrieved by Google image search from two sources: text-to-image and image-to-image search. For the text-to-image search, we formulate a text query for each class using its class name and broader category to retrieve the top images. For image-to-image search, we query the search engine using every training image in Mini-ImageNet and Stanford Cars. 
Finally, we union the two search results, where the text-to-image search results account for 82\% of our final benchmark. Note that the rationale for including a small amount of image-to-image results is to enrich the types of web label noises in our benchmark. We show that an alternative way to construct the dataset by removing all image-to-image results leads to consistent results in the Appendix C.

In the second step of deduplication, following~\citep{kornblith2019better}, we run a CNN-based duplicate detector over all images to remove near-duplicates to any of the images in the validation set. All images are retrieved under the usage rights ``free to use or share''~\footnote{\scriptsize \url{https://support.google.com/websearch/answer/29508}}. But we still recommend checking their actual usage right for the image.

Finally, the images are annotated using the Google Cloud labeling service. The annotators are asked to provide a binary question: ``is the label correct for this image?''. Every image is independently annotated by 3-5 workers and the final label is reached by majority voting. Statistics show annotators disagree only on a small proportion (11\%) of the total images. The remainder have unanimous labels agreed by at least 3 annotators.

\subsection{Dataset Overview}
In total, we collect about 800,000 annotations over 212,588 images, out of which there are 54,400 images with incorrect labels on Mini-ImageNet and 12,629 images on Stanford Cars. The remainder of the images have correct labels.
Using web images with incorrect labels, we replace $p$\% of the original training images in the two datasets, and enumerate $p$ in 10 different levels to create the controlled web label noise: $\{0\%, 5\%, 10\%, 15\%, 20\%, 30\%, 40\%, 50\%, 60\%, 80\%\}$. Similar to synthetic noise, $p$ is made uniform across classes, \eg~$p=20\%$ means that every class has roughly 20\% incorrect labels. In Appendix C, we show constructing the dataset only using web images leading to consistent results.

Table~\ref{tab:dataset} summarizes our benchmark. For comparison, we also include synthetic (symmetric) labels of the same 10 noise levels. We use \textbf{\emph{blue noise}} to denote the synthetic label noise and \textbf{\emph{red noise}} for the web label noise. The sizes of the red and blue training sets are made similar to better compare their difference\footnote{As existing findings on synthetic noise hold on both the full (60K) and subset (50K) of Blue Mini-ImageNet, we choose to report the results on the 60K full set. This results in a slightly larger Blue Mini-ImageNet but may not affect our main contributions}. Only a subset of web images with incorrect labels is used in our dataset. But we release all 212,588 images along with their annotations, which can be downloaded at {\footnotesize \url{http://www.lujiang.info/cnlw.html}} under the license of Creative Commons.

For lack of space, we use Fig. 1 in the Appendix to illustrate noisy images in each dataset. In summary, there are three clear distinctions between images with the synthetic and web label noise. Images with label noise from the web (or red noise) (1) a higher degree of similarity to the true positive images, (2) exist at the instance-level, and (3) come from an open vocabulary outside the class vocabulary of Mini-ImageNet or Stanford Cars.

\section{Method}\label{sec:methods}
In this section, we introduce a simple method called \emph{MentorMix} to overcome both synthetic and web label noise. As its name suggests, our idea is inspired by MentorNet~\cite{jiang2018mentornet} and Mixup~\cite{zhang2018mixup}. The main idea is to design a new robust loss to overcome noisy labels using curriculum learning and vicinal risk minimization.

\subsection{Background on MentorNet and Mixup}

Consider a classification problem with training set $\mathcal{D} = \{\bx_1,y_1),\cdots, (\bx_n,y_n)\}$, where $\bx_i$ denotes the $i^{th}$ training image and $y_i \in [1,m]$ is an integer-valued noisy label over $m$ possible classes and $\by_i$ is the corresponding one-hot label. Note that no clean labels are allowed to be used in training.
Let $g_{s}(\bx_i;\vw)$ denote the prediction of a DNN, parameterized by $\vw \in \mathbb{R}^d$. MentorNet~\citep{jiang2018mentornet} minimizes the following objective:
\begin{equation}
\label{eq:mentornet_obj}
\begin{aligned}
\vw^* = \argmin_{\vw \in \mathbb{R}^d, \bv \in [0,1]^n} \mathbb{F}(\bv, \vw) \\
= \frac{1}{n}\sum_{i=1}^n  v_i \ell(g_s(\bx_i;\!\vw), y_i) + \theta \|\vw\|_2^2 + G(\bv; \gamma)
\end{aligned}
\end{equation}
where $\ell(g_s(\bx_i;\!\vw), y_i)$, or $\ell(\bx_i, y_i)$ for short, is the cross-entropy loss with the softmax function. $\theta$ is the weight decay parameter on the $l_2$ norm of the model parameters. For convenience, we make the weight decay regularization, data augmentation and dropout all subsumed inside $g_{s}$.

Eq.~\eqref{eq:mentornet_obj} introduces the latent weight variable $\bv \in [0,1]^n$ for every training example. The regularization term $G$ determines a curriculum~\cite{jiang2015self,jiang2014self,fan2017self} or equivalently a weighting scheme to compute the latent weight $v_i$ to each example. In~\citep{jiang2018mentornet}, the weighting scheme is computed by a neural network called MentorNet. In training, $\vw$ and $\bv$ are alternatively minimized inside a mini-batch, one at a time while the other is held fixed. Only $\vw$ is used at test time.

Mixup~\citep{zhang2018mixup} minimizes the empirical vicinal risk calculated from:
\begin{equation}
\label{eq:mixup_obj}
\vw^* = \argmin_{\vw} \frac{1}{n}\sum_{i=1}^n \frac{1}{n}\sum_{j=1}^n \mathop{\mathbb{E}}\limits_{\lambda} \lbrack \ell(g_s(\tbx_{ij};\!\vw), \tby_{ij})\rbrack
\end{equation}
$\tbx$ and $\tby$ are computed by the mixup function:
\begin{align}
\tbx_{ij} = \lambda \bx_i + (1-\lambda) \bx_j \\
\tby_{ij} = \lambda \by_i + (1-\lambda) \by_j
\end{align}
where $\lambda$ is drawn from the Beta distribution $\text{Beta}(\alpha, \alpha)$ controlled by hyperparameter $\alpha$. In practice, only the examples from the same mini-batch are mixed up during training.

\subsection{MentorMix}
In the proposed MentorMix, we minimize the empirical vicinal risk using curriculum learning. For simplicity, the self-paced regularizer $G(\bv)=-\gamma \|\bv\|_1$~\cite{kumar2010self,jiang2015self} is used and we have:
\begin{equation}
\label{eq:mentormix_obj}
\mathbb{F}(\tbv, \vw) = \frac{1}{n^2}\sum_{i=1}^n \sum_{j=1}^n \mathop{\mathbb{E}}\limits_{\lambda} \lbrack \tv_{ij} \ell(\tbx_{ij}, \tby_{ij}) - \gamma \tv_{ij} \rbrack,
\end{equation}
where $\gamma$ is a hyperparameter. It is easy to derive the optimal weighting scheme when the network parameter $\vw$ is fixed:
\begin{equation}
\label{eq:vstarij}
\tv_{ij}^* = \argmin_{\tbv \in [0,1]^{n \times n}} \mathbb{F}_{\vw}(\tbv) = \bm{1}(\ell(\tbx_{ij}, \tby_{ij}) \le \gamma),
\end{equation}
where $\bm{1}$ is the indicator function and $\mathbb{F}_{\vw}$ denotes the objective when $\vw$ is fixed.

Although Eq.~\eqref{eq:vstarij} gives a closed-form solution for computing the optimal weight, it is intractable to compute as this requires enumerating all pairs of training examples. 
We therefore resort to importance sampling to find the ``important'' examples. To do so, we define a stratum for each $\bx_i$ and draw an example from the following distribution:
\begin{equation}
P_{\bv}(v_i=1 | \bx_i, \by_i) = \frac{\exp(v_i^*/t)}{\sum_{j=1}^n \exp(v_j^*/t)},
\end{equation}
where $t$ is the temperature in the softmax function and is fixed to 1 in our experiments. $P_{\bv}$ specifies a density function over individual training examples. In theory, the distribution is defined over all training examples but, in practice to enable mini-batch training, we compute the distribution within each mini-batch (See Algorithm 1). $v_i^*$ is the optimal weight for $\bx_i$. $v_i^*$ is calculated from $v_i^* = \argmin_{v_i} \mathbb{F}_{\vw}(\bv)$ in Eq.~\eqref{eq:mentornet_obj} and can be conveniently obtained by MentorNet. As the optimal $v_{ij}^*$ can only have binary values according to Eq.~\eqref{eq:vstarij}, under the importance sampling we rewrite part of the objective in Eq.~\eqref{eq:mentormix_obj} as:
\begin{equation}
\begin{aligned}
\sum_{(\bx_i, \by_i) \sim \mathcal{D}}  \sum_{(\bx_j, \by_j) \sim \mathcal{D}} \mathop{\mathbb{E}}\limits_{\lambda} \lbrack v_{ij} \ell(\tbx_{ij}, \tby_{ij}) - \gamma v_{ij} \rbrack  \\
= \sum_{(\bx_i, \by_i) \sim \mathcal{D}}  \sum_{(\bx_j, \by_j) \sim P_{\bv}} \mathop{\mathbb{E}}\limits_{\lambda} \lbrack \ell(\tbx_{ij}, \tby_{ij}) \rbrack - \gamma
\end{aligned}
\end{equation}
where the constant $\gamma$ will be dropped during training. According to Eq.~\eqref{eq:vstarij}, our goal is to find the mixed-up examples of smaller loss. For a given example ($\bx_i$, $\by_i$), the loss of the mixed-up example $\ell(\tbx_{ij}, \tby_{ij})$ tends to be smaller when $\ell(\bx_j, \by_j)$ is small. Inspired by this idea, we sample $\bx_j$ from $P_{\bv}$ with respect to the weight $v_j^*$ that is monotonically decreasing with its loss $\ell(\bx_j, y_j)$ . In this way examples of lower loss are more likely to be selected in the mixup.

{
\begin{algorithm}
\small
\SetKwData{Left}{left}\SetKwData{This}{this}\SetKwData{Up}{up}
\SetKwInOut{Input}{Input}\SetKwInOut{Output}{Output}
\LinesNumbered
\Input{mini-batch $\mathcal{D}_{m}$; two hyperparameters $\gamma_p$ and $\alpha$}
\Output{the loss of the mini-batch}

For every $(\bx_i, y_i)$ in $\mathcal{D}_{m}$ compute $\ell(\bx_i, y_i)$ \;

Set $\ell_p(\mathcal{D}_{m})$ to be the $\gamma_p$-th percentile of the loss $\{\ell(\bx_i, y_i)\}$. \;

$\gamma \leftarrow \text{EMA}(\ell_p(\mathcal{D}_{m}))$ \; {\footnotesize\tcp{update the moving average}}

$v_i^* \leftarrow \text{MentorNet}(\ell(\bx_i, y_i), \gamma)$  \; {\footnotesize\tcp{MentorNet weight}}

Compute $P_{\bv} = \text{softmax}(\bv^*)$, where  $\bv^* = [v_1^*,\cdots, v_{|\mathcal{D}_{m}|}^*]$\;

Stop gradient \;

\ForEach{ $(\bx_i, \by_i)$} {
    Draw a sample $(\bx_j, \by_j)$ with replacement from $P_{\bv}$  \;
    
    $\lambda \leftarrow Beta (\alpha, \alpha)$ \;
    
    $\lambda \leftarrow v_i^* \max(\lambda, 1-\lambda) + (1-v_i^*) \min(\lambda, 1-\lambda)$  \;
    
    
    $\tbx_{ij} \leftarrow \lambda \bx_i + (1-\lambda) \bx_j$ \;
    
    $\tby_{ij} \leftarrow \lambda \by_i + (1-\lambda) \by_j$ \;
    
    Compute $\ell_i = \ell(\tbx_{ij}, \tby_{ij})$ \;
    
    Weight $\ell_i$ using a separate MentorNet \; {\footnotesize\tcp{optional}}
}
\Return $(1/|\mathcal{D}_{m}|) \sum_{i=1}^{|\mathcal{D}_{m}|} \ell_i$
\caption{{The proposed MentorMix method.}}
\label{alg:overall}
\end{algorithm}
\vspace{-1mm}
}

Algorithm 1 shows the four key steps to compute the loss for a mini-batch: weight (Step 2-4), sample (Step 5 and 8), mixup (Step 9-12), and weight again (Step 14), where the weighting is achieved by the MentorNet. Following our previous work~\citep{jiang2018mentornet}, we avoid directly setting $\gamma$ and adopt a moving average (in Step 3) to track the exponential moving average of the $\gamma_p$-th percentile of the mini-batch loss, in which $\gamma_p$ becomes the new hyperparameter. Step 4 computes the weight for individual example using a fixed MentorNet. The simplified MentorNet is used in our paper which is equivalent to computing the weight by a thresholding function $v_i^* = \bm{1}(\ell(\bx_{i}, y_i) \le \gamma)$. In Step 10, we assign a bigger (binary) weight between $\lambda$ and $1-\lambda$ to $\bx_i$ unless its $v^*_i$ is small. 
This trick is to stabilize importance sampling by encouraging each stratum to receive a bigger weight in the mixup. It leads to marginal performance gains but makes the training more robust to the choice of hyperparameters. In Step 14, the second weighting can be applied optionally using the weights produced by a separate MentorNet. This step is optional for low noise levels but is useful for high noise levels. Our algorithm has the same time and space complexity as that of the Mixup algorithm.

\section{Experiments}\label{sec:experiments}
This section verifies the proposed method on four datasets and presents new findings on web label noise. Specifically, in Section~\ref{sec:exp_method} we first verify the proposed method on our dataset and then compare it with the state-of-the-art on two public benchmarks of synthetic and real-world noisy labels. In Section~\ref{sec:exp_understanding}, we empirically examine DNNs trained on controlled noisy labels under various settings and present our findings that challenge our previous understandings.

\begin{table*}[ht]
\centering
\caption{Peak accuracy (\%) of the best trial of each method averaged across 10 noise levels. -- denotes the method failed to train.}
\label{tab:baseline}
\begin{tabular}{|c|cc|cc|cc|cc|}
\toprule
\multicolumn{1}{|c|}{\multirow{4}{*}{Method}} & \multicolumn{4}{c|}{Mini-ImageNet} & \multicolumn{4}{c|}{Stanford Cars} \\
\multicolumn{1}{|c|}{} & \multicolumn{2}{c|}{Fine-tuned} & \multicolumn{2}{c|}{Trained from scratch} & \multicolumn{2}{c|}{Fine-tuned} & \multicolumn{2}{c|}{Trained from scratch} \\
\multicolumn{1}{|c|}{} & Blue & Red & Blue & Red & Blue & Red & Blue & \multicolumn{1}{c|}{Red} \\
\hline
Vanilla&	82.3$\pm$1.9&	81.6$\pm$1.9&	58.3$\pm$10.3&	64.9$\pm$5.2&	70.0$\pm$16.8&	82.4$\pm$6.9&	53.8$\pm$24.4&	77.7$\pm$10.4\\
WeightDecay&	81.9$\pm$1.8&	81.5$\pm$1.8&	---& ---&	72.2$\pm$17.5&	84.3$\pm$6.6&	---&	---\\
Dropout&	82.8$\pm$1.3&	81.8$\pm$1.8&	59.3$\pm$9.5&	65.7$\pm$5.0&	71.7$\pm$16.9&	83.8$\pm$6.6&	62.8$\pm$23.5&	84.1$\pm$6.7\\
S-Model&	82.3$\pm$1.8&	82.0$\pm$1.9&	58.7$\pm$10.2&	64.6$\pm$5.1&	69.7$\pm$16.8&	82.4$\pm$7.1&	53.9$\pm$23.5&	77.6$\pm$10.2\\
Bootstrap&	83.1$\pm$1.6&	82.7$\pm$1.8&	60.1$\pm$9.7&	65.5$\pm$4.9&	71.7$\pm$16.9&	82.8$\pm$6.7&	55.6$\pm$23.9&	78.9$\pm$9.6\\
Mixup&	81.7$\pm$1.8&	82.4$\pm$1.7&	60.7$\pm$9.8&	66.0$\pm$4.9&	73.1$\pm$16.6&	85.0$\pm$6.2&	64.2$\pm$21.6&	82.5$\pm$8.0\\
MentorNet&	82.9$\pm$1.7&	82.4$\pm$1.7&	61.8$\pm$10.3&	65.1$\pm$5.0&	75.9$\pm$16.8&	82.6$\pm$6.6&	56.8$\pm$23.1&	78.9$\pm$8.9\\
{Our MentorMix}&	{\bf 84.2$\pm$0.7}&	{\bf 83.3$\pm$1.9}&	{\bf 70.9$\pm$3.4}&	{\bf 67.0$\pm$5.0}&	{\bf 78.2$\pm$16.2}&	{\bf 86.9$\pm$5.5}&	{\bf 67.7$\pm$23.0} &	{\bf 83.6$\pm$7.5}\\
\bottomrule
\end{tabular}
\vspace{-3mm}
\end{table*}
\subsection{Method Comparison}\label{sec:exp_method}

\noindent\textbf{Evaluation metrics}: Following prior works, the \emph{peak accuracy} is used as the primary evaluation metric that denotes the maximum accuracy on the clean validation set throughout the training. In addition, the \emph{final accuracy} is also reported in the Appendix D \ie~the validation accuracy after training has converged at the final training step.

\noindent\textbf{Training setting}: On our benchmark, all methods are trained on the noisy training sets of two noise types (blue and red) under 10 noise levels (from 0\% to 80\%), and tested on the same clean validation set. Two training settings are considered: (i) training from scratch and (ii) fine-tuning from an ImageNet checkpoint where the checkpoint is pretrained on the ImageNet training data. See details in the Appendix. On our datasets, Inception-ResNet-v2~\citep{szegedy2017inception} is used as the default backbone for all methods, where we upsample the images in the Mini-ImageNet dataset from 84x84 to 299x299 so that we can keep using the same pretrained ImageNet checkpoint. On the public benchmarks in Section~\ref{exp:method_soa}, we train networks from scratch using ResNet-32 for CIFAR and Inception-ResNet-v2 for WebVision.

\noindent\textbf{Baselines and our method}: On our dataset, MentorMix is compared against the following baselines. 
We extensively search the hyperparameter for each method on every noise level.
\textit{\textbf{Vanilla}} is the standard training using $l_2$ weight decay, dropout, and data augmentation. \textit{\textbf{Weight Decay}} and \textit{\textbf{Dropout}}~\citep{srivastava2014dropout} are classical regularization methods. We search the hyperparameter for the weight decay in $\{e^{-5}, e^{-4}, e^{-3}, e^{-2}\}$ and the dropout ratio in $\{0.9, 0.8, 0.7, 0.6, 0.5\}$ as suggested in~\citep{arpit2017closer}. \textbf{\emph{Bootstrap}}~\citep{reed2014training} corrects the loss with the learned label. The soft version is used and the hyperparameter for the learned label is tuned in $\{0.05, 0.25, 0.5, 0.7\}$. \textit{\textbf{S-model}}~\citep{goldberger2017training} is another way to ``correct'' the predictions by appending a new layer to a DNN to learn noise transformation. \textit{\textbf{MentorNet}}~\citep{jiang2018mentornet} is an example-weighting method. We employ the predefined MentorNet and search the hyperparameter $p$-percentile in $\{85\%, 75\%, 55\%, 35\%\}$. \textit{\textbf{Mixup}}~\citep{zhang2018mixup} is a robust learning method that minimizes the empirical vicinal risk. Following the advice in~\citep{zhang2018mixup} its hyperparameter $\alpha$ is searched in $\{1, 2, 4, 8\}$.

We implement MentorMix in Algorithm~\ref{alg:overall} in TensorFlow. We search two hyperparameters $\alpha$ in the $\gamma_p$ in the range $\alpha=\{0.4, 1, 2\}$ and $\gamma_p=\{90\%, 80\%, 70\%\}$. The code will be released to reproduce our results.

\subsubsection{Baseline Comparison}
We first show the comparison to the baseline methods on our dataset. For the lack of space, Table~\ref{tab:baseline} shows a high-level summary of the comparison result, in which each cell shows a method's average best peak accuracy (and 95\% confidence interval) across all 10 noise levels. Table 4 - Table 7 in the Appendix D list detailed results on each noise level.

As shown, the proposed method consistently outperforms the baseline methods across noise types (red and blue) and training settings (finetuning and training from scratch). This is desirable as baseline methods that work well on blue noise may not show consistent improvement over red noise, or vice versa. The red noise appears to be less harmful. Yet it is more difficult to overcome, which suggests the need for a new benchmark for a more comprehensive evaluation. Nevertheless, our method yields consistent improvements on both synthetic and web noisy labels. 

\begin{figure}[ht]
\vspace{-2mm}
\centering
\begin{subfigure}[b]{0.23\textwidth}
    \includegraphics[width=\textwidth]{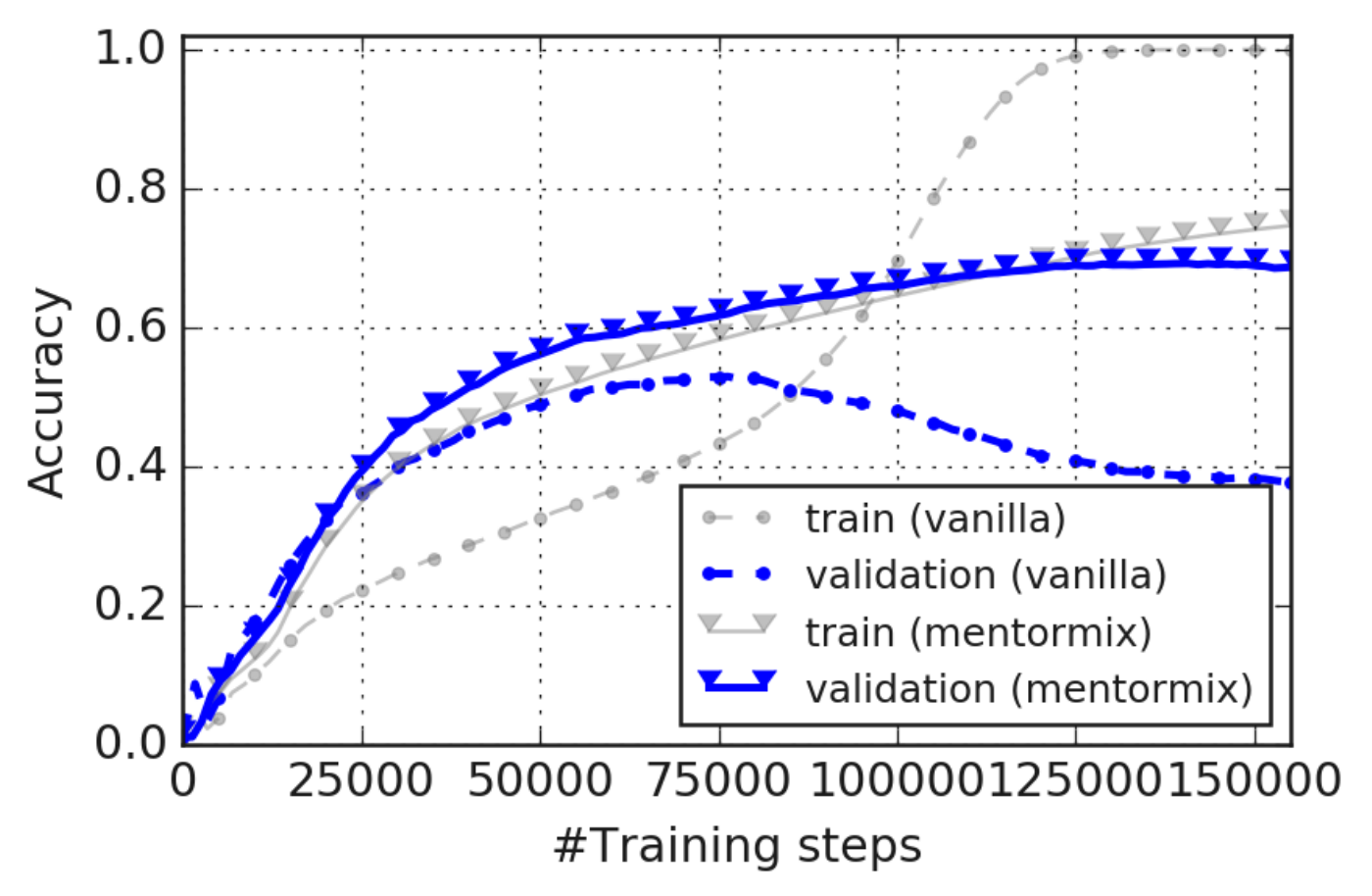}
    \vspace{-5mm}
    \caption{Blue Mini-ImageNet}
\end{subfigure}
\begin{subfigure}[b]{0.23\textwidth}
    \includegraphics[width=\textwidth]{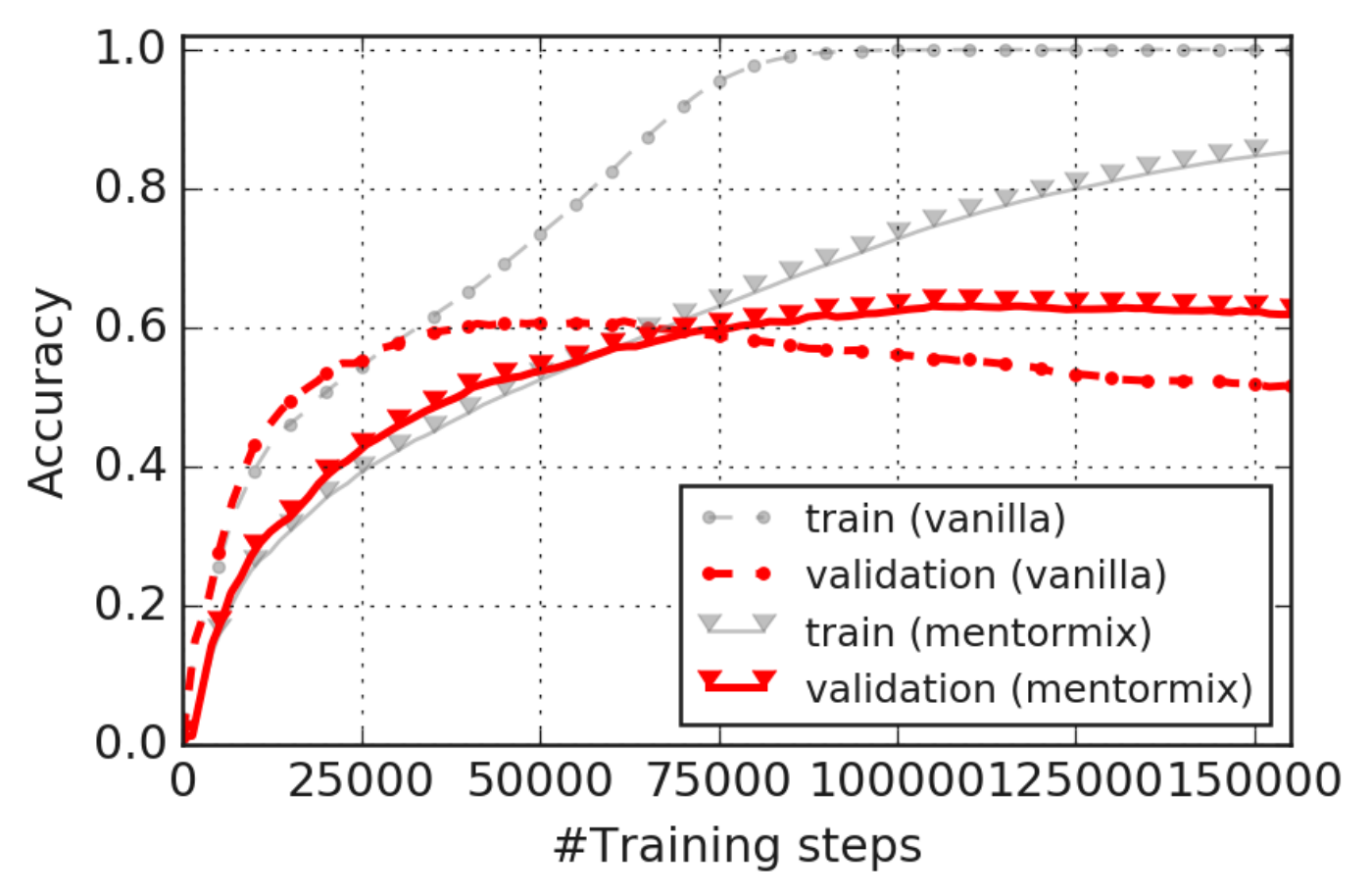}
    \vspace{-5mm}
    \caption{Red Mini-ImageNet}
\end{subfigure}
\vspace{-2mm}
\caption{\label{fig:mentormix_training}Comparison of training and validation accuracy during training. The dataset is Mimi-ImageNet at the 50\% noise level.}
\vspace{-3mm}
\end{figure}

The primary reason for our superior performance is that MentorMix can leverage MentorNet and Mixup in a complementary way. Technically, it uses MentorNet weight to identify examples with ``cleaner'' labels and encourages them to be used in the mixup operation. \cite{jiang2018mentornet} showed that MentorNet optimizes an underlying robust loss in empirical risk minimization. From this perspective, our MentorMix introduces a new robust loss to minimize the vicinal risk which turns out to be more resilient to noisy training labels. For example, Fig.~\ref{fig:mentormix_training} compares Vanilla and MentorMix on the blue and red Mini-ImageNet at 50\% noise level. It shows that MentorMix's loss is more robust to noisy labels, and MentorMix improves the peak accuracy of Vanilla by 16.3\% on blue noise and 2.4\% on red noise. 

As the noise levels span across a wide range, we find the hyperparameters of robust
learning methods are important. For the same method, a careful hyperparameter search could well be the difference between good and bad performance. As shown in the Appendix, we find that our method is relatively less sensitive to the choice of hyperparameters.

\begin{figure*}[ht!]
\vspace{-3mm}
\centering
\includegraphics[width=0.9\textwidth]{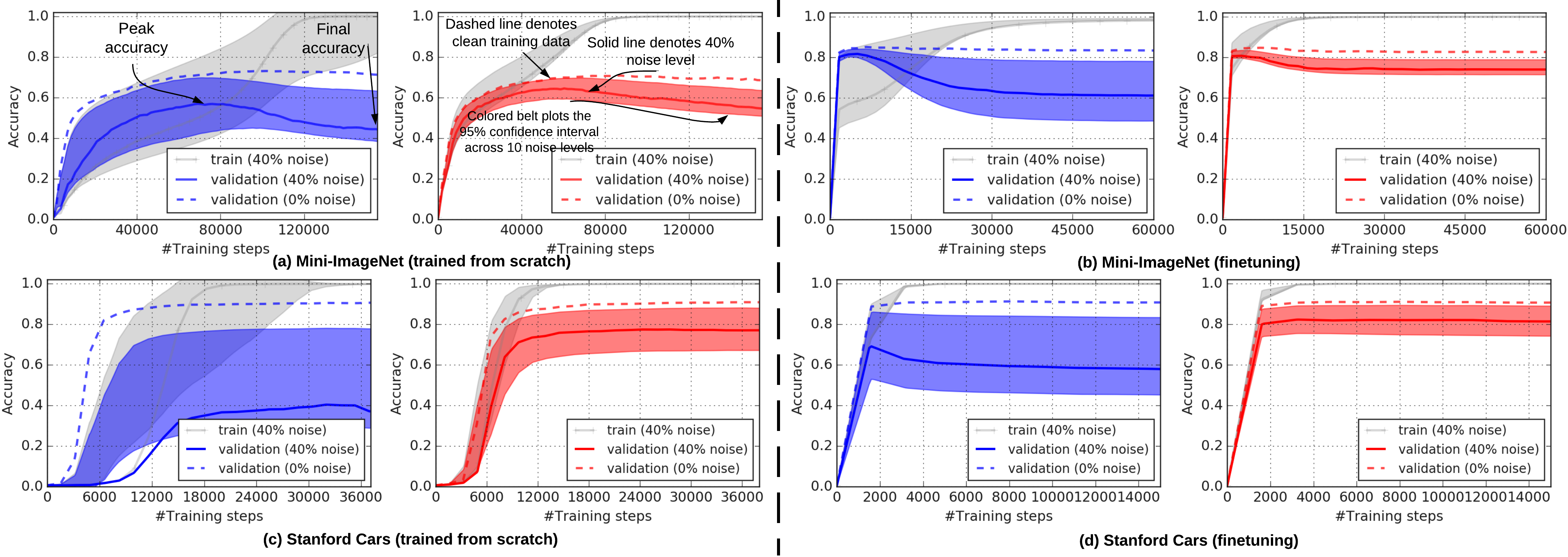}
\vspace{-3mm}
\caption{\label{fig:vanilla}DNNs trained on synthetic (blue) and web label noise (red) on Mini-ImageNet (top) and Stanford Cars (bottom).}
\vspace{-3mm}
\end{figure*}

\subsubsection{Comparison to the state-of-the-art}\label{exp:method_soa}
In this subsection, we compare MentorMix on two public benchmarks of synthetic and real-world noisy labels. Table~\ref{tab:cifar} compares with the state-of-the-art on the CIFAR dataset with symmetric label noise, where the top shows the classification accuracy on the clean validation set of CIFAR-100 and the bottom is for CIFAR-10. As all methods are trained using networks of similar capacity (ours is ResNet-32), we cite the numbers reported in their papers except for MentorNet and Mixup. Our method achieves the best accuracy across all noise levels. The results validate that our method is effective for synthetic noisy labels.

\begin{table}[ht!]
\vspace{-3mm}
\centering
\small
\caption{Comparison with the state-of-the-art in terms of the validation accuracy on CIFAR-100 (top) and CIFAR-10 (bottom).}
\label{tab:cifar}
\begin{tabular}{cccccc}
\toprule
\multirow{2}{*}{Data} & \multirow{2}{*}{ Method} & \multicolumn{4}{c}{ Noise level (\%)} \\
 &  & 20 & 40 & 60 & 80 \\ 
\midrule
\multirow{5}{*}{\rotatebox[origin=c]{90}{CIFAR100}} 
 & \citet{arazo2019unsupervised} & 73.7 & 70.1 & 59.5 & 39.5 \\
 & \citet{zhang2018generalized} & 67.6 & 62.6 &54.0 &29.6 \\
 & MentorNet~(\citeyear{jiang2018mentornet}) & 73.5 & 68.5 & 61.2 & 35.5 \\
 & Mixup~(\citeyear{zhang2018mixup}) & 73.9 & 66.8 & 58.8 & 40.1 \\
 & \citet{huang2019o2u}& 74.1&69.2&39.4&-\\
 & Ours (MentorMix) & {\bf 78.6} & {\bf 71.3} & {\bf 64.6} & {\bf 41.2} \\
 \hline
 \multirow{7}{*}{\rotatebox[origin=c]{90}{CIFAR10}} 
 & \citet{arazo2019unsupervised} & 94.0 & 92.8 & 90.3 & 74.1 \\
 & \citet{zhang2018generalized} & 89.7 & 87.6 &82.7 &67.9 \\
 & \citet{lee2019robust} & 87.1& 81.8& 75.4& -\\
 & \citet{chen2019understanding} & 89.7 & - & - & 52.3 \\
 & \citet{huang2019o2u}& 92.6&90.3&43.4&-\\
 & MentorNet~(\citeyear{jiang2018mentornet}) & 92.0 & 91.2 & 74.2 & 60.0 \\
 & Mixup~(\citeyear{zhang2018mixup}) & 94.0& 91.5& 86.8& 76.9\\
 & Ours (MentorMix)$\dagger$ & {\bf 95.6} & {\bf 94.2}& {\bf 91.3} & {\bf 81.0} \\ 
\bottomrule
\end{tabular}
\vspace{-3mm}
\end{table}

We then compare with the state-of-the-art on the challenging benchmark WebVision 1.0~\cite{li2017webvision} that contains 2.4 million training images of noisy labels from the web. It uses the same 1,000 classes in the ImageNet ILSVRC12 challenge and thus is evaluated on the same validation set. Following prior studies, we train our method on both the full training set (2.4M images on 1K classes) and the mini subset (61K images on 50 classes), and test it on two clean validation sets from ImageNet ILSVRC12 and WebVision.

Table~\ref{tab:webvision} shows the comparison results, where the method marked by $\dagger$ uses extra clean labels during training. As shown, the proposed MentorMix improves the prior state-of-the-art by about 3\% in the top-1 accuracy on the ILSVRC12 validation set without using any extra labels. It is worth noting that $3\%$ is a significant improvement on the ILSVRC12 validation~\citep{deng2009imagenet}. To the best of our knowledge, it achieves the best-published result on the WebVision 1.0 benchmark under the same training setting. The results show that our method is effective for real-world noisy labels. We also apply our method on the Clothing-1M dataset where we train only on the 1M noisy training examples. Our model gets 74.3\% accuracy which is competitive to recent published works.

\begin{table}[ht!]
\vspace{-3mm}
\centering
\caption{Comparison with the state-of-the-art on the clean validation set of ILSVRC12 and WebVision. The number outside (inside) the parentheses denotes the top-1 (top-5) classification accuracy(\%). $\dagger$ marks the method trained using extra clean labels.}
\label{tab:webvision}
\small
\begin{tabular}{c|lcc}
\toprule
{Data} & {Method} & {ILSVRC12} & {WebVision} \\
\midrule
Full& \citet{lee2017cleannet}$\dagger$& 61.0(82.0) & 69.1(86.7) \\
Full&Vanilla & 61.7(82.4) & 70.9(88.0) \\
Full& MentorNet~(\citeyear{jiang2018mentornet})$\dagger$& 64.2(84.8) & 72.6(88.9) \\
Full& \citet{guo2018curriculumnet}$\dagger$& 64.8(84.9) & 72.1(89.2) \\
Full& \citet{saxena2019data} & ---& 67.5(-----) \\
Full&Ours (MentorMix) & {\bf 67.5(87.2)} & {\bf 74.3(90.5)} \\
\hline
Mini& MentorNet~(\citeyear{jiang2018mentornet}) & 63.8(85.8) & --- \\
Mini& \citet{chen2019understanding} & 61.6(85.0) & 65.2(85.3) \\
Mini& Ours (MentorMix) & {\bf 72.9(91.1)} & {\bf 76.0(90.2)} \\
\bottomrule
\end{tabular}
\vspace{-3mm}
\end{table}

\subsection{Understanding DNNs trained on noisy labels} \label{sec:exp_understanding}
In this subsection, we conduct a large study into understanding DNNs trained on noisy labels across noise levels, noise types, training settings, and network architectures. We focus on three important findings~\citep{zhang2017understanding,arpit2017closer,kornblith2019better}. These works examine vanilla DNN training either on controlled synthetic labels (the former two) or on clean training labels (the last one). Our goal is to revisit them on our benchmark in a controlled setting where the noise in training sets varies from completely clean (0\%) to the level where 80\% of training labels are incorrect. 

As in these works, we learn DNNs using vanilla training which allows us to compare their findings. Training DNNs using robust learning methods would probably lead to different findings but that would make the findings undesirably depend on the specific method being used. Regarding the network architectures, by default we use Inception-ResNet-v2 and also compare six other architectures: EfficientNet-B5~\citep{tan2019efficientnet}, MobileNet-V2~\citep{sandler2018mobilenetv2}, ResNet-50 and ResNet-101~\citep{he2016deep}, Inception-V2~\citep{ioffe2015batch}, and Inception-V3~\citep{szegedy2016rethinking}. We select the above architectures to be representative of diverse capacities, the accuracy of which on the ILSVRC 2012 validation covers a wide range from 71.6\% to 83.6\%. See more details in the Appendix B.

{\bf DNNs generalize much better on red label noise.} \citet{zhang2017understanding} found that the generalization performance of DNNs drops sharply as the level of noisy training labels increases. This pivotal finding that DNNs generalize poorly on noisy training labels has influenced many works. We first confirm \citet{zhang2017understanding}'s finding on blue noise in Fig~\ref{fig:vanilla}, where the training and validation accuracies are shown along with the training steps. The dashed and solid curves represent the validation accuracies for 0\% (or clean) and 40\% noise levels, respectively, and the color belt plots the 95\% confidence interval of the 10 noise levels.

After the training converges in Fig~\ref{fig:vanilla}, the difference between the dashed and solid curves (in blue) indicates a palpable performance degradation between the clean (0\%) and noisy (40\%) labels. This can also be seen from the greater width of the blue belt which denotes the accuracy's confidence interval over all 10 noise levels. This confirms \citet{zhang2017understanding}'s finding on synthetic noisy labels. However, the difference is much smaller on the red curves, suggesting DNNs generalize much better on the red noise. This pattern is consistent across our two datasets using both fine-tuning and training from scratch. We hypothesize that DNNs are more robust to web labels because they are more relevant (visually or semantically) to the clean training images. See Fig. 1 in the Appendix.

\begin{figure}[ht]
\vspace{-3mm}
\centering
\includegraphics[width=0.35\textwidth]{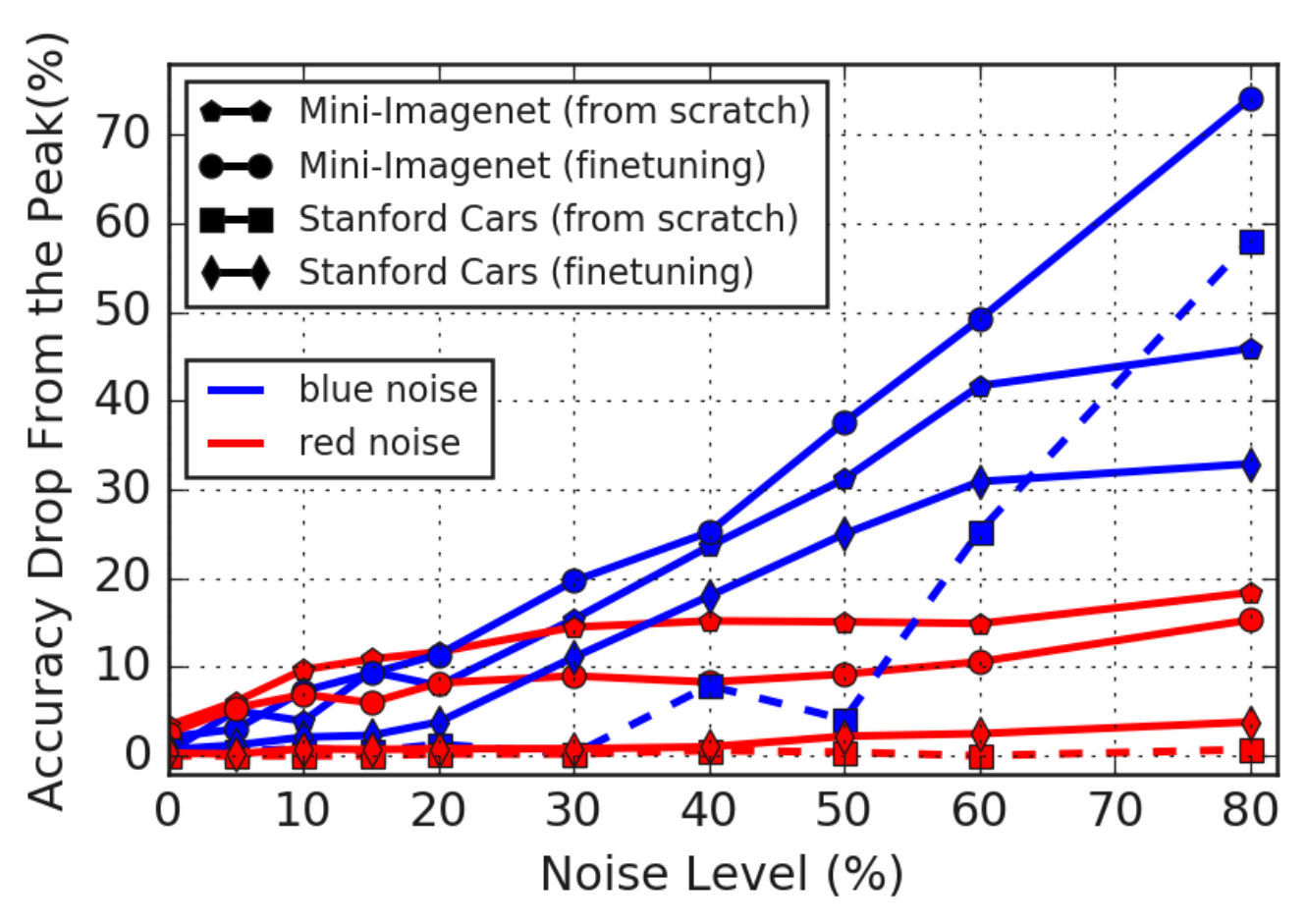}
\vspace{-5mm}
\caption{\label{fig:drop}Performance drop from the peak accuracy at different noise levels. Colors are used to differentiate noise types.}
\vspace{-3mm}
\end{figure}

{\bf DNNs may not learn patterns first on red label noise.} \citet{arpit2017closer} found that DNNs learn patterns first, revealing an interesting property that DNNs are able to automatically learn generalizable ``patterns'' in the early training stage before memorizing all noisy training labels. This can be manifested by the gap between the peak and final accuracy, as shown in Fig.~\ref{fig:vanilla}(a). A larger drop suggests a better pattern is found during the early training stage. For better visualization, Fig.~\ref{fig:drop} computes the relative difference, namely the drop, between the peak and final accuracy across noise levels. As shown, the blue curves show a significant drop as the noise level grows. This is consistent with \citet{arpit2017closer}'s finding on synthetic label noise.

Interestingly, the drop on the red noise is considerably smaller and even approaches zero on the Stanford Cars dataset. This suggests DNNs may not learn patterns first on red label noise at least for the fine-grained classification task. Our hypothesis is that images of real-world label noise are more complicated than those of the synthetic noise because they are sampled non-uniformly from an infinite number of classes. Therefore, it is much more difficult for DNNs to capture meaningful patterns automatically.

\textbf{ImageNet architectures generalize on noisy labels when the networks are fine-tuned.} \citet{kornblith2019better} found that fine-tuning better architectures trained on ImageNet tend to perform better on downstream tasks of clean training labels. It is important to verify whether this holds on noisy training labels because if so, one can conveniently transfer better architectures to better overcome the noisy labels.
\begin{figure}[ht]
\vspace{-1mm}
\centering
\begin{subfigure}[b]{0.35\textwidth}
    \includegraphics[width=\textwidth]{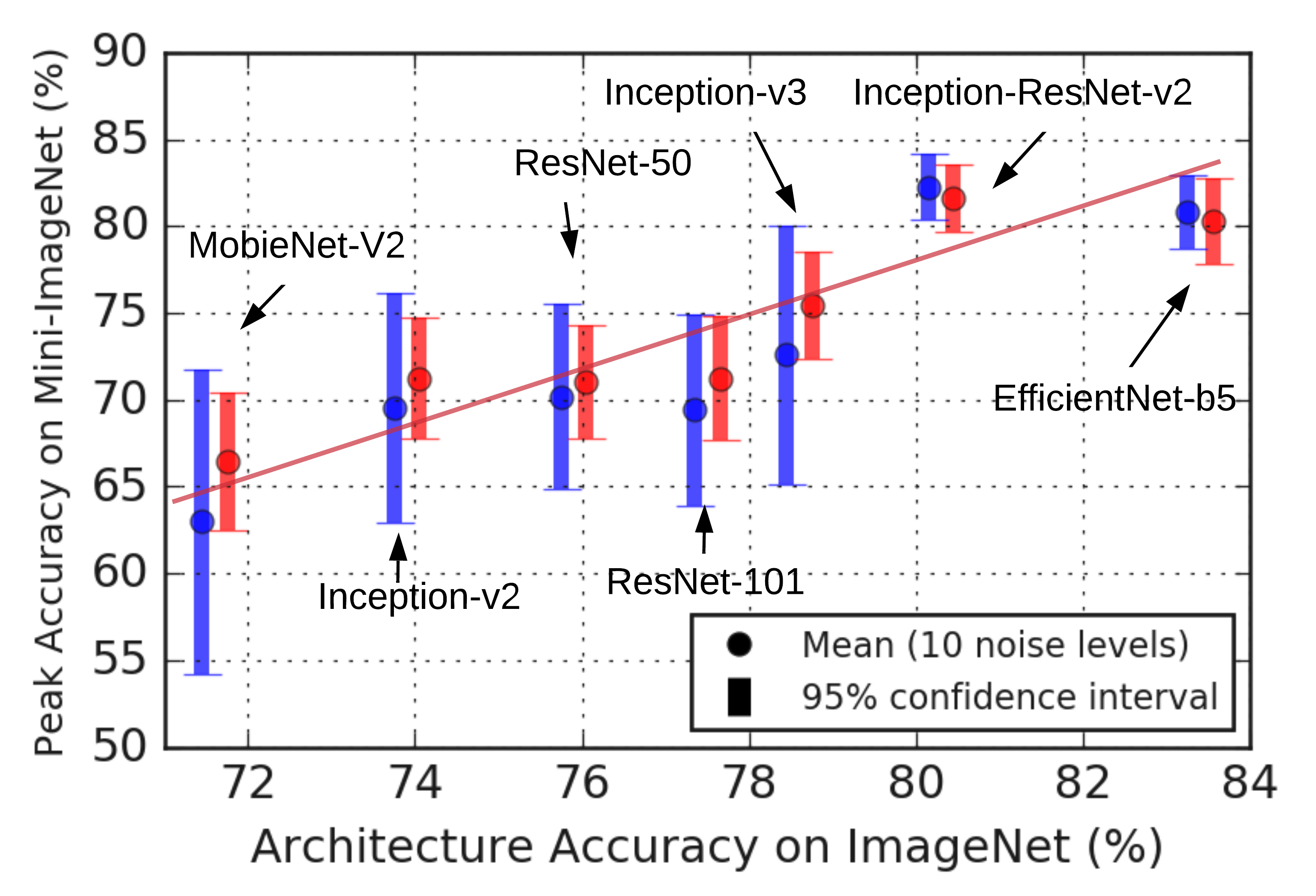}
    \vspace{-5mm}
    \caption{Mini-ImageNet ($r=0.91$)}
\end{subfigure}
\begin{subfigure}[b]{0.35\textwidth}
    \includegraphics[width=\textwidth]{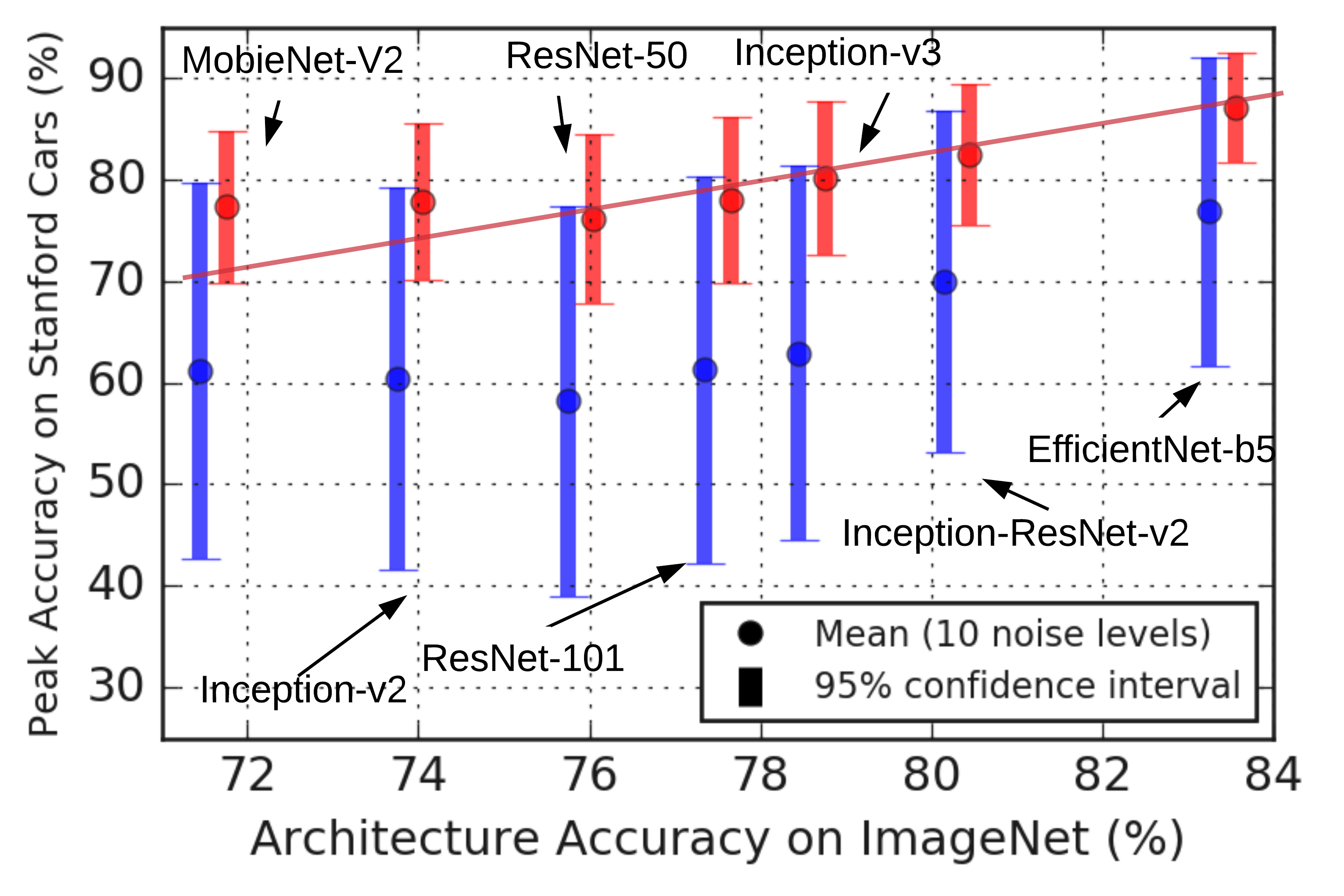}
    \vspace{-5mm}
    \caption{Stanford Cars ($r=0.88$)}
    \vspace{-3mm}
\end{subfigure}
\vspace{-2mm}
\caption{\label{fig:backbones}Fine-tuning seven ImageNet architectures on the red and blue datasets. The number in parentheses is the Pearson correlation between the architecture's ImageNet accuracy and the fine-tuning accuracy on our dataset of red noise.}
\vspace{-3mm}
\end{figure}

Following~\citep{kornblith2019better}, in Fig.~\ref{fig:backbones}, we compare the fine-tuning performance using ImageNet architectures, where the $x$-axis is the accuracy of the pretrained architectures on ImageNet, and the $y$-axis denotes the peak accuracy on our datasets. The bar plots the 95\% confidence interval across 10 noise levels, where the center dot marks the mean. As it shows, there is a reasonable correlation between the ImageNet accuracy and the validation accuracy on both red and blue noisy labels. The Pearson correlation for the red noise is 0.91 on Mini-ImageNet and 0.88 on Stanford Cars. This indicates a better-pretrained architecture is likely to perform better even when it is fine-tuned on noisy training labels. We do not find such correlation when these architectures are trained from scratch. These results extend \citet{kornblith2019better}'s finding to noisy training data, and suggest when possible one may use more advanced pretrained architectures to overcome noisy training labels.

\section{Conclusions}
In this paper, we study web label noise in a controlled setting. We make three contributions. First, we establish the first benchmark of controlled web noise obtained from image search. Second, a simple but effective method is proposed to overcome both synthetic and real-world noisy labels. Our method achieves state-of-the-art results on multiple datasets. 

Finally, we conduct the largest study by far into understanding deep learning on noisy data across a variety of settings. Our studies reveal several new findings: (1) DNNs generalize much better on web label noise; (2) DNNs may not learn patterns first on web label noise in which early stopping may not be very effective; (3) when networks are fine-tuned, ImageNet architectures generalize well on both symmetric and web label noise; (4) methods that perform well on synthetic noise may not work as well on the real-world noisy labels from the web; (5) the proposed method yields consistent improvements on both synthetic and real-world noisy labels from the web.

Based on our observations, we arrive at the following recommendations for training deep neural networks on noisy training data.
\begin{itemize}
    \item A simple way to deal with noisy labels is to fine-tune a model that is pre-trained on clean datasets, like ImageNet. The better the pre-trained model is, the better it may generalize on downstream noisy training tasks.
    \item Early stopping may not be as effective on the label noise from the web, especially on the fine-grained classification task.
    \item Methods that perform well on synthetic noise may not work as well on the real-world noisy labels from the web. The proposed MentorMix can better overcome both synthetic and real-world web noisy labels.
    \item The real-world label noise from the web appears to be less harmful, yet it is more difficult for our current robust learning methods to tackle. This encourages more future research to be carried out on controlled label noise from the web. 
\end{itemize}

\section*{Acknowledgements}
The authors would like to thank anonymous reviewers for helpful comments and Boqing Gong and Fei Sha for meaningful discussions and kind support. The data labeling is supported by the Google Cloud labeling service.


\bibliography{cr_paper}

\begin{thebibliography}{65}
\providecommand{\natexlab}[1]{#1}
\providecommand{\url}[1]{\texttt{#1}}
\expandafter\ifx\csname urlstyle\endcsname\relax
  \providecommand{\doi}[1]{doi: #1}\else
  \providecommand{\doi}{doi: \begingroup \urlstyle{rm}\Url}\fi

\bibitem[Arazo et~al.(2019)Arazo, Ortego, Albert, O'Connor, and
  McGuinness]{arazo2019unsupervised}
Arazo, E., Ortego, D., Albert, P., O'Connor, N.~E., and McGuinness, K.
\newblock Unsupervised label noise modeling and loss correction.
\newblock In \emph{International Conference on Machine Learning (ICML)}, 2019.

\bibitem[Arpit et~al.(2017)Arpit, Jastrz{k{e}}bski, Ballas, Krueger, Bengio,
  Kanwal, Maharaj, Fischer, Courville, Bengio, et~al.]{arpit2017closer}
Arpit, D., Jastrz{k{e}}bski, S., Ballas, N., Krueger, D., Bengio, E., Kanwal,
  M.~S., Maharaj, T., Fischer, A., Courville, A., Bengio, Y., et~al.
\newblock A closer look at memorization in deep networks.
\newblock In \emph{International Conference on Machine Learning (ICML)}, 2017.

\bibitem[Azadi et~al.(2016)Azadi, Feng, Jegelka, and
  Darrell]{azadi2016auxiliary}
Azadi, S., Feng, J., Jegelka, S., and Darrell, T.
\newblock Auxiliary image regularization for deep cnns with noisy labels.
\newblock In \emph{International Conference on Learning Representations
  (ICLR)}, 2016.

\bibitem[Bootkrajang \& Kab{\'a}n(2012)Bootkrajang and
  Kab{\'a}n]{bootkrajang2012label}
Bootkrajang, J. and Kab{\'a}n, A.
\newblock Label-noise robust logistic regression and its applications.
\newblock In \emph{European Conference on Machine Learning and Principles and
  Practice of Knowledge Discovery in Databases (ECML PKDD)}, 2012.

\bibitem[Charoenphakdee et~al.(2019)Charoenphakdee, Lee, and
  Sugiyama]{charoenphakdee2019symmetric}
Charoenphakdee, N., Lee, J., and Sugiyama, M.
\newblock On symmetric losses for learning from corrupted labels.
\newblock \emph{International Conference on Machine Learning (ICML)}, 2019.

\bibitem[Chen et~al.(2019)Chen, Liao, Chen, and Zhang]{chen2019understanding}
Chen, P., Liao, B., Chen, G., and Zhang, S.
\newblock Understanding and utilizing deep neural networks trained with noisy
  labels.
\newblock \emph{International Conference on Machine Learning (ICML)}, 2019.

\bibitem[Chen \& Gupta(2015)Chen and Gupta]{chen2015webly}
Chen, X. and Gupta, A.
\newblock Webly supervised learning of convolutional networks.
\newblock In \emph{International Conference on Computer Vision (ICCV)}, 2015.

\bibitem[Cheng et~al.(2019)Cheng, Jiang, and Macherey]{cheng2019robust}
Cheng, Y., Jiang, L., and Macherey, W.
\newblock Robust neural machine translation with doubly adversarial inputs.
\newblock In \emph{Annual Conference of the Association for Computational
  Linguistics (ACL)}, 2019.

\bibitem[Cheng et~al.(2020)Cheng, Jiang, Macherey, and
  Eisenstein]{cheng2020advaug}
Cheng, Y., Jiang, L., Macherey, W., and Eisenstein, J.
\newblock Advaug: Robust adversarial augmentation for neural machine
  translation.
\newblock In \emph{Annual Conference of the Association for Computational
  Linguistics (ACL)}, 2020.

\bibitem[Deng et~al.(2009)Deng, Dong, Socher, Li, Li, and
  Fei-Fei]{deng2009imagenet}
Deng, J., Dong, W., Socher, R., Li, L.-J., Li, K., and Fei-Fei, L.
\newblock Imagenet: A large-scale hierarchical image database.
\newblock In \emph{Conference on Computer Vision and Pattern Recognition
  (CVPR)}, 2009.

\bibitem[Fan et~al.(2017)Fan, He, Liang, and Hu]{fan2017self}
Fan, Y., He, R., Liang, J., and Hu, B.
\newblock Self-paced learning: an implicit regularization perspective.
\newblock In \emph{AAAI Conference on Artificial Intelligence (AAAI)}, 2017.

\bibitem[Goldberger \& Ben-Reuven(2017)Goldberger and
  Ben-Reuven]{goldberger2017training}
Goldberger, J. and Ben-Reuven, E.
\newblock Training deep neural-networks using a noise adaptation layer.
\newblock In \emph{International Conference on Learning Representations
  (ICLR)}, 2017.

\bibitem[Guo et~al.(2018)Guo, Huang, Zhang, Zhuang, Dong, Scott, and
  Huang]{guo2018curriculumnet}
Guo, S., Huang, W., Zhang, H., Zhuang, C., Dong, D., Scott, M.~R., and Huang,
  D.
\newblock Curriculumnet: Weakly supervised learning from large-scale web
  images.
\newblock In \emph{European Conference on Computer Vision (ECCV)}, 2018.

\bibitem[Han et~al.(2018{\natexlab{a}})Han, Yao, Niu, Zhou, Tsang, Zhang, and
  Sugiyama]{han2018masking}
Han, B., Yao, J., Niu, G., Zhou, M., Tsang, I., Zhang, Y., and Sugiyama, M.
\newblock Masking: A new perspective of noisy supervision.
\newblock In \emph{Conference on Neural Information Processing Systems
  (NeurIPS)}, 2018{\natexlab{a}}.

\bibitem[Han et~al.(2018{\natexlab{b}})Han, Yao, Yu, Niu, Xu, Hu, Tsang, and
  Sugiyama]{han2018co}
Han, B., Yao, Q., Yu, X., Niu, G., Xu, M., Hu, W., Tsang, I., and Sugiyama, M.
\newblock Co-teaching: Robust training of deep neural networks with extremely
  noisy labels.
\newblock In \emph{Conference on Neural Information Processing Systems
  (NeurIPS)}, 2018{\natexlab{b}}.

\bibitem[He et~al.(2016)He, Zhang, Ren, and Sun]{he2016deep}
He, K., Zhang, X., Ren, S., and Sun, J.
\newblock Deep residual learning for image recognition.
\newblock In \emph{Conference on Computer Vision and Pattern Recognition
  (CVPR)}, 2016.

\bibitem[Hendrycks \& Dietterich(2019)Hendrycks and
  Dietterich]{hendrycks2019benchmarking}
Hendrycks, D. and Dietterich, T.
\newblock Benchmarking neural network robustness to common corruptions and
  perturbations.
\newblock In \emph{International Conference on Learning Representations
  (ICLR)}, 2019.

\bibitem[Hendrycks et~al.(2018)Hendrycks, Mazeika, Wilson, and
  Gimpel]{hendrycks2018using}
Hendrycks, D., Mazeika, M., Wilson, D., and Gimpel, K.
\newblock Using trusted data to train deep networks on labels corrupted by
  severe noise.
\newblock In \emph{Conference on Neural Information Processing Systems
  (NeurIPS)}, 2018.

\bibitem[Huang et~al.(2019)Huang, Qu, Jia, and Zhao]{huang2019o2u}
Huang, J., Qu, L., Jia, R., and Zhao, B.
\newblock O2u-net: A simple noisy label detection approach for deep neural
  networks.
\newblock In \emph{International Conference on Computer Vision (ICCV)}, 2019.

\bibitem[Ioffe \& Szegedy(2015)Ioffe and Szegedy]{ioffe2015batch}
Ioffe, S. and Szegedy, C.
\newblock Batch normalization: Accelerating deep network training by reducing
  internal covariate shift.
\newblock In \emph{International Conference on Machine Learning (ICML)}, 2015.

\bibitem[Jiang et~al.(2014)Jiang, Meng, Yu, Lan, Shan, and
  Hauptmann]{jiang2014self}
Jiang, L., Meng, D., Yu, S.-I., Lan, Z., Shan, S., and Hauptmann, A.
\newblock Self-paced learning with diversity.
\newblock In \emph{Conference on Neural Information Processing Systems
  (NeurIPS)}, 2014.

\bibitem[Jiang et~al.(2015)Jiang, Meng, Zhao, Shan, and
  Hauptmann]{jiang2015self}
Jiang, L., Meng, D., Zhao, Q., Shan, S., and Hauptmann, A.~G.
\newblock Self-paced curriculum learning.
\newblock In \emph{AAAI Conference on Artificial Intelligence (AAAI)}, 2015.

\bibitem[Jiang et~al.(2018)Jiang, Zhou, Leung, Li, and
  Fei-Fei]{jiang2018mentornet}
Jiang, L., Zhou, Z., Leung, T., Li, L.-J., and Fei-Fei, L.
\newblock Mentornet: Learning data-driven curriculum for very deep neural
  networks on corrupted labels.
\newblock \emph{International Conference on Machine Learning (ICML)}, 2018.

\bibitem[Kornblith et~al.(2019)Kornblith, Shlens, and Le]{kornblith2019better}
Kornblith, S., Shlens, J., and Le, Q.~V.
\newblock Do better imagenet models transfer better?
\newblock In \emph{Conference on Computer Vision and Pattern Recognition
  (CVPR)}, 2019.

\bibitem[Krause et~al.(2013)Krause, Deng, Stark, and
  Fei-Fei]{krause2013collecting}
Krause, J., Deng, J., Stark, M., and Fei-Fei, L.
\newblock Collecting a large-scale dataset of fine-grained cars.
\newblock In \emph{Second Workshop on Fine-Grained Visual Categorization},
  2013.

\bibitem[Krause et~al.(2016)Krause, Sapp, Howard, Zhou, Toshev, Duerig,
  Philbin, and Fei-Fei]{krause2016unreasonable}
Krause, J., Sapp, B., Howard, A., Zhou, H., Toshev, A., Duerig, T., Philbin,
  J., and Fei-Fei, L.
\newblock The unreasonable effectiveness of noisy data for fine-grained
  recognition.
\newblock In \emph{European Conference on Computer Vision (ECCV)}, 2016.

\bibitem[Kumar et~al.(2010)Kumar, Packer, and Koller]{kumar2010self}
Kumar, M.~P., Packer, B., and Koller, D.
\newblock Self-paced learning for latent variable models.
\newblock In \emph{Conference on Neural Information Processing Systems
  (NeurIPS)}, 2010.

\bibitem[Lee et~al.(2019)Lee, Yun, Lee, Lee, Li, and Shin]{lee2019robust}
Lee, K., Yun, S., Lee, K., Lee, H., Li, B., and Shin, J.
\newblock Robust inference via generative classifiers for handling noisy
  labels.
\newblock \emph{International Conference on Machine Learning (ICML)}, 2019.

\bibitem[Lee et~al.(2018)Lee, He, Zhang, and Yang]{lee2017cleannet}
Lee, K.-H., He, X., Zhang, L., and Yang, L.
\newblock Cleannet: Transfer learning for scalable image classifier training
  with label noise.
\newblock \emph{Conference on Computer Vision and Pattern Recognition (CVPR)},
  2018.

\bibitem[Li et~al.(2019)Li, Wong, Zhao, and Kankanhalli]{li2019learning}
Li, J., Wong, Y., Zhao, Q., and Kankanhalli, M.~S.
\newblock Learning to learn from noisy labeled data.
\newblock In \emph{Conference on Computer Vision and Pattern Recognition
  (CVPR)}, 2019.

\bibitem[Li et~al.(2020)Li, Socher, and Hoi]{li2020dividemix}
Li, J., Socher, R., and Hoi, S.~C.
\newblock Dividemix: Learning with noisy labels as semi-supervised learning.
\newblock In \emph{International Conference on Learning Representations
  (ICLR)}, 2020.

\bibitem[Li et~al.(2017{\natexlab{a}})Li, Wang, Li, Agustsson, and
  Van~Gool]{li2017webvision}
Li, W., Wang, L., Li, W., Agustsson, E., and Van~Gool, L.
\newblock Webvision database: Visual learning and understanding from web data.
\newblock \emph{arXiv preprint arXiv:1708.02862}, 2017{\natexlab{a}}.

\bibitem[Li et~al.(2017{\natexlab{b}})Li, Yang, Song, Cao, Li, and
  Luo]{Li2017ICCV}
Li, Y., Yang, J., Song, Y., Cao, L., Li, J., and Luo, J.
\newblock Learning from noisy labels with distillation.
\newblock In \emph{International Conference on Computer Vision (ICCV)},
  2017{\natexlab{b}}.

\bibitem[Liang et~al.(2016)Liang, Jiang, Meng, and
  Hauptmann]{liang2016learning}
Liang, J., Jiang, L., Meng, D., and Hauptmann, A.~G.
\newblock Learning to detect concepts from webly-labeled video data.
\newblock In \emph{International Joint Conference on Artificial Intelligence
  (IJCAI)}, 2016.

\bibitem[Liang et~al.(2020)Liang, Jiang, and Hauptmann]{liang2020simaug}
Liang, J., Jiang, L., and Hauptmann, A.
\newblock Simaug: Learning robust representations from simulation for
  trajectory prediction.
\newblock In \emph{European Conference on Computer Vision (ECCV)}, 2020.

\bibitem[Ma et~al.(2018)Ma, Wang, Houle, Zhou, Erfani, Xia, Wijewickrema, and
  Bailey]{ma2018dimensionality}
Ma, X., Wang, Y., Houle, M.~E., Zhou, S., Erfani, S.~M., Xia, S.-T.,
  Wijewickrema, S., and Bailey, J.
\newblock Dimensionality-driven learning with noisy labels.
\newblock In \emph{International Conference on Machine Learning (ICML)}, 2018.

\bibitem[Mahajan et~al.(2018)Mahajan, Girshick, Ramanathan, He, Paluri, Li,
  Bharambe, and van~der Maaten]{mahajan2018exploring}
Mahajan, D., Girshick, R., Ramanathan, V., He, K., Paluri, M., Li, Y.,
  Bharambe, A., and van~der Maaten, L.
\newblock Exploring the limits of weakly supervised pretraining.
\newblock In \emph{European Conference on Computer Vision (ECCV)}, 2018.

\bibitem[Mnih \& Hinton(2012)Mnih and Hinton]{mnih2012learning}
Mnih, V. and Hinton, G.~E.
\newblock Learning to label aerial images from noisy data.
\newblock In \emph{International Conference on Machine Learning (ICML)}, 2012.

\bibitem[Noh et~al.(2017)Noh, You, Mun, and Han]{noh2017regularizing}
Noh, H., You, T., Mun, J., and Han, B.
\newblock Regularizing deep neural networks by noise: its interpretation and
  optimization.
\newblock In \emph{Conference on Neural Information Processing Systems
  (NeurIPS)}, 2017.

\bibitem[Northcutt et~al.(2019)Northcutt, Jiang, and
  Chuang]{northcutt2019confident}
Northcutt, C.~G., Jiang, L., and Chuang, I.~L.
\newblock Confident learning: Estimating uncertainty in dataset labels.
\newblock \emph{arXiv preprint arXiv:1911.00068}, 2019.

\bibitem[Patrini et~al.(2017)Patrini, Rozza, Krishna~Menon, Nock, and
  Qu]{patrini2017making}
Patrini, G., Rozza, A., Krishna~Menon, A., Nock, R., and Qu, L.
\newblock Making deep neural networks robust to label noise: A loss correction
  approach.
\newblock In \emph{Conference on Computer Vision and Pattern Recognition
  (CVPR)}, 2017.

\bibitem[Reed et~al.(2014)Reed, Lee, Anguelov, Szegedy, Erhan, and
  Rabinovich]{reed2014training}
Reed, S., Lee, H., Anguelov, D., Szegedy, C., Erhan, D., and Rabinovich, A.
\newblock Training deep neural networks on noisy labels with bootstrapping.
\newblock \emph{arXiv preprint arXiv:1412.6596}, 2014.

\bibitem[Reeve \& Kab{\'a}n(2019)Reeve and Kab{\'a}n]{reeve2019fast}
Reeve, H.~W. and Kab{\'a}n, A.
\newblock Fast rates for a knn classifier robust to unknown asymmetric label
  noise.
\newblock \emph{International Conference on Machine Learning (ICML)}, 2019.

\bibitem[Ren et~al.(2018)Ren, Zeng, Yang, and Urtasun]{ren2018learning}
Ren, M., Zeng, W., Yang, B., and Urtasun, R.
\newblock Learning to reweight examples for robust deep learning.
\newblock In \emph{International Conference on Machine Learning (ICML)}, 2018.

\bibitem[Rolnick et~al.(2017)Rolnick, Veit, Belongie, and
  Shavit]{rolnick2017deep}
Rolnick, D., Veit, A., Belongie, S., and Shavit, N.
\newblock Deep learning is robust to massive label noise.
\newblock \emph{arXiv preprint arXiv:1705.10694}, 2017.

\bibitem[Sandler et~al.(2018)Sandler, Howard, Zhu, Zhmoginov, and
  Chen]{sandler2018mobilenetv2}
Sandler, M., Howard, A., Zhu, M., Zhmoginov, A., and Chen, L.-C.
\newblock Mobilenetv2: Inverted residuals and linear bottlenecks.
\newblock In \emph{Conference on Computer Vision and Pattern Recognition
  (CVPR)}, 2018.

\bibitem[Saxena et~al.(2019)Saxena, Tuzel, and DeCoste]{saxena2019data}
Saxena, S., Tuzel, O., and DeCoste, D.
\newblock Data parameters: A new family of parameters for learning a
  differentiable curriculum.
\newblock In \emph{Conference on Neural Information Processing Systems
  (NeurIPS)}, 2019.

\bibitem[Seo et~al.(2019)Seo, Kim, and Han]{seo2019combinatorial}
Seo, P.~H., Kim, G., and Han, B.
\newblock Combinatorial inference against label noise.
\newblock In \emph{Conference on Neural Information Processing Systems
  (NeurIPS)}, 2019.

\bibitem[Shu et~al.(2019)Shu, Xie, Yi, Zhao, Zhou, Xu, and Meng]{shu2019meta}
Shu, J., Xie, Q., Yi, L., Zhao, Q., Zhou, S., Xu, Z., and Meng, D.
\newblock Meta-weight-net: Learning an explicit mapping for sample weighting.
\newblock In \emph{Conference on Neural Information Processing Systems
  (NeurIPS)}, 2019.

\bibitem[Song et~al.(2019)Song, Kim, and Lee]{song2019selfie}
Song, H., Kim, M., and Lee, J.-G.
\newblock Selfie: Refurbishing unclean samples for robust deep learning.
\newblock In \emph{International Conference on Machine Learning (ICML)}, 2019.

\bibitem[Srivastava et~al.(2014)Srivastava, Hinton, Krizhevsky, Sutskever, and
  Salakhutdinov]{srivastava2014dropout}
Srivastava, N., Hinton, G., Krizhevsky, A., Sutskever, I., and Salakhutdinov,
  R.
\newblock Dropout: a simple way to prevent neural networks from overfitting.
\newblock \emph{Journal of Machine Learning Research (JMLR)}, 15\penalty0
  (1):\penalty0 1929--1958, 2014.

\bibitem[Szegedy et~al.(2016)Szegedy, Vanhoucke, Ioffe, Shlens, and
  Wojna]{szegedy2016rethinking}
Szegedy, C., Vanhoucke, V., Ioffe, S., Shlens, J., and Wojna, Z.
\newblock Rethinking the inception architecture for computer vision.
\newblock In \emph{Conference on Computer Vision and Pattern Recognition
  (CVPR)}, 2016.

\bibitem[Szegedy et~al.(2017)Szegedy, Ioffe, Vanhoucke, and
  Alemi]{szegedy2017inception}
Szegedy, C., Ioffe, S., Vanhoucke, V., and Alemi, A.~A.
\newblock Inception-v4, inception-resnet and the impact of residual connections
  on learning.
\newblock In \emph{AAAI Conference on Artificial Intelligence (AAAI)}, 2017.

\bibitem[Tan \& Le(2019)Tan and Le]{tan2019efficientnet}
Tan, M. and Le, Q.~V.
\newblock Efficientnet: Rethinking model scaling for convolutional neural
  networks.
\newblock In \emph{International Conference on Machine Learning (ICML)}, 2019.

\bibitem[Vahdat(2017)]{vahdat2017toward}
Vahdat, A.
\newblock Toward robustness against label noise in training deep discriminative
  neural networks.
\newblock In \emph{Conference on Neural Information Processing Systems
  (NeurIPS)}, 2017.

\bibitem[Van~Rooyen et~al.(2015)Van~Rooyen, Menon, and
  Williamson]{van2015learning}
Van~Rooyen, B., Menon, A., and Williamson, R.~C.
\newblock Learning with symmetric label noise: The importance of being
  unhinged.
\newblock In \emph{Conference on Neural Information Processing Systems
  (NeurIPS)}, 2015.

\bibitem[Veit et~al.(2017)Veit, Alldrin, Chechik, Krasin, Gupta, and
  Belongie]{veit2017learning}
Veit, A., Alldrin, N., Chechik, G., Krasin, I., Gupta, A., and Belongie, S.
\newblock Learning from noisy large-scale datasets with minimal supervision.
\newblock In \emph{Conference on Computer Vision and Pattern Recognition
  (CVPR)}, 2017.

\bibitem[Vinyals et~al.(2016)Vinyals, Blundell, Lillicrap, Wierstra,
  et~al.]{vinyals2016matching}
Vinyals, O., Blundell, C., Lillicrap, T., Wierstra, D., et~al.
\newblock Matching networks for one shot learning.
\newblock In \emph{Conference on Neural Information Processing Systems
  (NeurIPS)}, 2016.

\bibitem[Wang et~al.(2018)Wang, Liu, Ma, Bailey, Zha, Song, and
  Xia]{wang2018iterative}
Wang, Y., Liu, W., Ma, X., Bailey, J., Zha, H., Song, L., and Xia, S.-T.
\newblock Iterative learning with open-set noisy labels.
\newblock In \emph{Conference on Computer Vision and Pattern Recognition
  (CVPR)}, 2018.

\bibitem[Xiao et~al.(2015)Xiao, Xia, Yang, Huang, and Wang]{xiao2015learning}
Xiao, T., Xia, T., Yang, Y., Huang, C., and Wang, X.
\newblock Learning from massive noisy labeled data for image classification.
\newblock In \emph{Conference on Computer Vision and Pattern Recognition
  (CVPR)}, 2015.

\bibitem[Zhang et~al.(2017)Zhang, Bengio, Hardt, Recht, and
  Vinyals]{zhang2017understanding}
Zhang, C., Bengio, S., Hardt, M., Recht, B., and Vinyals, O.
\newblock Understanding deep learning requires rethinking generalization.
\newblock In \emph{International Conference on Learning Representations
  (ICLR)}, 2017.

\bibitem[Zhang et~al.(2018)Zhang, Cisse, Dauphin, and
  Lopez-Paz]{zhang2018mixup}
Zhang, H., Cisse, M., Dauphin, Y.~N., and Lopez-Paz, D.
\newblock mixup: Beyond empirical risk minimization.
\newblock In \emph{International Conference on Learning Representations
  (ICLR)}, 2018.

\bibitem[Zhang et~al.(2019)Zhang, Yu, Jiao, Xing, Ghaoui, and
  Jordan]{zhang2019theoretically}
Zhang, H., Yu, Y., Jiao, J., Xing, E.~P., Ghaoui, L.~E., and Jordan, M.~I.
\newblock Theoretically principled trade-off between robustness and accuracy.
\newblock \emph{International Conference on Machine Learning (ICML)}, 2019.

\bibitem[Zhang \& Sabuncu(2018)Zhang and Sabuncu]{zhang2018generalized}
Zhang, Z. and Sabuncu, M.
\newblock Generalized cross entropy loss for training deep neural networks with
  noisy labels.
\newblock In \emph{Conference on Neural Information Processing Systems
  (NeurIPS)}, 2018.

\bibitem[Zhang et~al.(2020)Zhang, Zhang, Arik, Lee, and
  Pfister]{zhang2020distilling}
Zhang, Z., Zhang, H., Arik, S.~O., Lee, H., and Pfister, T.
\newblock Distilling effective supervision from severe label noise.
\newblock In \emph{Conference on Computer Vision and Pattern Recognition
  (CVPR)}, 2020.

\end{thebibliography}


\begin{thebibliography}{12}
\providecommand{\natexlab}[1]{#1}
\providecommand{\url}[1]{\texttt{#1}}
\expandafter\ifx\csname urlstyle\endcsname\relax
  \providecommand{\doi}[1]{doi: #1}\else
  \providecommand{\doi}{doi: \begingroup \urlstyle{rm}\Url}\fi

\bibitem[Goyal et~al.(2017)Goyal, Doll{\'a}r, Girshick, Noordhuis, Wesolowski,
  Kyrola, Tulloch, Jia, and He]{goyal2017accurate}
Goyal, P., Doll{\'a}r, P., Girshick, R., Noordhuis, P., Wesolowski, L., Kyrola,
  A., Tulloch, A., Jia, Y., and He, K.
\newblock Accurate, large minibatch sgd: Training imagenet in 1 hour.
\newblock \emph{arXiv preprint arXiv:1706.02677}, 2017.

\bibitem[He et~al.(2016)He, Zhang, Ren, and Sun]{he2016deep}
He, K., Zhang, X., Ren, S., and Sun, J.
\newblock Deep residual learning for image recognition.
\newblock In \emph{Conference on Computer Vision and Pattern Recognition
  (CVPR)}, 2016.

\bibitem[Ioffe \& Szegedy(2015)Ioffe and Szegedy]{ioffe2015batch}
Ioffe, S. and Szegedy, C.
\newblock Batch normalization: Accelerating deep network training by reducing
  internal covariate shift.
\newblock In \emph{International Conference on Machine Learning (ICML)}, 2015.

\bibitem[Jiang et~al.(2018)Jiang, Zhou, Leung, Li, and
  Fei-Fei]{jiang2018mentornet}
Jiang, L., Zhou, Z., Leung, T., Li, L.-J., and Fei-Fei, L.
\newblock Mentornet: Learning data-driven curriculum for very deep neural
  networks on corrupted labels.
\newblock \emph{International Conference on Machine Learning (ICML)}, 2018.

\bibitem[Kornblith et~al.(2019)Kornblith, Shlens, and Le]{kornblith2019better}
Kornblith, S., Shlens, J., and Le, Q.~V.
\newblock Do better imagenet models transfer better?
\newblock In \emph{Conference on Computer Vision and Pattern Recognition
  (CVPR)}, 2019.

\bibitem[Krizhevsky \& Hinton(2009)Krizhevsky and
  Hinton]{krizhevsky2009learning}
Krizhevsky, A. and Hinton, G.
\newblock Learning multiple layers of features from tiny images.
\newblock 2009.

\bibitem[Li et~al.(2017)Li, Wang, Li, Agustsson, and Van~Gool]{li2017webvision}
Li, W., Wang, L., Li, W., Agustsson, E., and Van~Gool, L.
\newblock Webvision database: Visual learning and understanding from web data.
\newblock \emph{arXiv preprint arXiv:1708.02862}, 2017.

\bibitem[Sandler et~al.(2018)Sandler, Howard, Zhu, Zhmoginov, and
  Chen]{sandler2018mobilenetv2}
Sandler, M., Howard, A., Zhu, M., Zhmoginov, A., and Chen, L.-C.
\newblock Mobilenetv2: Inverted residuals and linear bottlenecks.
\newblock In \emph{Conference on Computer Vision and Pattern Recognition
  (CVPR)}, 2018.

\bibitem[Szegedy et~al.(2016)Szegedy, Vanhoucke, Ioffe, Shlens, and
  Wojna]{szegedy2016rethinking}
Szegedy, C., Vanhoucke, V., Ioffe, S., Shlens, J., and Wojna, Z.
\newblock Rethinking the inception architecture for computer vision.
\newblock In \emph{Conference on Computer Vision and Pattern Recognition
  (CVPR)}, 2016.

\bibitem[Szegedy et~al.(2017)Szegedy, Ioffe, Vanhoucke, and
  Alemi]{szegedy2017inception}
Szegedy, C., Ioffe, S., Vanhoucke, V., and Alemi, A.~A.
\newblock Inception-v4, inception-resnet and the impact of residual connections
  on learning.
\newblock In \emph{AAAI Conference on Artificial Intelligence (AAAI)}, 2017.

\bibitem[Tan \& Le(2019)Tan and Le]{tan2019efficientnet}
Tan, M. and Le, Q.~V.
\newblock Efficientnet: Rethinking model scaling for convolutional neural
  networks.
\newblock In \emph{International Conference on Machine Learning (ICML)}, 2019.

\bibitem[Xiao et~al.(2015)Xiao, Xia, Yang, Huang, and Wang]{xiao2015learning}
Xiao, T., Xia, T., Yang, Y., Huang, C., and Wang, X.
\newblock Learning from massive noisy labeled data for image classification.
\newblock In \emph{Conference on Computer Vision and Pattern Recognition
  (CVPR)}, 2015.

\end{thebibliography}
\bibliographystyle{icml2020}

\end{document}


\onecolumn

\icmltitle{Supplementary Materials for
Beyond Synthetic Noise: Deep Learning on Controlled Noisy Labels}

\begin{icmlauthorlist}
\icmlauthor{Lu Jiang}{}
\icmlauthor{Di Huang}{}
\icmlauthor{Mason Liu}{}
\icmlauthor{Weilong Yang}{}
\end{icmlauthorlist}



\appendix

\begin{figure}[ht!]
\centering
\includegraphics[width=0.99\linewidth]{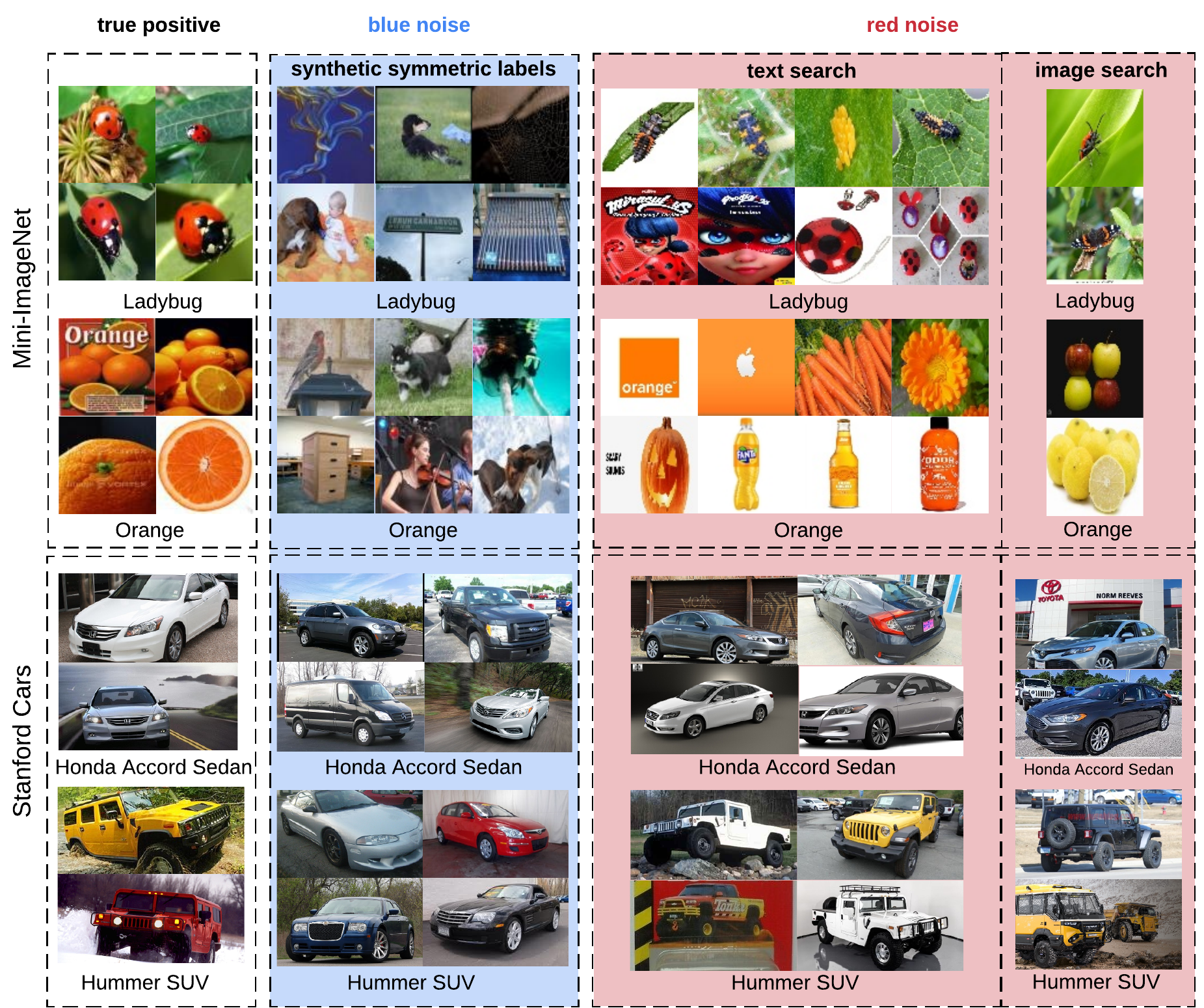}
\caption{\label{fig:dataset}Comparison of symmetric label noise (Blue noise) and web label noise (Red noise). From left to right, columns are true positive images, images with incorrect symmetric labels, and images with incorrect web labels from text-to-image search and image-to-image search, respectively. The image-to-image search results (the last column) only account for 18\% in our dataset and fewer images are shown as a result.}
\end{figure}

\section{Dataset Overview:}

\begin{table}[ht]
\centering
\caption{Summary of the difference of images with blue and red noisy labels.}
\label{tab:noise_difference}
\begin{tabular}{lll}
\toprule
Difference & Blue Noise & Red Noise \\
\midrule
Visual \& semantic similarity to true positive images & Low & High \\
Level of the label noise&  instance-level & class-level \\
Latent class vocabulary from which images are sampled& Fixed vocabulary &  Open vocabulary\\
\bottomrule
\end{tabular}
\end{table}
 
Fig.~\ref{fig:dataset} shows some example images with correct labels and incorrect labels of symmetric label noise (blue noise) and web label noise (red noise). There are three clear distinctions between images with the synthetic and web label noise as summarized in Table~\ref{tab:noise_difference}. First, images with noise from the web (or red noise) are more relevant (visually or semantically) to true positive images. Second, synthetic noise (symmetric or asymmetric) is at class-level which means all examples in the same class are treated equally. Web label noise is at instance-level in which certain images are more likely to be mislabelled than others. For example,  ``Honda Civic'' images taken from the side view are more likely to be confused with ``Honda Accord'' as the two models are lookalike from the side view. Such confusion is rare for car images taken from the front view. Third, images with noise from the web come from an open vocabulary outside the class vocabulary of Mini-ImageNet or Stanford Cars. For example, the noisy images of ``ladybug'' include ``fly'' and other bugs that do not belong to any of the classes in Mini-ImageNet.

\begin{figure}[ht]
\centering
\begin{subfigure}[b]{0.48\textwidth}
    \includegraphics[width=\textwidth]{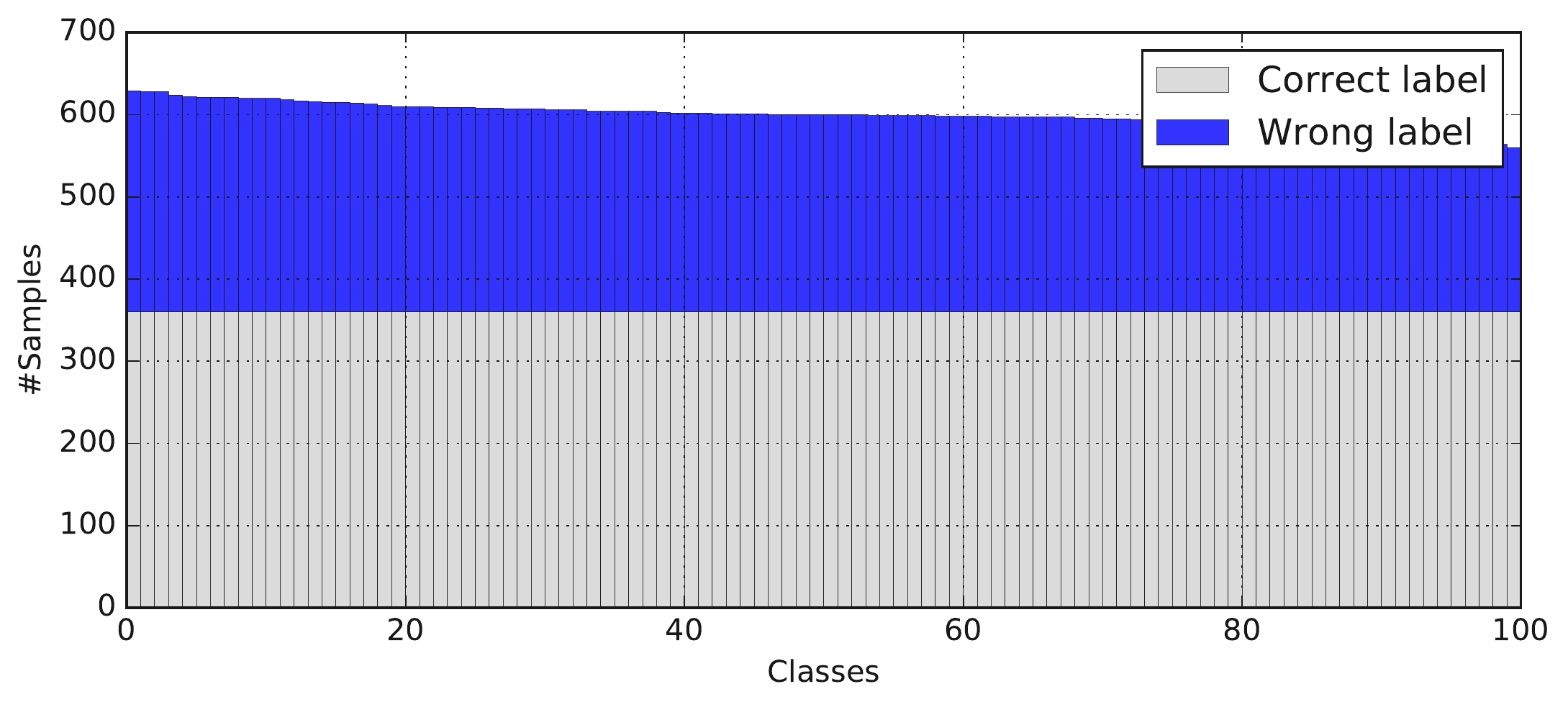}
    \caption{Blue Mini-ImageNet 40\% Noise}
\end{subfigure}
\begin{subfigure}[b]{0.48\textwidth}
    \includegraphics[width=\textwidth]{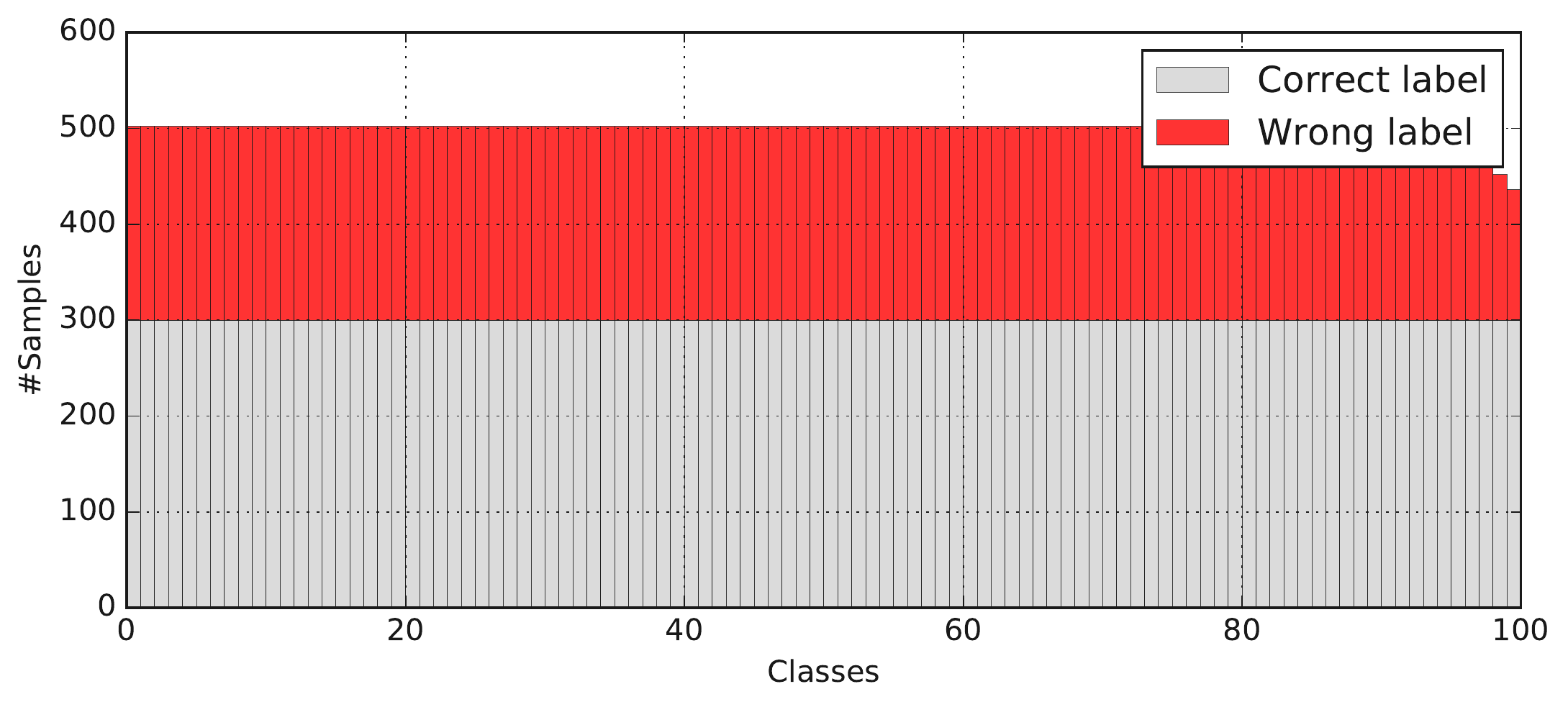}
    \caption{Red Mini-ImageNet 40\% Noise}
\end{subfigure}
\begin{subfigure}[b]{0.48\textwidth}
    \includegraphics[width=\textwidth]{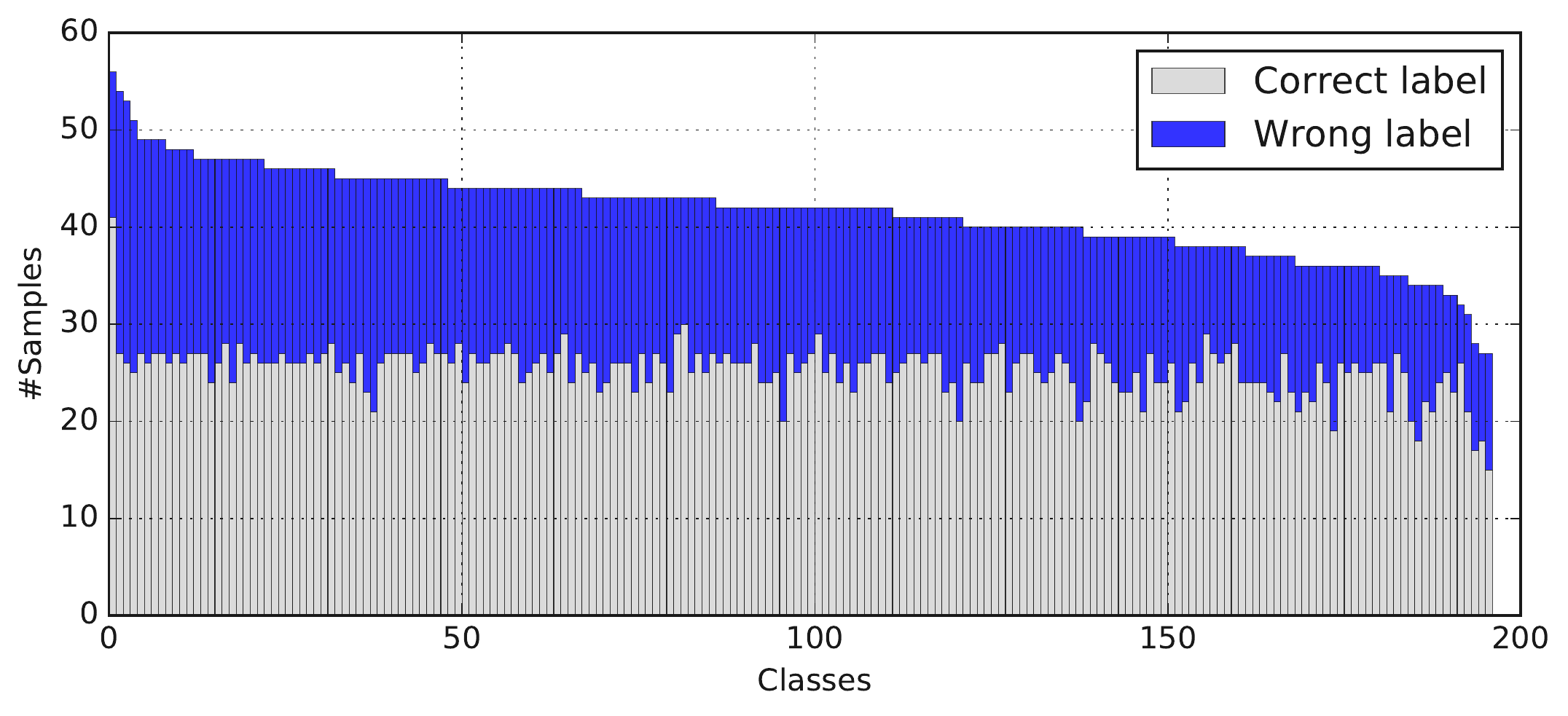}
    \caption{Blue Stanford Cars 20\% Noise}
\end{subfigure}
\begin{subfigure}[b]{0.48\textwidth}
    \includegraphics[width=\textwidth]{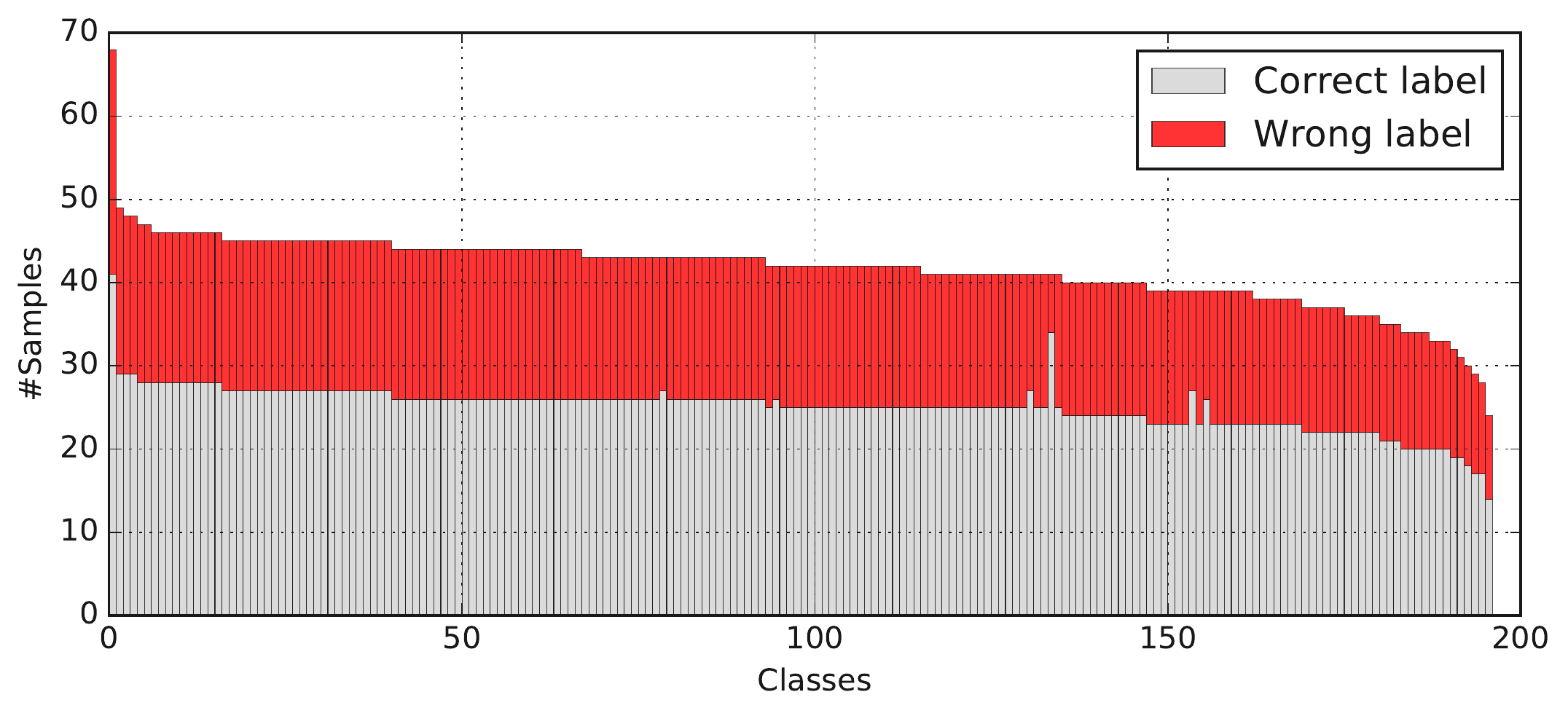}
    \caption{Red Stanford Cars 20\% Noise}
\end{subfigure}
\caption{\label{fig:dataset_dist}The distribution of images with correct and incorrect labels in Mini-ImageNet and Stanford Cars. The grey, blue, and red bar represent images of correct labels, symmetric noisy labels, and web noisy labels, respectively. Classes are ranked by the number of training examples. Better viewed in color.}
\end{figure}

Fig.~\ref{fig:dataset_dist} illustrates the distribution of correctly-labeled and incorrectly-labeled images across classes, where symmetric label noise is in blue and web label noise is in red. It is worth noting that it is not a feasible option for us to annotate existing datasets of web labels, \eg~WebVision~\citep{li2017webvision} or Clothing-1M~\citep{xiao2015learning}. Due to their imbalanced class distribution, for many classes, we simply cannot find sufficient images with incorrect labels to label in these datasets.

As incorrect images are rare in a few common classes (\eg~hotdog), we need to limit the size of Red Mini-ImageNet to 50K such that every class can get sufficient incorrect images at every noise level. Recall the role of the blue noisy datasets is to confirm existing findings on symmetric noise. Initially, we made the Blue and Red Mini-ImageNet to be the same size of 50K examples. However, we found that existing findings on symmetric noise hold on both the full (60K) and subset (50K) of Blue Mini-ImageNet. In the end, we decided to report the results on the 60K full set which results in a larger size of Blue Mini-ImageNet. This design would not affect our main contributions for the following reasons. First, on our second dataset Stanford Cars, the blue and red set have the same size. Second, our method has been verified by extensive experiments on many other datasets. Third, our main findings are on red noise which may not be affected by the size of the blue set.

\section{Detailed Experimental Setups}

This section presents detailed setups for training and testing used in our experiments.

\subsection{Setup on the proposed dataset}
\textbf{Network architectures}. Table~\ref{tab:backbone} lists the parameter count and input image size for each network architecture used in our experiments. We obtained their model checkpoints trained on the ImageNet 2012 dataset from TensorFlow Slim\footnote{https://github.com/tensorflow/models/tree/master/research/slim}, EfficienNet TPU\footnote{https://github.com/tensorflow/tpu/tree/master/models/official/efficientnet}, and from~\citet{kornblith2019better}. The last two columns list the top-1 accuracy of our obtained models along with the accuracy reported in the original paper. As shown, the top-1 accuracy of these architectures on the ImageNet ILSVRC 2012 validation ranges from 71.6\% to 83.6\%. We select the above architectures to be representative of diverse capacities.

\begin{table}[ht]
\centering
\caption{Overview of the ImageNet architectures used in our study.}
\label{tab:backbone}
\begin{tabular}{@{}lrccc@{}}
\toprule
\multirow{2}{*}{Network} & \multirow{2}{*}{Parameters} & \multirow{2}{*}{Image Size} & \multicolumn{2}{c}{ImageNet Top-1 Acc.} \\
 &  &  & Paper& \multicolumn{1}{c}{Our checkpoint} \\ 
\midrule
EfficientNet B5~\citep{tan2019efficientnet}&28.3M&456&83.3&83.3\\
Inception V2~\citep{ioffe2015batch}&10.2M&224&74.8&73.9\\
Inception V3~\citep{szegedy2016rethinking}&21.8M&299&78.8&78.6\\
Inception-ResNet V2~\citep{szegedy2017inception} &54.2M&299&80.0&80.3\\
MobileNet V2~\citep{sandler2018mobilenetv2}&2.2M&224&72.0&71.6\\
ResNet 50 V1~\citep{he2016deep}&23.5M&224&75.2&75.9\\
ResNet 101 V1~\citep{he2016deep}&42.5M&224&76.4&77.5\\
\bottomrule
\end{tabular}
\end{table}

\textbf{Training from scratch (random initialization)}. For vanilla training, we trained each architecture on the clean dataset (0\% noise level) to find the optimal training setting by grid search. Our grid search consisted of 6 start learning rates of $\{1.6, 0.16, 1.0, 0.5, 0.1, 0.01\}$ and 3 learning rate decay epochs of $\{1, 2, 3\}$. The exponential learning rate decay factor was fixed to 0.975. We trained the network to full convergence, and the maximum epoch to train was 200 on Mini-ImageNet (Red and Blue) and 300 epochs on Stanford Cars (Blue and Red), where the learning rate warmup~\citep{goyal2017accurate} was used in the first 5 epochs. The training was using Nesterov momentum with a momentum parameter of 0.9 with a batch size of 64, taking an exponential moving average of the weights with a decay factor of 0.9999. We had to reduce the batch size to 8 for EfficientNet for its larger image input. Following~\cite{kornblith2019better}, our vanilla training was with batch normalization layers but without label smoothing, dropout, or auxiliary heads. We employed the standard prepossessing in EfficientNet\footnote{https://github.com/tensorflow/tpu/blob/master/models/official/efficientnet/preprocessing.py} for data augmentation and evaluated on the central cropped images on the validation set. Training in this way, we obtained reasonable performance on the clean Stanford Cars validation set. For example, our Inception-ResNet-V2 got 90.8 (without dropout) and 92.4 (with dropout) versus 89.9 reported in~\citep{kornblith2019better}.

\textbf{Fine-tuning from ImageNet checkpoints}. For fine-tuning experiments, we initialized networks with ImageNet-pretrained weights. We used a similar training protocol for fine-tuning as training from scratch. The start learning rate was stable in fine-tuning so we fixed it to 0.01 and only searched the learning rate decay epochs in $\{1, 3, 5, 10\}$. Learning rate warmup was not used in fine-tuning. As fine-tuning converges faster, we scaled down the maximum number of epochs to train by a factor of 2 and trained the network to full convergence. Training in this way, we obtained reasonable performance on the clean Stanford Cars test set. For example, our Inception-ResNet-V2 got 92.4 versus 92.0 reported in~\citep{kornblith2019better} and our EfficientNet-B5 got 93.8\% versus 93.6\% reported in~\citep{tan2019efficientnet}. 

\textbf{Baseline comparison}. For method comparison, we used Inception-ResNet as the default network. All methods employed the identical setting discussed above, including the same start learning rate, learning rate decay factor, batch size, and the maximum number of epochs to train. For Dropout, as it converges slower, we added another 100 epochs to its \#maximum epochs to train. We extensively searched the hyperparameter for each method on every noise level using the hyperparameter range discussed in the main manuscript. The performance variance under different hyperparameters can be found in Fig.~\ref{fig:comp_imagenet_sc}, Fig.~\ref{fig:comp_imagenet_ft}, Fig.~\ref{fig:comp_cars_sc}, and Fig.~\ref{fig:comp_cars_ft}, where the black line shows the 95\% confidence interval of the accuracy under all searched hyperparameters.

\subsection{Setup on public benchmarks: CIFAR and WebVsion}
CIFAR-10 and CIFAR-100~\cite{krizhevsky2009learning} consist of 32 $\times$ 32 color images arranged in 10 and 100 classes. Both datasets contain 50,000 training and 10,000 validation images. The ResNet-32~\cite{he2016deep} with standard data augmentation function was used as our network backbone. In training, the batch size was set to 128 and we trained 400K iterations for the ResNet model by a distributed asynchronized momentum SGD optimizer (momentum = 0.9) on 8 V100 GPUs. We used the common learning rate scheduling strategy: setting the starting learning rate as 0.1 and using the step-wise exponential learning rate decay which multiplies it by 0.9 every 20K iterations. In this way, training ResNet-32 on the clean training dataset, the validation accuracy reaches 95.2\% on CIFAR-10, and 78.0\% on CIFAR-100. We searched the hyperparameters of MentorMix in the following range: $\alpha = \{2,4,8,32\}$ and $\gamma_p=\{0.8,0.7,0.6,0.5,0.4,0.2,0.1\}$. We used a simple MentorNet called ``Predefined MentorNet'' from~\cite{jiang2018mentornet}\footnote{https://github.com/google/mentornet} which, as discussed in the main paper, computes the weight by a thresholding function $v_i^* = \bm{1}(\ell(\bx_{i}, y_i) \le \gamma)$.

WebVision 1.0~\citep{li2017webvision} contains 2.4 million images of real-world noisy labels, crawled from the web using the 1,000 concepts in ImageNet ILSVRC12. We downloaded the resized images from the official website\footnote{https://www.vision.ee.ethz.ch/webvision/download.html}. The inception-resnet v2~\cite{szegedy2017inception} was used as our network backbone. In training, the batch size was set to 64 and we trained 4M iterations using a distributed asynchronized momentum optimizer on 32 V100 GPUs. The start learning rate was 0.01 and was discounted by a factor of 0.95 every 562K steps. The weight decay was $4e{-5}$. The batch norm was used and its decay was set to 0.9997 and the epsilon was 0.001. The default data augmentation for the ResNet model is used. We also tested our method on the WebVision mini-training set that contains about 61K Google images on the first 50 classes. All the models were evaluated on the clean ILSVRC12 and WebVision validation set. The best hyperparameter is $\gamma_p=0.7$ and $\alpha=0.4$ for both the WebVision full training set and the WebVision mini-training set.

\section{Alternative Approaches for Dataset Construction}
In this section, we study two alternative approaches to construct our red datasets. Our goal is to verify whether our findings in Section 5.3 of the main paper are consistent when the datasets are constructed differently. Please note that it may not be necessary to verify the proposed method MentorMix on these new datasets as it has already been verified in Section 5.2 on the two public benchmarks (CIFAR and WebVision) that contain both synthetic and real-world noisy labels. We consider the following settings to construct our red datasets:

\begin{itemize}
    \item {\bf Setting 0 (default)} uses the approach discussed in the main paper where we replace the clean images in Mini-ImageNet and Stanford Cars with incorrectly labeled images from the web while leaving the label unchanged. The advantage of our approach is that we closely match the construction of synthetic datasets while still being able to introduce controlled levels of noise that better resembles realistic label noise distributions.
    \item {\bf Setting 1 (web images only)}: in this setting, the red datasets only contain images from the web (with both correct and incorrect labels). No clean images in the original Mini-ImageNet or Stanford Cars datasets are used. This setting is used to understand the impact of the domain difference in Setting 0.
    \item {\bf Setting 2 (no image-to-image results)}: this setting is the same as setting 0 except only web images obtained from the text-to-image search are used. This setting examines the impact after removing the image-to-image search label noise.
\end{itemize}

First, we show that DNNs generalize much better on the red label noise under all three settings. We compare the standard deviation of the final accuracies at 10 noise levels (0\%-80\%) in Table~\ref{tab:std}. A higher standard deviation suggests a poorer generalization performance when DNNs are trained on noisy labels. Ideally, we expect to observe a significantly smaller standard deviation in web noise. Table~\ref{tab:std} shows the standard deviation of the red noise is at least two times less than that of the blue noise. The standard deviations of red noise are comparable among all three settings. These results show that DNNs generalize much better on red label noise despite specific approaches used to construct the dataset.

\begin{table}[ht]
\centering
\caption{Comparison of the standard deviation of the final accuracies across 10 noise levels. A higher standard deviation suggests a poorer generalization performance of DNNs trained on noisy labels.}
\label{tab:std}
\begin{tabular}{|c|c|c|c|c|c|c|c|c}
\hline
\multirow{2}{*}{Noise Settings} & \multicolumn{2}{c|}{Mini-ImageNet} & \multicolumn{2}{c|}{Stanford Cars} \\
 & Fine-tuned & Trained from scratch & Fine-tuned & Trained from scratch \\ 
 \hline
Blue Noise & 0.205 & 0.195 & 0.268& 0.347 \\
\hline
Red Noise (setting 0) & 0.051 & 0.088 &0.104 & 0.146 \\
Red Noise (setting 1) & 0.046 & 0.068 &0.104 & 0.127\\
Red Noise (setting 2) & 0.056 & 0.077 &0.067 & 0.096 \\
\hline
\end{tabular}
\end{table}

\begin{figure}[ht]
\centering
\begin{subfigure}[b]{0.33\textwidth}
    \includegraphics[width=\textwidth]{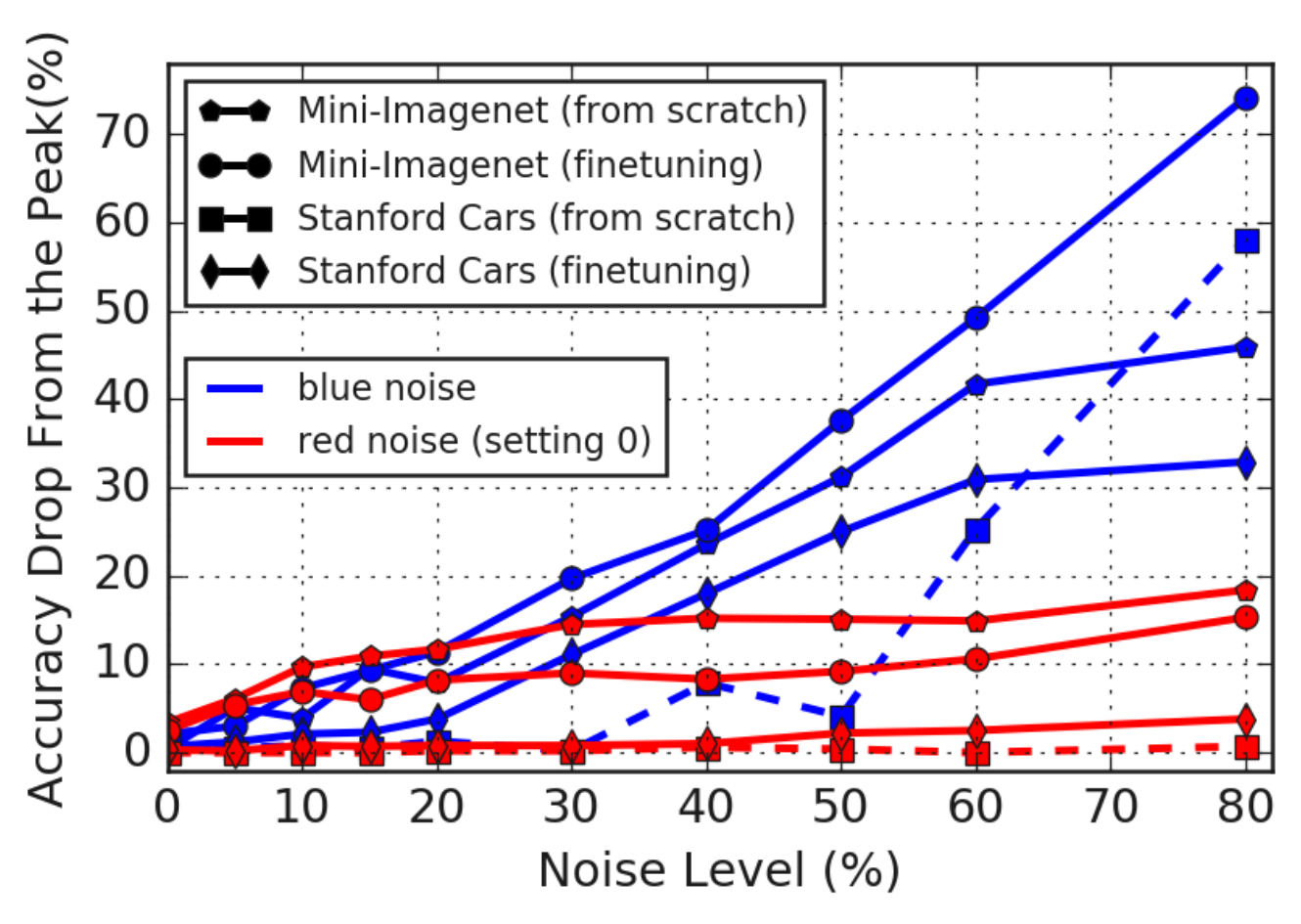}
    \caption{Red Noise (Setting 0)}
\end{subfigure}
\begin{subfigure}[b]{0.33\textwidth}
    \includegraphics[width=\textwidth]{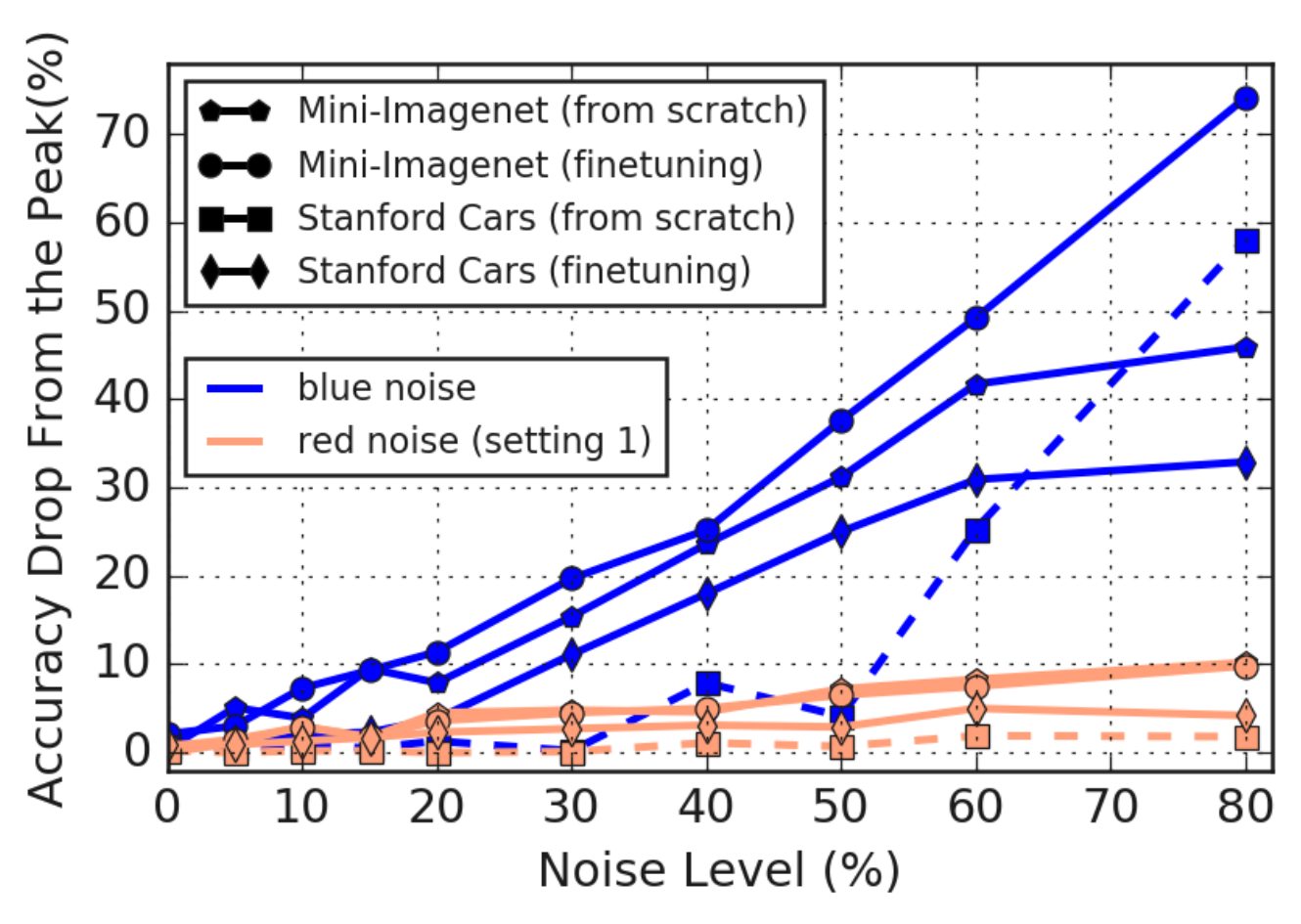}
    \caption{Red Noise (Setting 1)}
\end{subfigure}
\begin{subfigure}[b]{0.33\textwidth}
    \includegraphics[width=\textwidth]{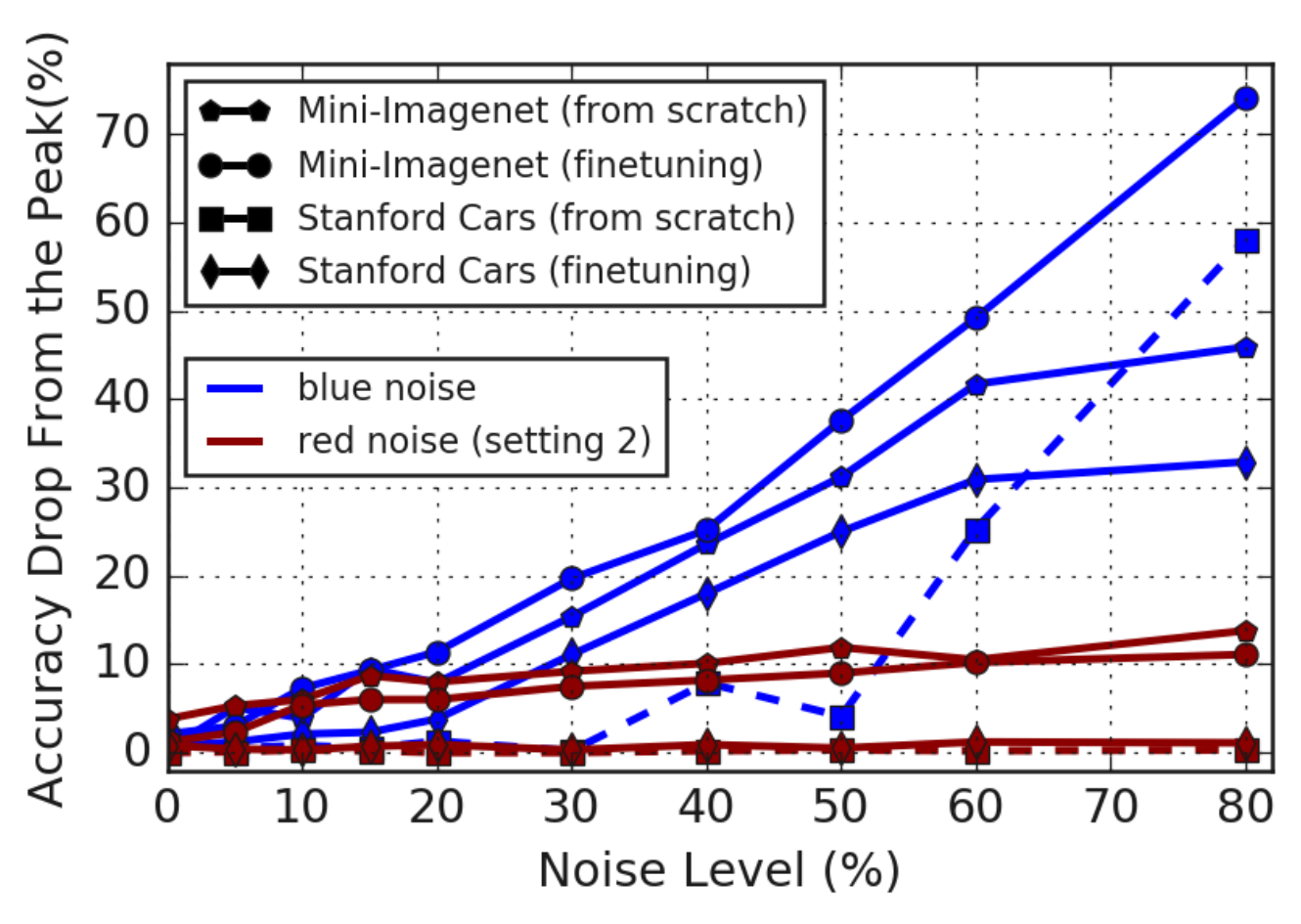}
    \caption{Red Noise (Setting 2)}
\end{subfigure}
\caption{\label{fig:drop_red_settings}Performance drop from the peak accuracy across 10 noise levels. Different colors are used to differentiate noise types. A larger drop ($y$-axis) means a better pattern is found during the early training stage}
\end{figure}

Second, we show that DNNs may not learn patterns first on red noise \ie~DNNs are able to automatically learn generalizable ``patterns'' in the early training stage before memorizing all noisy training labels. This is manifested by the gap between the peak and final validation accuracy. Fig.~\ref{fig:drop_red_settings} illustrates the relative difference, namely the drop, between the peak and final accuracy on the clean validation set. Recall a larger drop between the peak and final validation accuracy means a better pattern is found in the early training stage. As it shows, the drops of red noise under all three settings are significantly and consistently smaller than that of the blue noise. These results show the domain differences and the types of label noise (image-to-image search) do not distort this finding.

\begin{figure}[!ht]
\vspace{-3mm}
\centering
\includegraphics[width=0.9\textwidth]{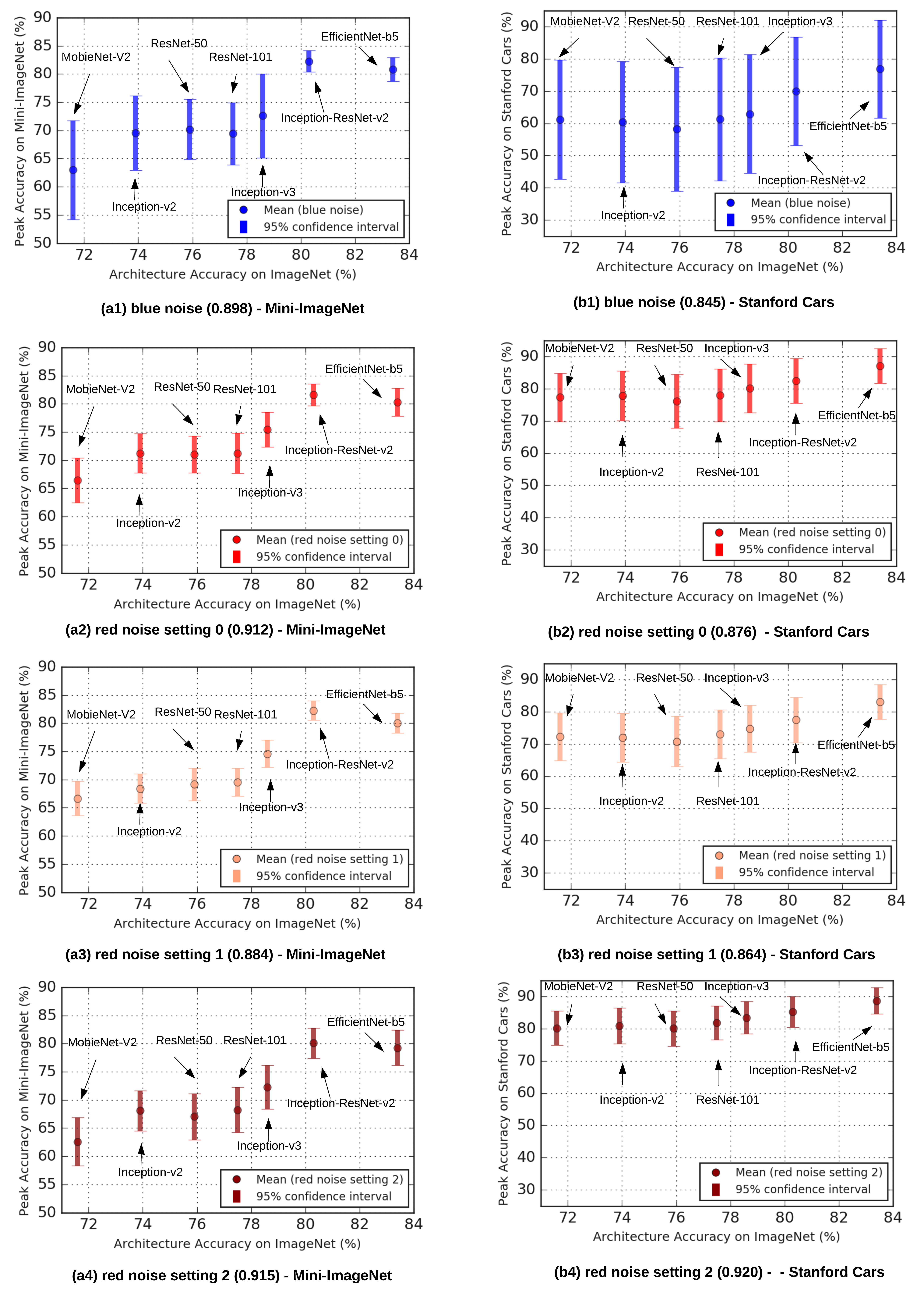}
\caption{\label{fig:backbones_all}Fine-tuning seven ImageNet architectures on the red and blue datasets. The number in parentheses is the Pearson correlation between the architecture's ImageNet accuracy and the performance on our red dataset. All three settings of red noise are illustrated. Better view in color.}
\end{figure}

Finally, we show that ImageNet architectures generalize on noisy labels when the networks are fine-tuned. To do so, we compute Pearson correlation $r$ for different red noise settings. The results are shown in Fig.~\ref{fig:backbones_all}, where the $x$-axis is the accuracy of the pretrained architectures on ImageNet and the $y$-axis shows the peak validation accuracy on noisy datasets. The bar plots the 95\% confidence interval across 10 noise levels, where the center dot marks the mean. As it shows, the correlation is consistent across all types of label noise where the Pearson correlations are shown in the parentheses. These results show a better pretrained architecture is likely to perform better when it is fine-tuned on noisy training labels. This finding seems to be consistent across all types of label noise.

\section{Detailed Method Comparison}
\begin{figure}[ht]
\vspace{-1mm}
\centering
\begin{subfigure}[b]{0.19\textwidth}
    \includegraphics[width=\textwidth]{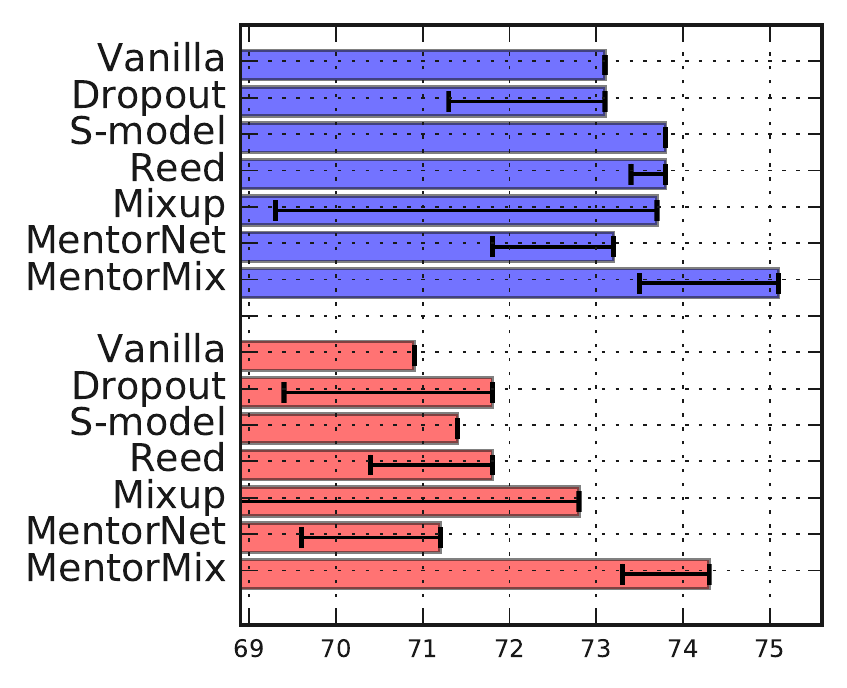}
    \caption{0\%}
\end{subfigure}
\begin{subfigure}[b]{0.19\textwidth}
    \includegraphics[width=\textwidth]{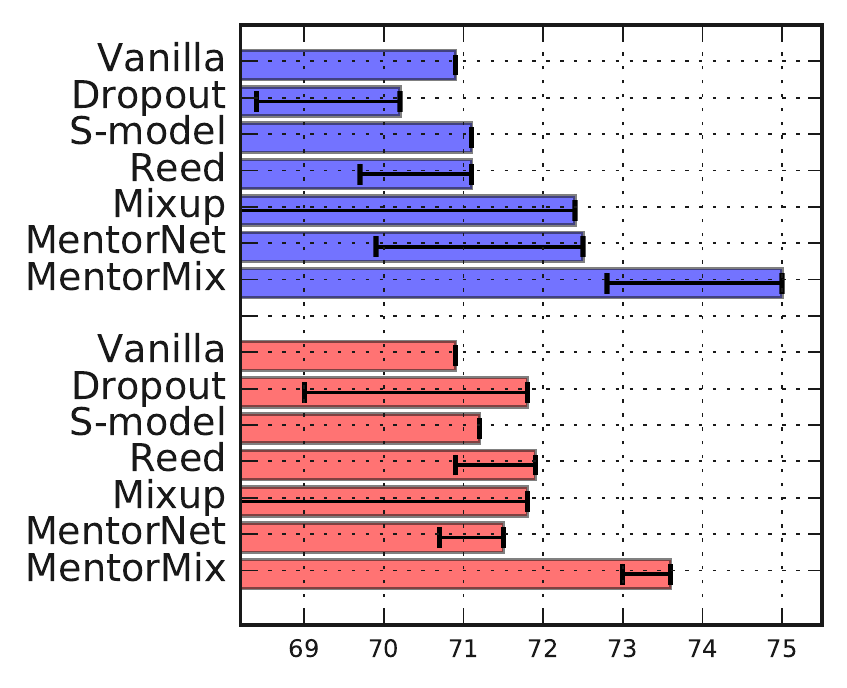}
    \caption{5\%}
\end{subfigure}
\begin{subfigure}[b]{0.19\textwidth}
    \includegraphics[width=\textwidth]{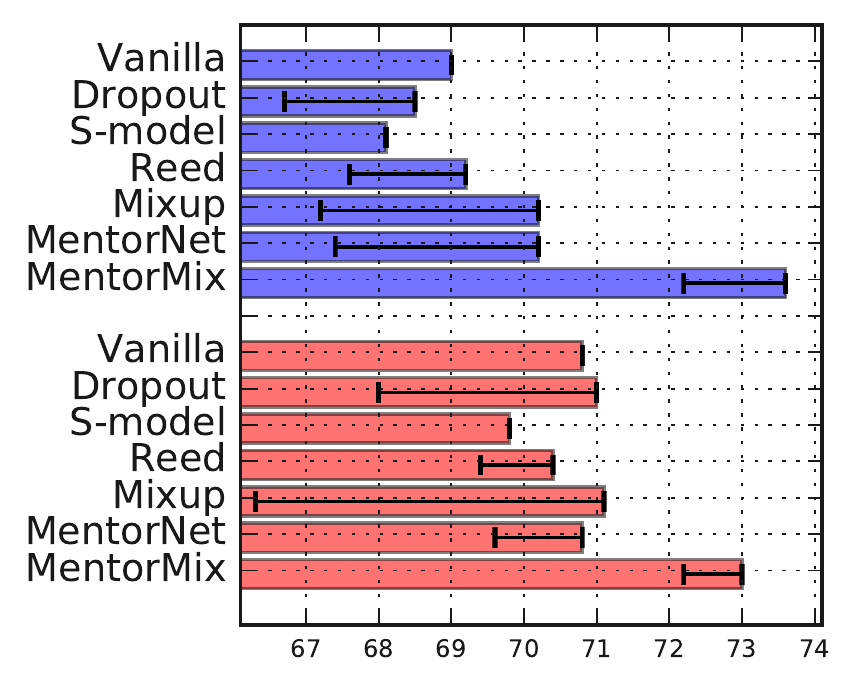}
    \caption{10\%}
\end{subfigure}
\begin{subfigure}[b]{0.19\textwidth}
    \includegraphics[width=\textwidth]{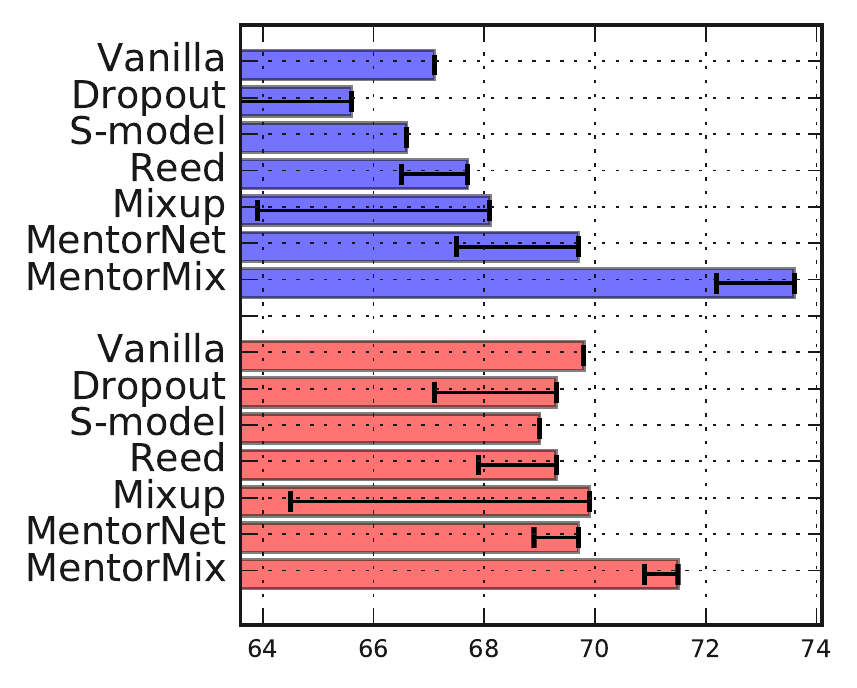}
    \caption{15\%}
\end{subfigure}
\begin{subfigure}[b]{0.19\textwidth}
    \includegraphics[width=\textwidth]{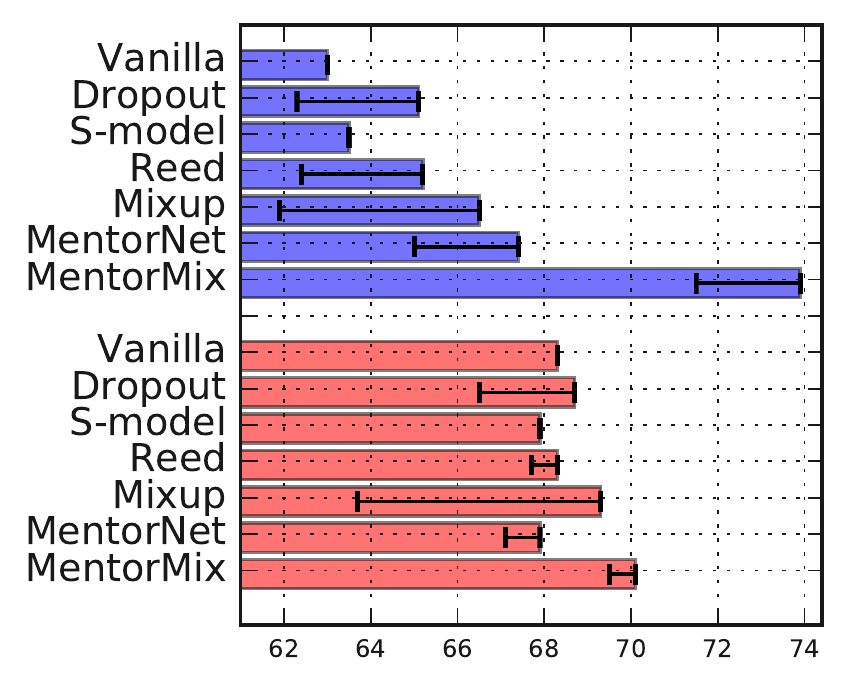}
    \caption{20\%}
\end{subfigure}
\begin{subfigure}[b]{0.19\textwidth}
    \includegraphics[width=\textwidth]{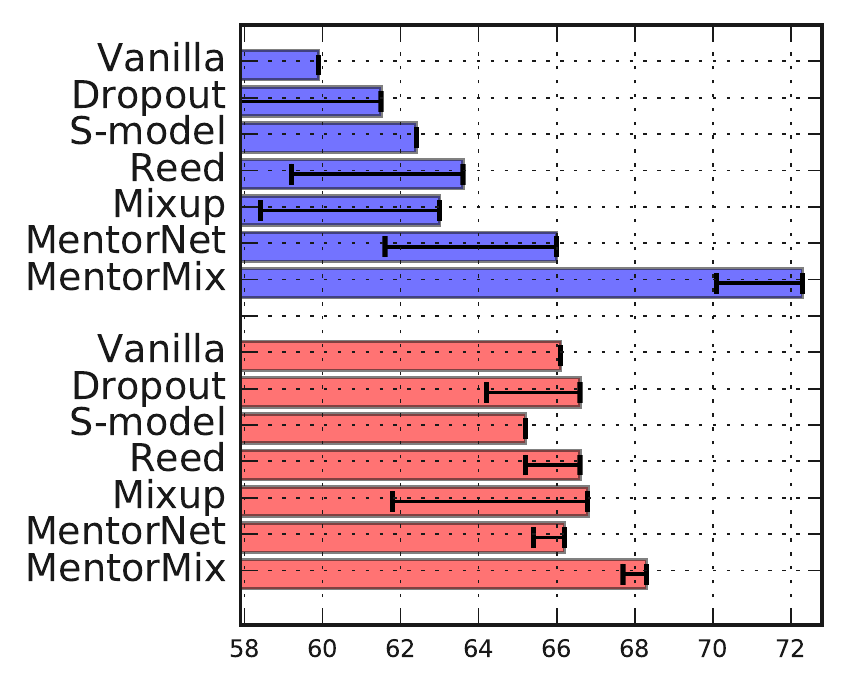}
    \caption{30\%}
\end{subfigure}
\begin{subfigure}[b]{0.19\textwidth}
    \includegraphics[width=\textwidth]{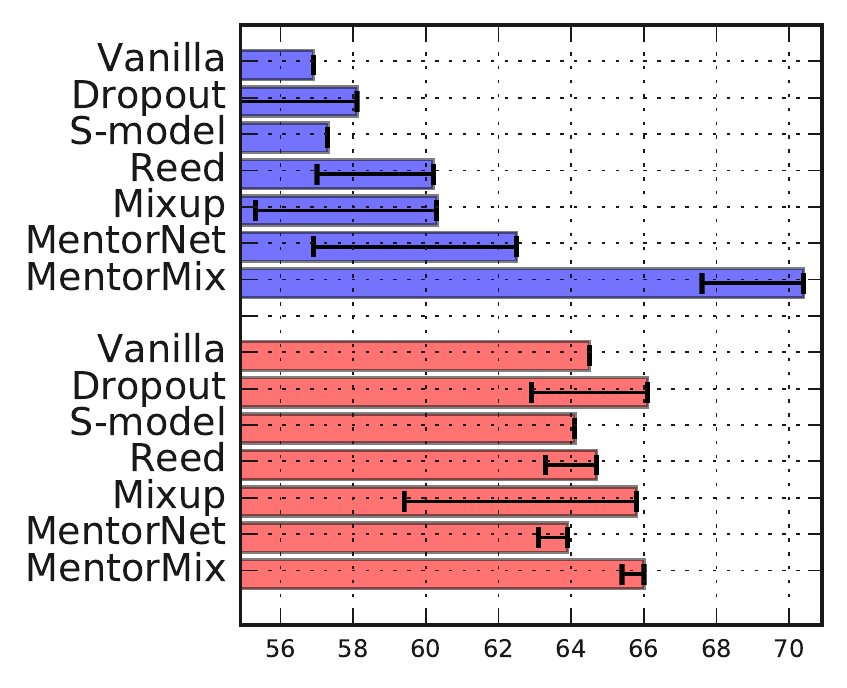}
    \caption{40\%}
\end{subfigure}
\begin{subfigure}[b]{0.19\textwidth}
    \includegraphics[width=\textwidth]{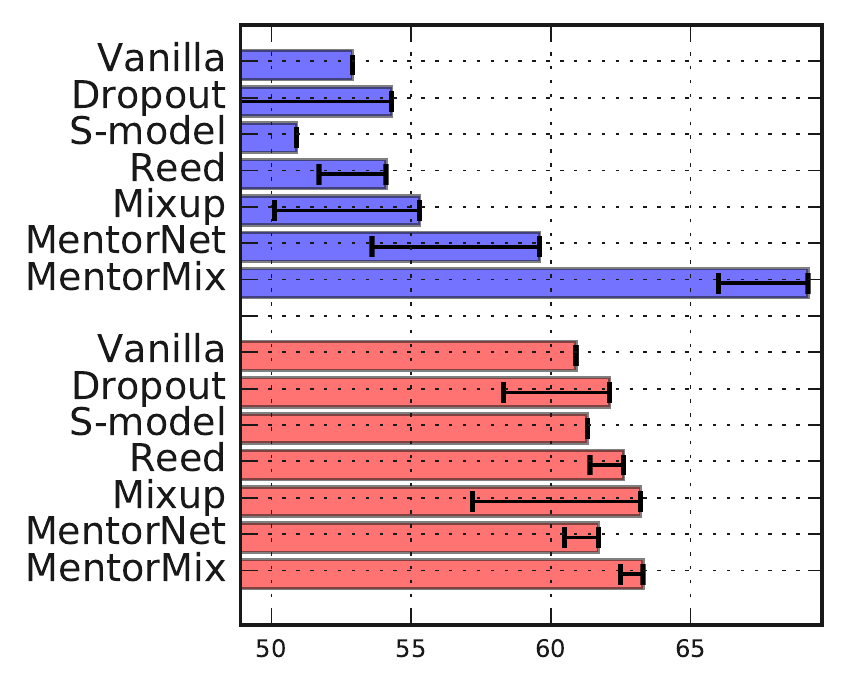}
    \caption{50\%}
\end{subfigure}
\begin{subfigure}[b]{0.19\textwidth}
    \includegraphics[width=\textwidth]{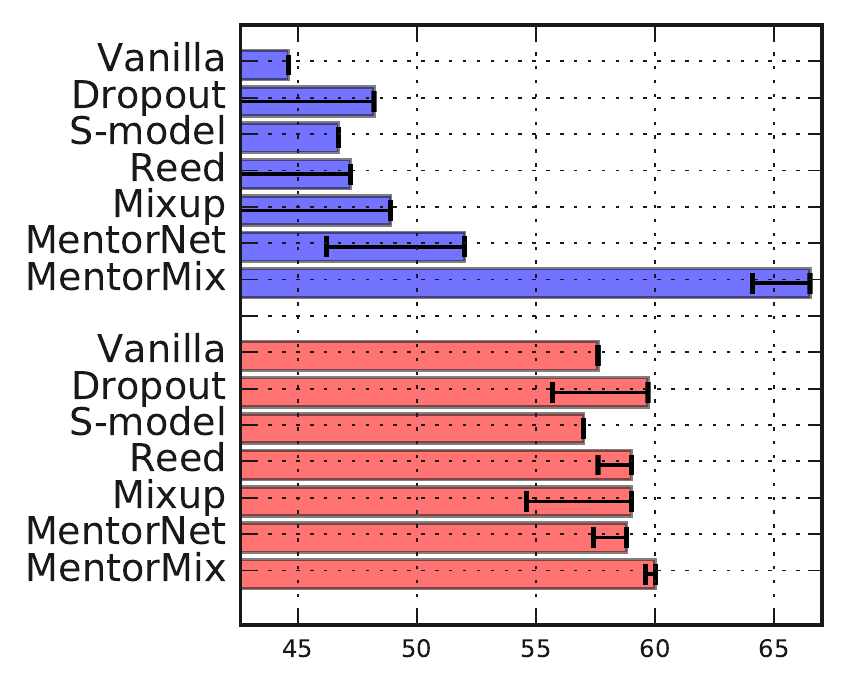}
    \caption{60\%}
\end{subfigure}
\begin{subfigure}[b]{0.19\textwidth}
    \includegraphics[width=\textwidth]{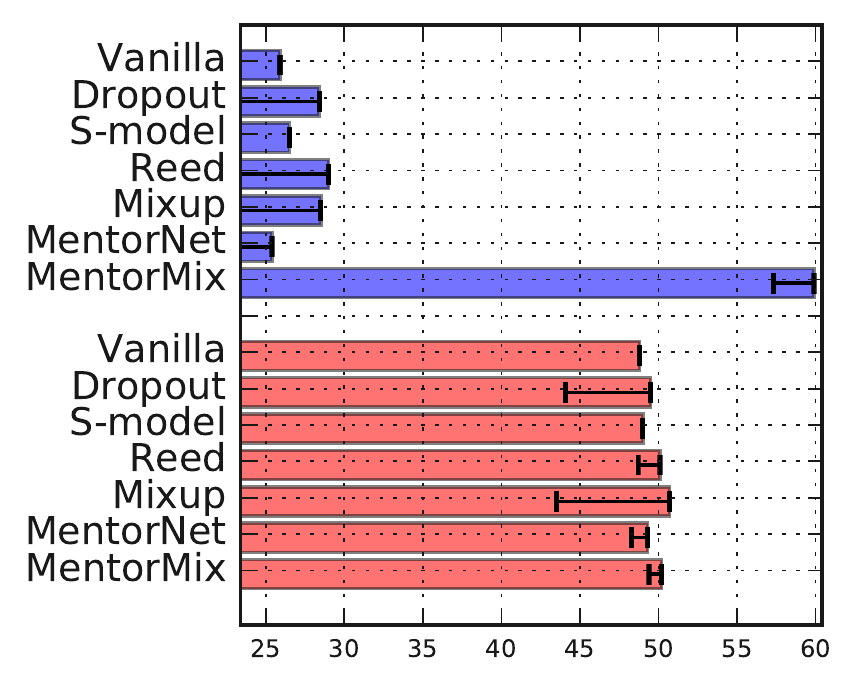}
    \caption{80\%}
\end{subfigure}
\vspace{-3mm}
\caption{\label{fig:comp_imagenet_sc}Peak accuracy of robust DNNs (trained from scratch) on Red and Blue Mini-ImageNet.}
\vspace{-2mm}
\end{figure}

\begin{figure}[!ht]
\vspace{-1mm}
\centering
\begin{subfigure}[b]{0.19\textwidth}
    \includegraphics[width=\textwidth]{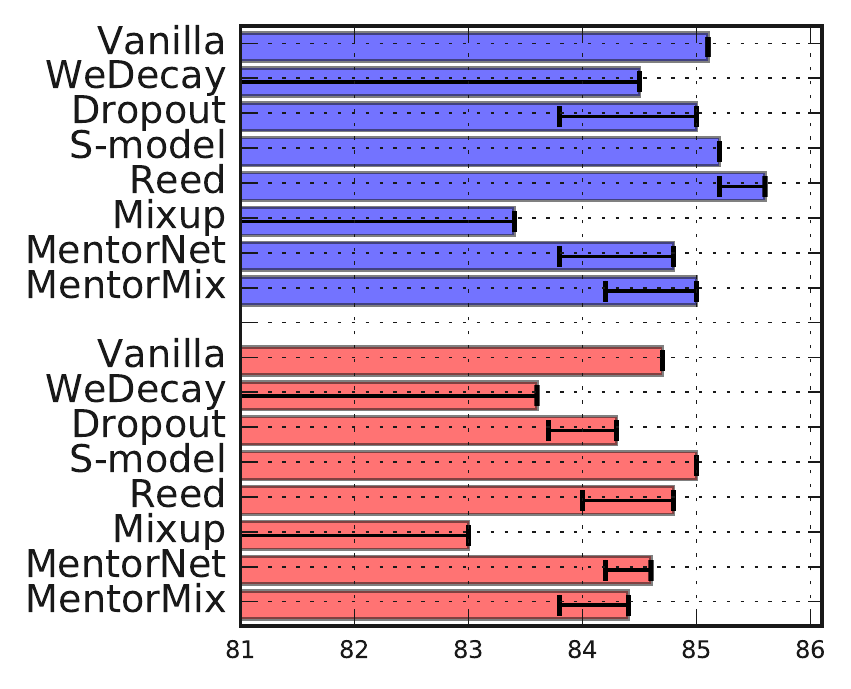}
    \caption{0\%}
\end{subfigure}
\begin{subfigure}[b]{0.19\textwidth}
    \includegraphics[width=\textwidth]{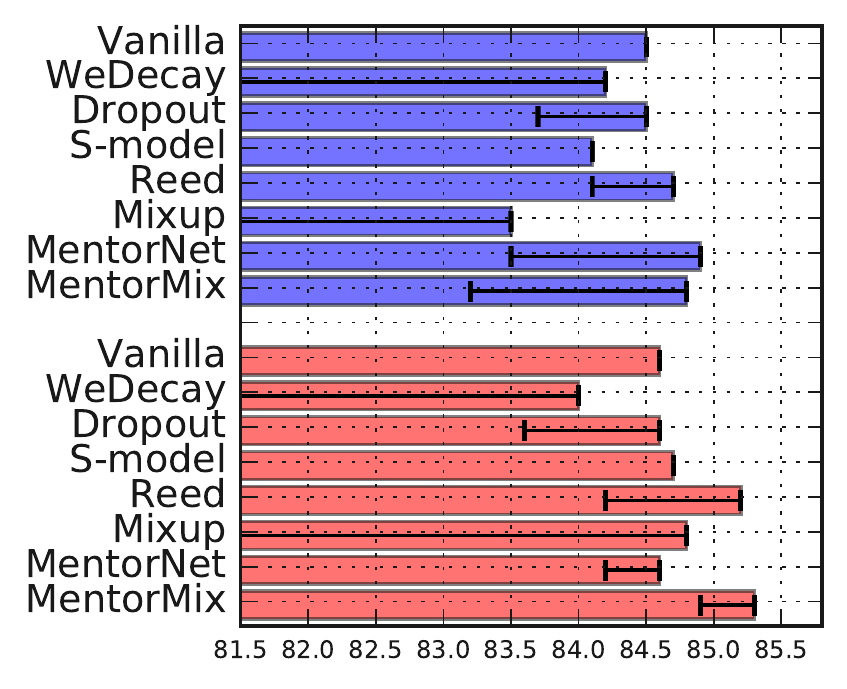}
    \caption{5\%}
\end{subfigure}
\begin{subfigure}[b]{0.19\textwidth}
    \includegraphics[width=\textwidth]{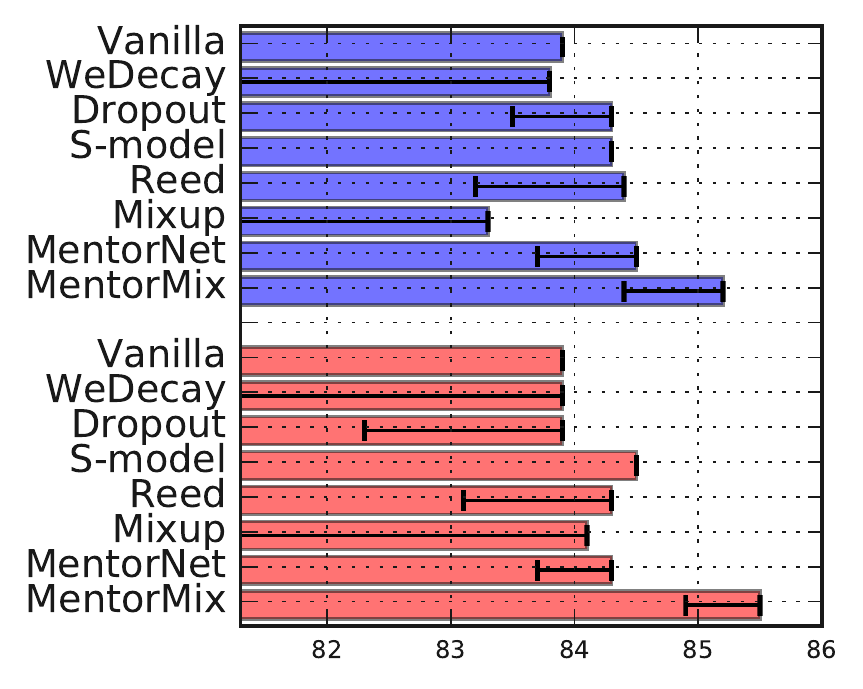}
    \caption{10\%}
\end{subfigure}
\begin{subfigure}[b]{0.19\textwidth}
    \includegraphics[width=\textwidth]{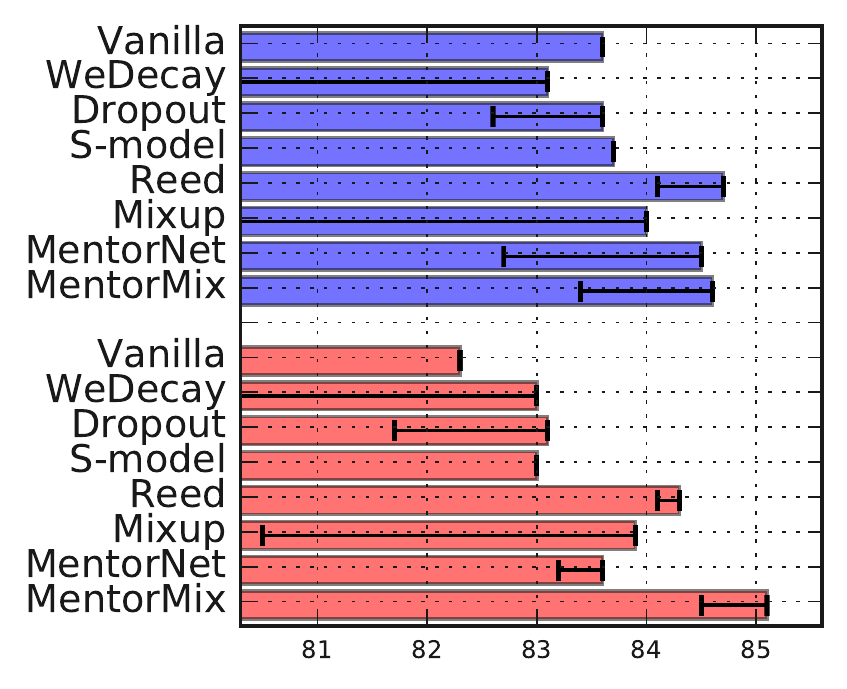}
    \caption{15\%}
\end{subfigure}
\begin{subfigure}[b]{0.19\textwidth}
    \includegraphics[width=\textwidth]{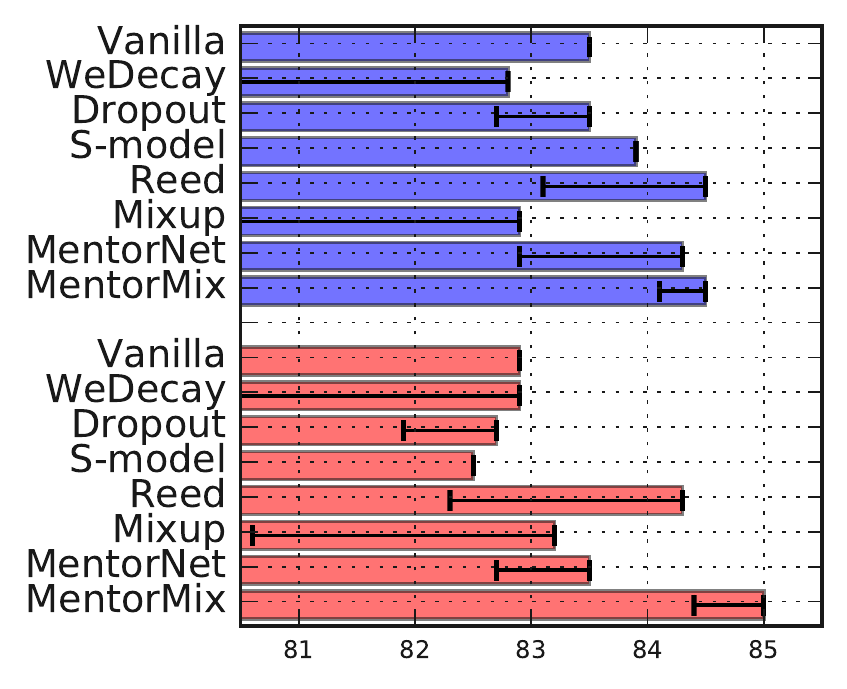}
    \caption{20\%}
\end{subfigure}
\begin{subfigure}[b]{0.19\textwidth}
    \includegraphics[width=\textwidth]{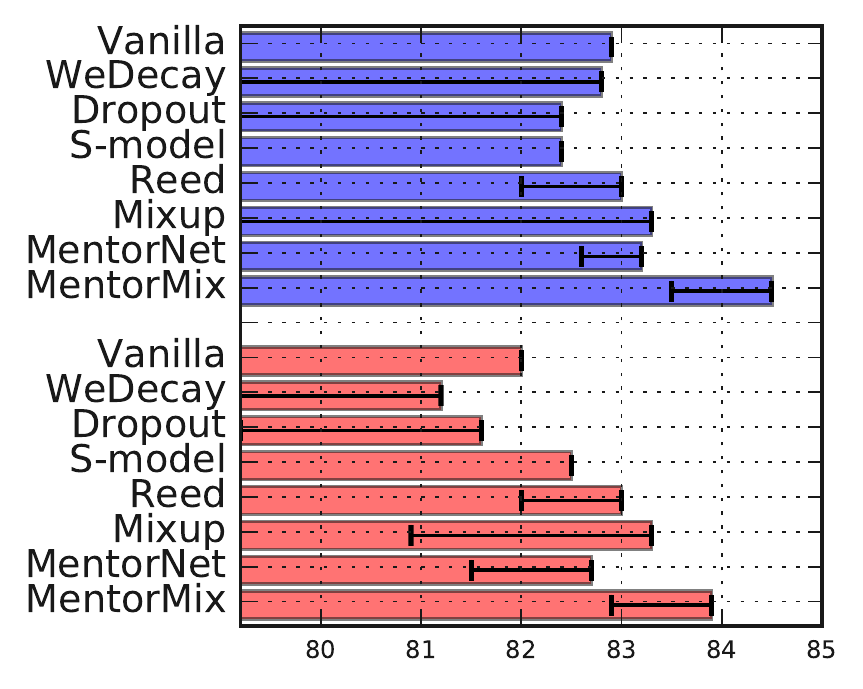}
    \caption{30\%}
\end{subfigure}
\begin{subfigure}[b]{0.19\textwidth}
    \includegraphics[width=\textwidth]{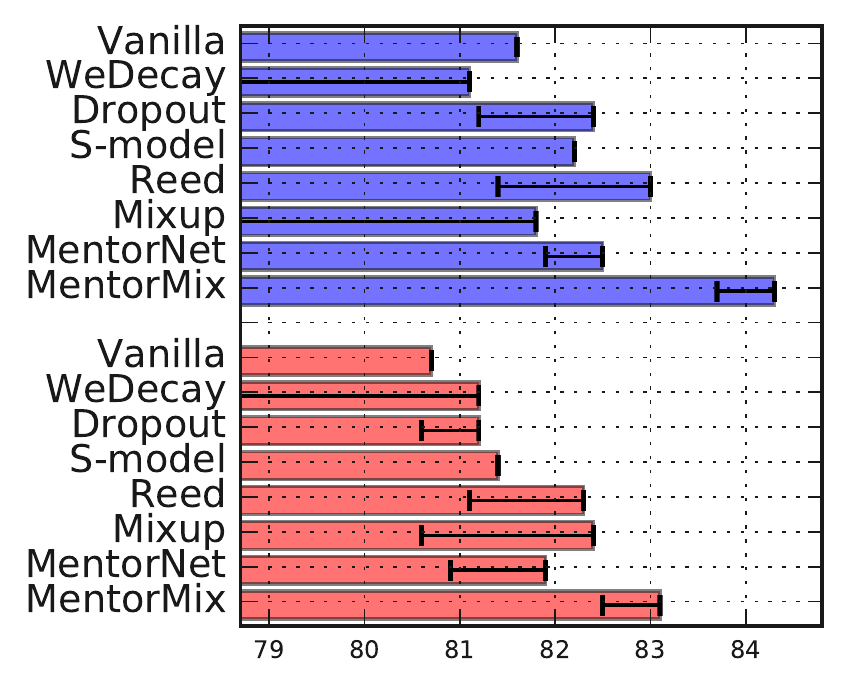}
    \caption{40\%}
\end{subfigure}
\begin{subfigure}[b]{0.19\textwidth}
    \includegraphics[width=\textwidth]{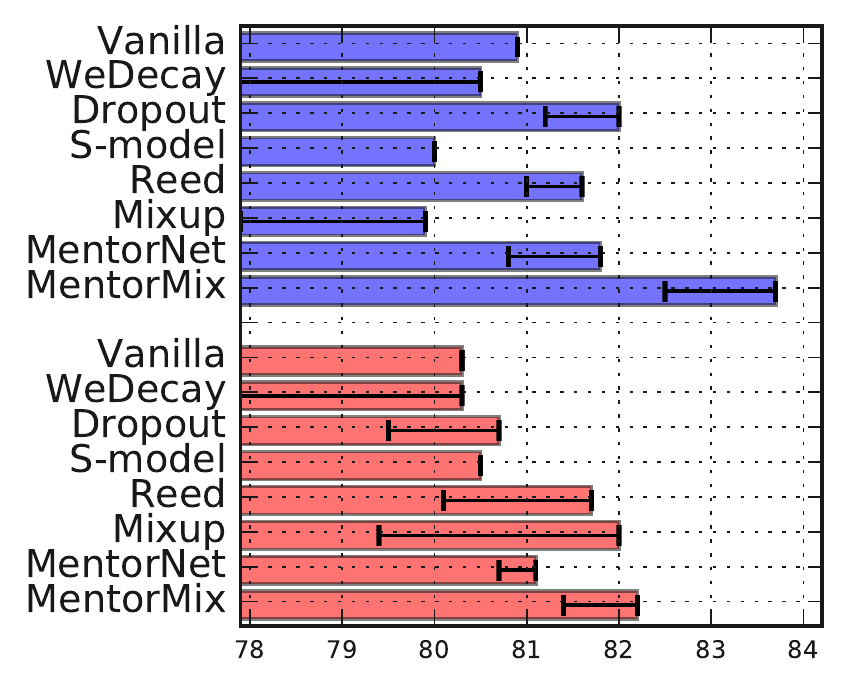}
    \caption{50\%}
\end{subfigure}
\begin{subfigure}[b]{0.19\textwidth}
    \includegraphics[width=\textwidth]{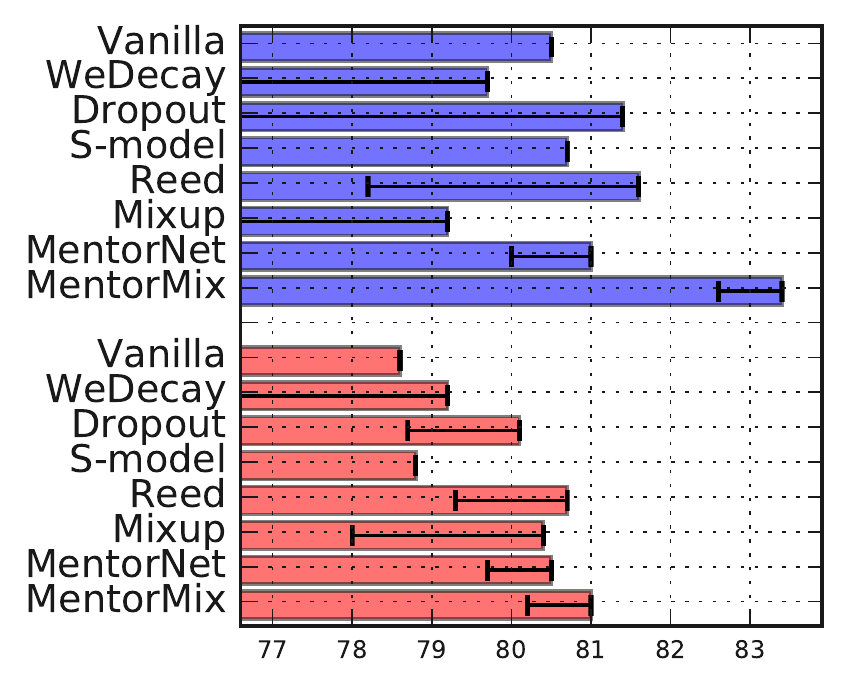}
    \caption{60\%}
\end{subfigure}
\begin{subfigure}[b]{0.19\textwidth}
    \includegraphics[width=\textwidth]{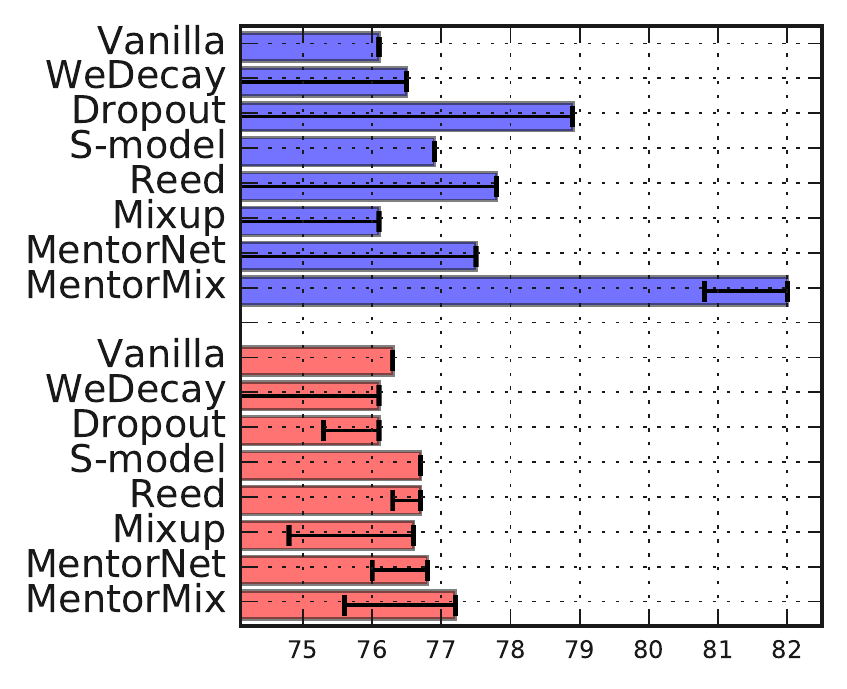}
    \caption{80\%}
\end{subfigure}
\vspace{-3mm}
\caption{\label{fig:comp_imagenet_ft}Peak accuracy of robust DNNs (fine-tuned) on Red and Blue Mini-ImageNet.}
\vspace{-2mm}
\end{figure}

\begin{figure}[!ht]
\centering
\begin{subfigure}[b]{0.19\textwidth}
    \includegraphics[width=\textwidth]{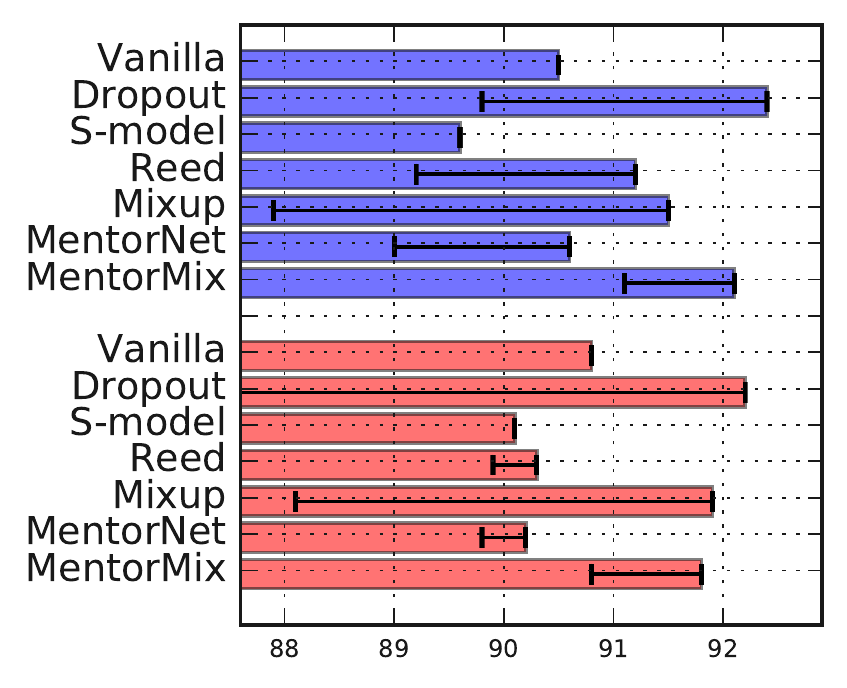}
    \caption{0\%}
\end{subfigure}
\begin{subfigure}[b]{0.19\textwidth}
    \includegraphics[width=\textwidth]{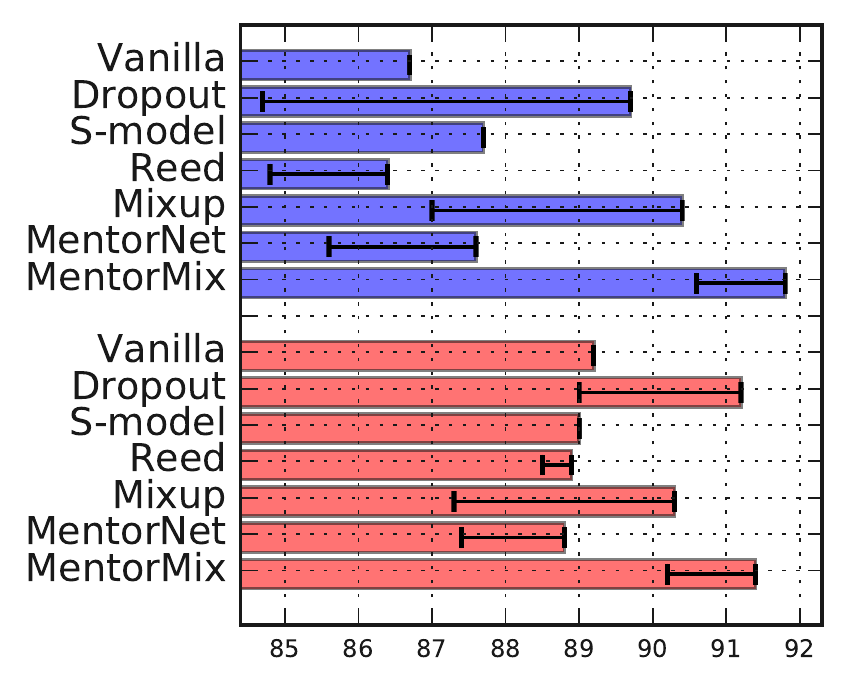}
    \caption{5\%}
\end{subfigure}
\begin{subfigure}[b]{0.19\textwidth}
    \includegraphics[width=\textwidth]{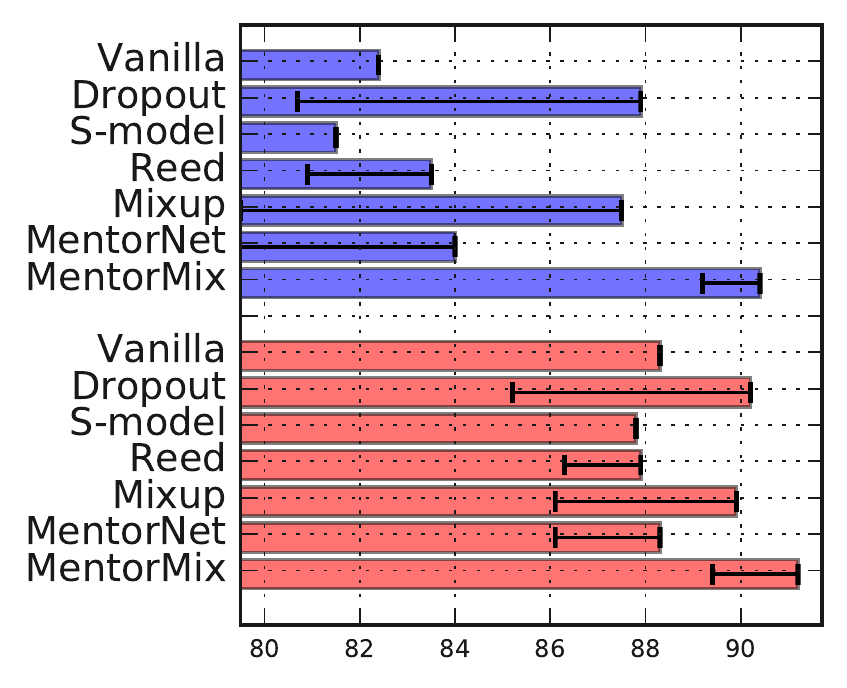}
    \caption{10\%}
\end{subfigure}
\begin{subfigure}[b]{0.19\textwidth}
    \includegraphics[width=\textwidth]{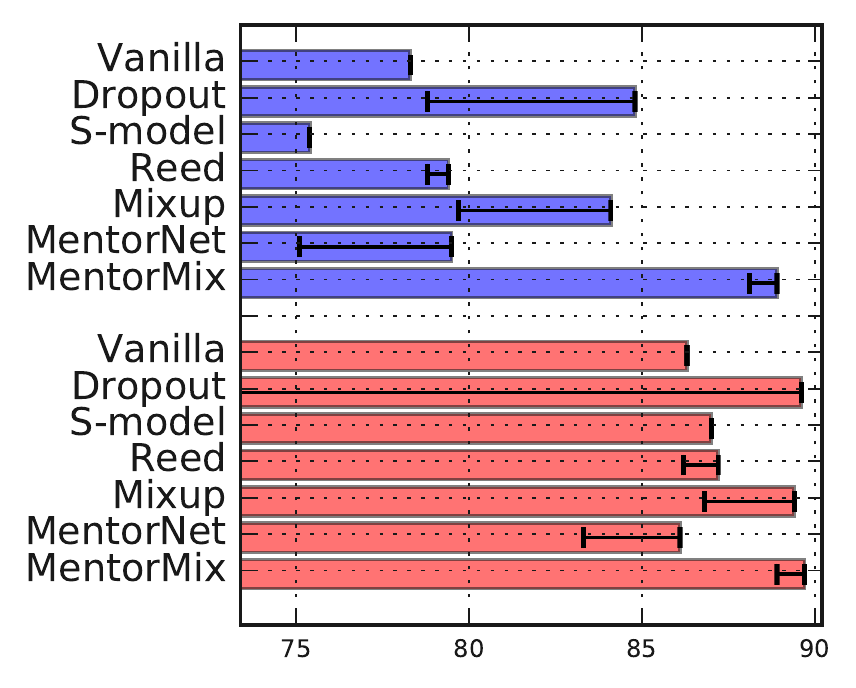}
    \caption{15\%}
\end{subfigure}
\begin{subfigure}[b]{0.19\textwidth}
    \includegraphics[width=\textwidth]{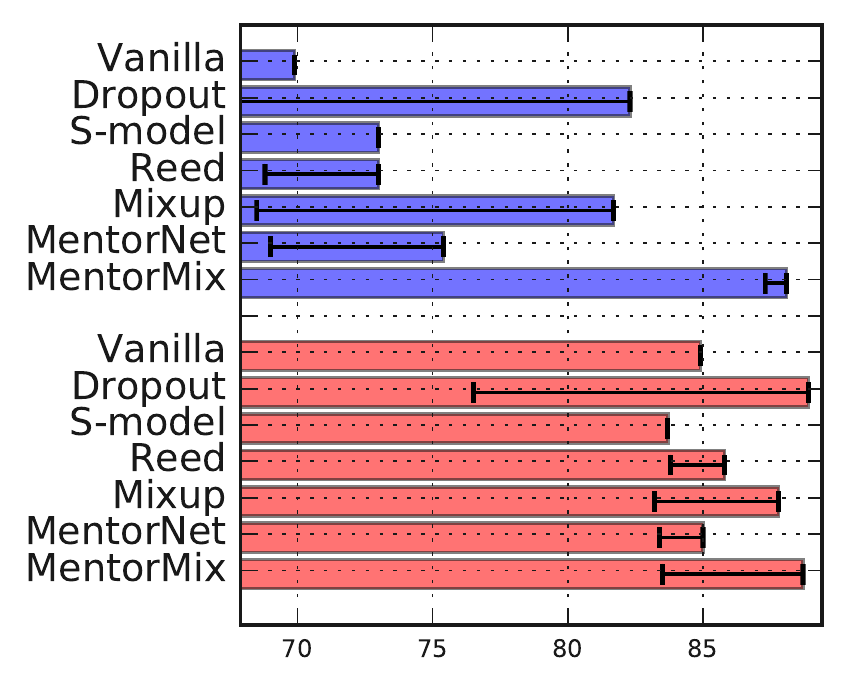}
    \caption{20\%}
\end{subfigure}
\begin{subfigure}[b]{0.19\textwidth}
    \includegraphics[width=\textwidth]{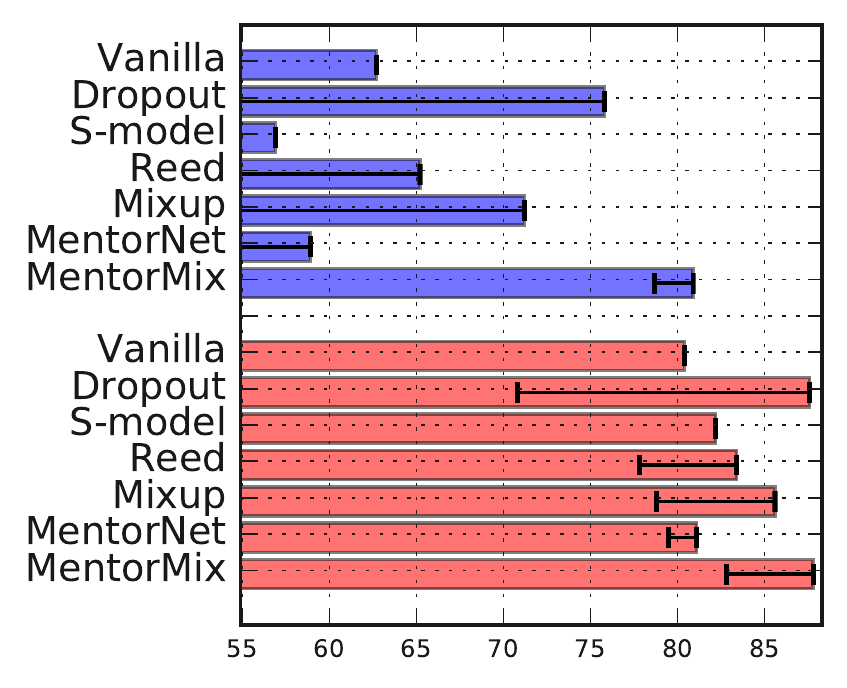}
    \caption{30\%}
\end{subfigure}
\begin{subfigure}[b]{0.19\textwidth}
    \includegraphics[width=\textwidth]{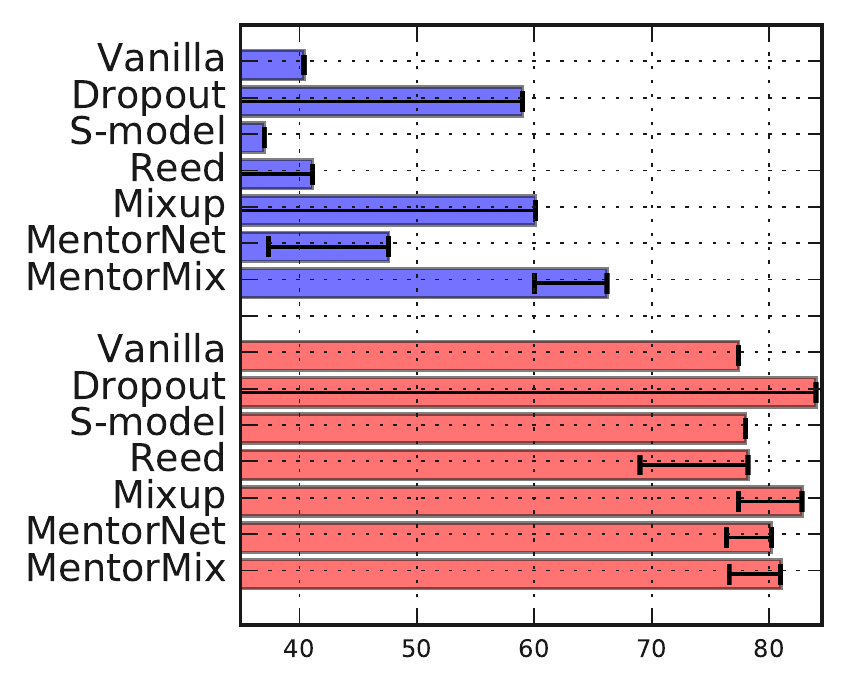}
    \caption{40\%}
\end{subfigure}
\begin{subfigure}[b]{0.19\textwidth}
    \includegraphics[width=\textwidth]{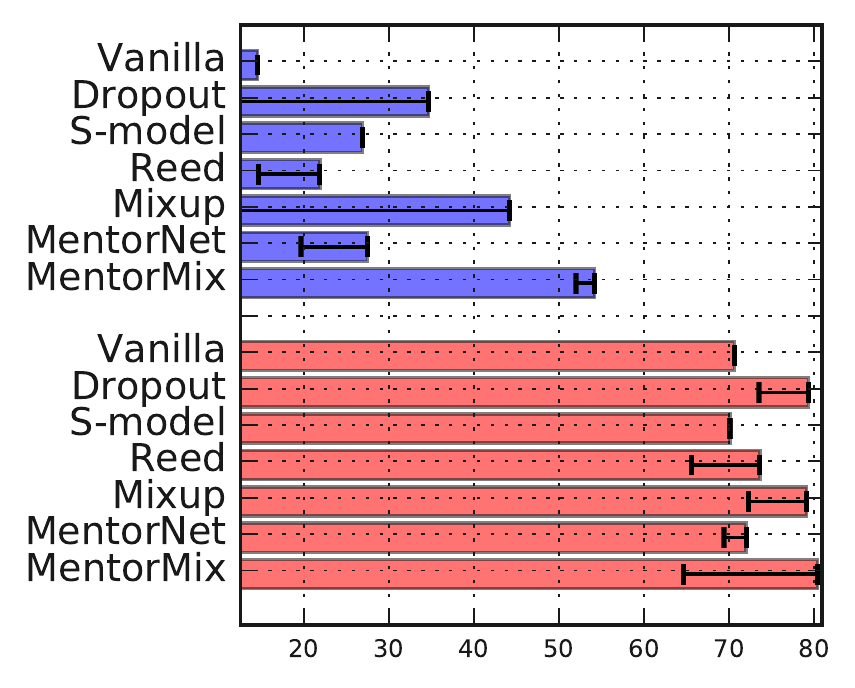}
    \caption{50\%}
\end{subfigure}
\begin{subfigure}[b]{0.19\textwidth}
    \includegraphics[width=\textwidth]{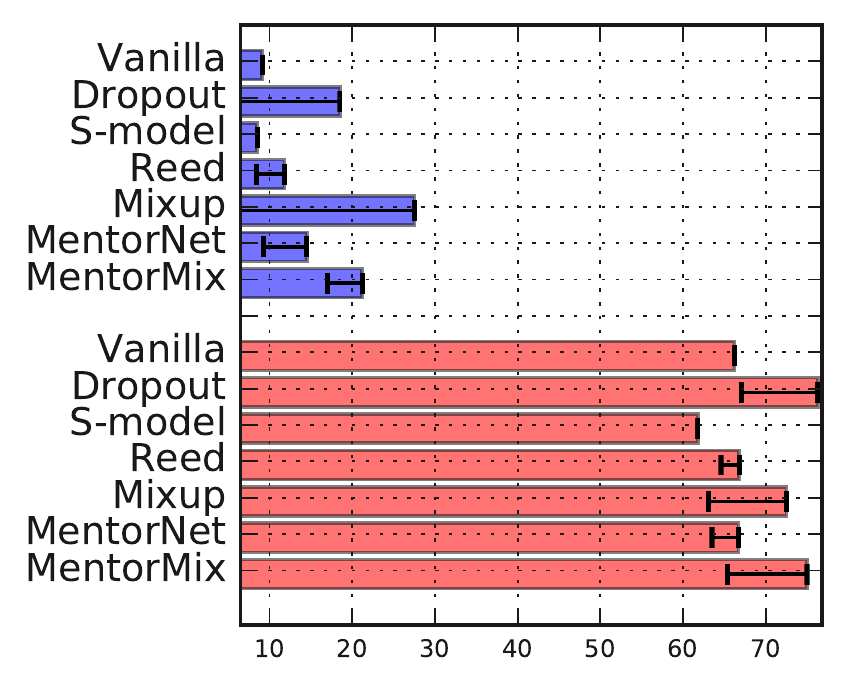}
    \caption{60\%}
\end{subfigure}
\begin{subfigure}[b]{0.19\textwidth}
    \includegraphics[width=\textwidth]{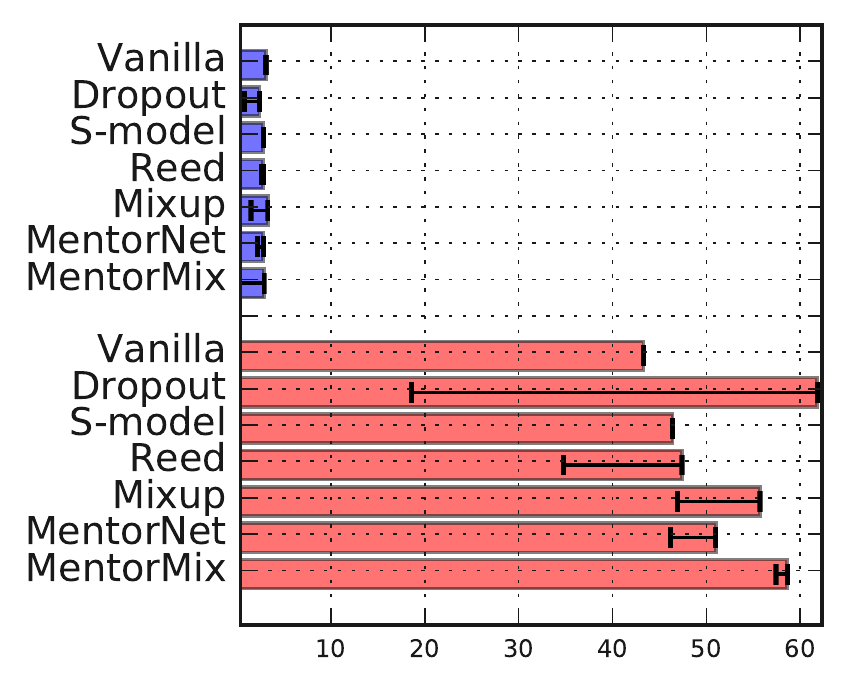}
    \caption{80\%}
\end{subfigure}
\vspace{-3mm}
\caption{\label{fig:comp_cars_sc}Peak accuracy of robust DNNs (trained from scratch) on Red and Blue Stanford Cars.}
\vspace{-3mm}
\end{figure}

\begin{figure}[!ht]
\vspace{-3mm}
\centering
\begin{subfigure}[b]{0.19\textwidth}
    \includegraphics[width=\textwidth]{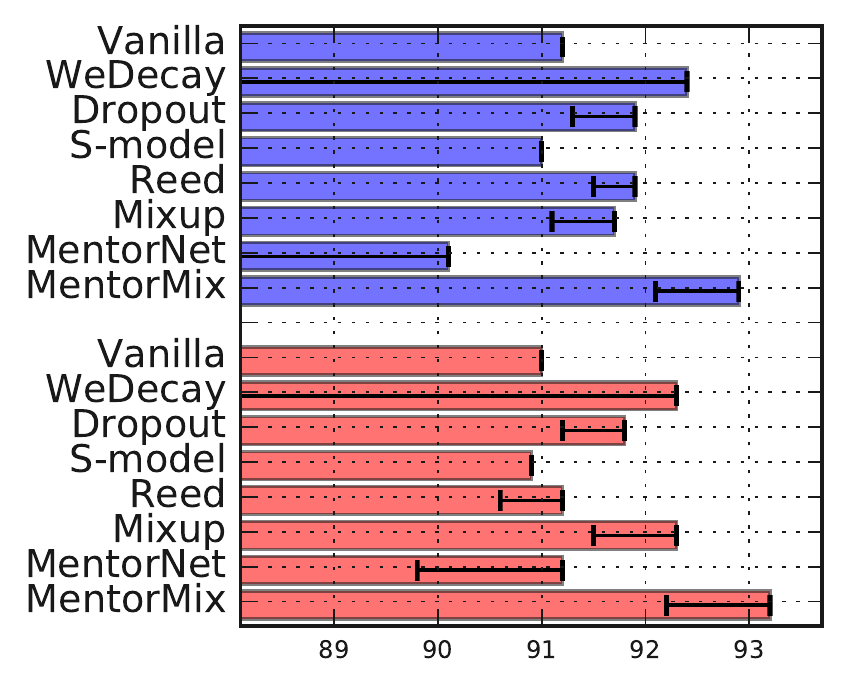}
    \caption{0\%}
\end{subfigure}
\begin{subfigure}[b]{0.19\textwidth}
    \includegraphics[width=\textwidth]{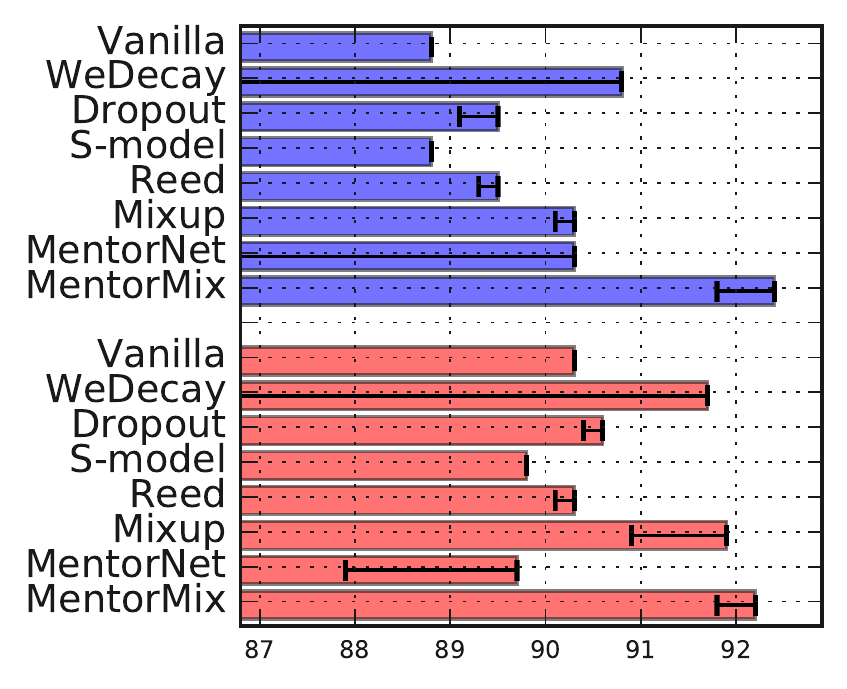}
    \caption{5\%}
\end{subfigure}
\begin{subfigure}[b]{0.19\textwidth}
    \includegraphics[width=\textwidth]{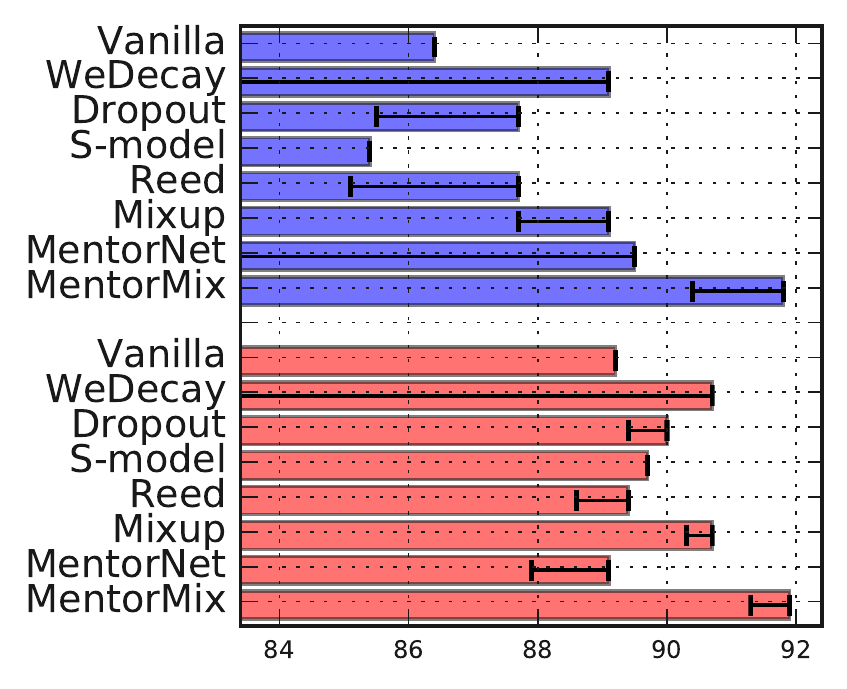}
    \caption{10\%}
\end{subfigure}
\begin{subfigure}[b]{0.19\textwidth}
    \includegraphics[width=\textwidth]{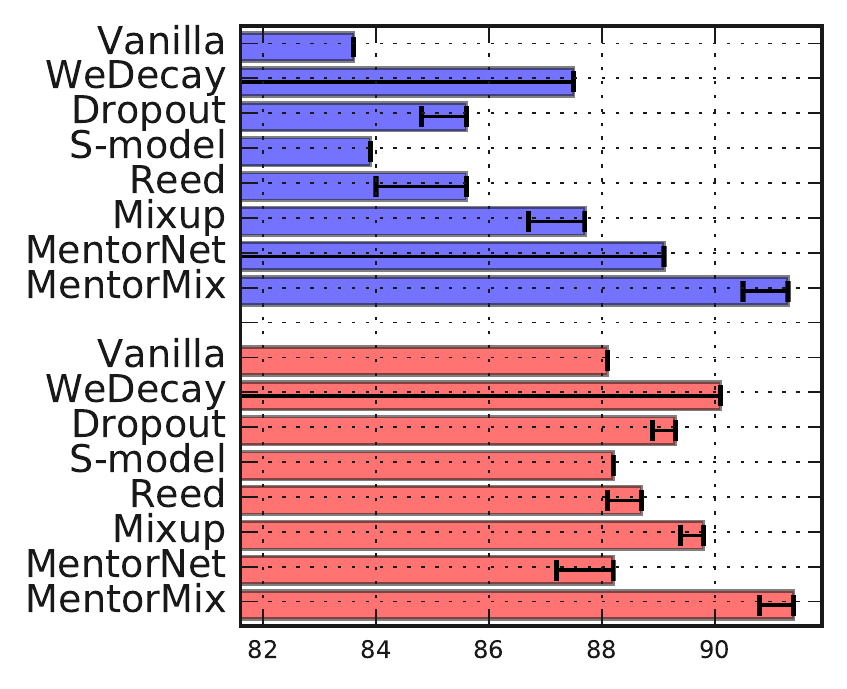}
    \caption{15\%}
\end{subfigure}
\begin{subfigure}[b]{0.19\textwidth}
    \includegraphics[width=\textwidth]{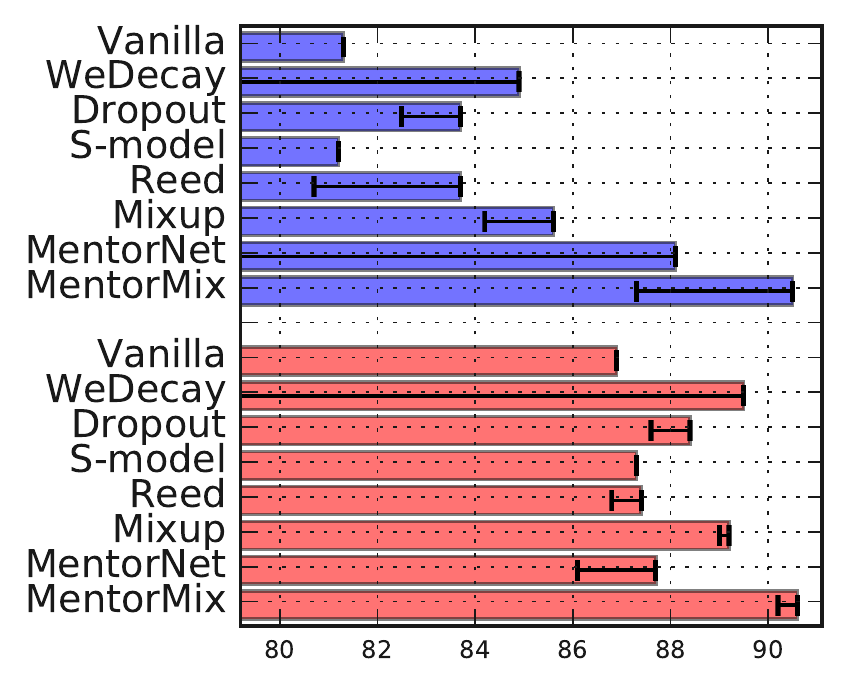}
    \caption{20\%}
\end{subfigure}
\begin{subfigure}[b]{0.19\textwidth}
    \includegraphics[width=\textwidth]{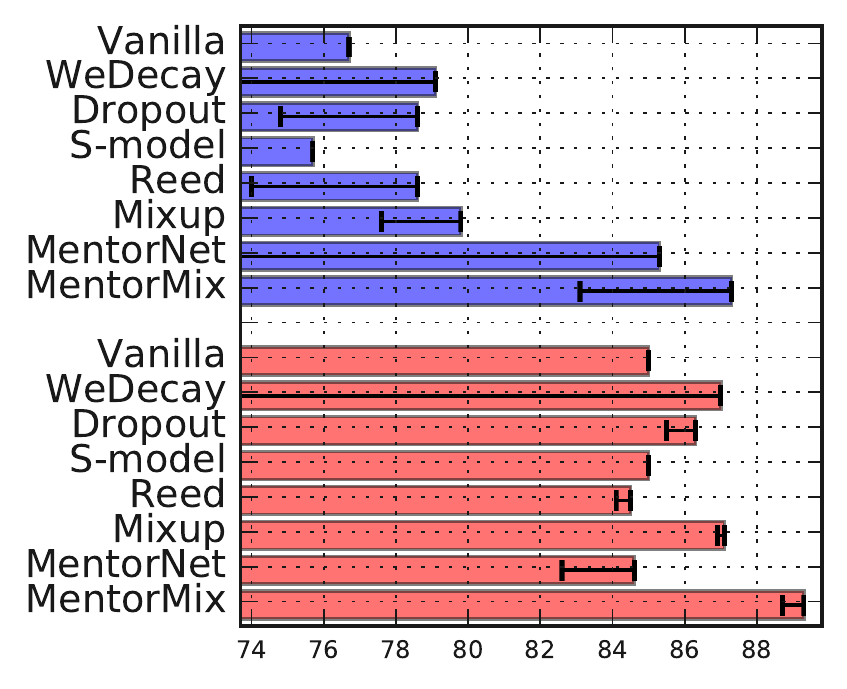}
    \caption{30\%}
\end{subfigure}
\begin{subfigure}[b]{0.19\textwidth}
    \includegraphics[width=\textwidth]{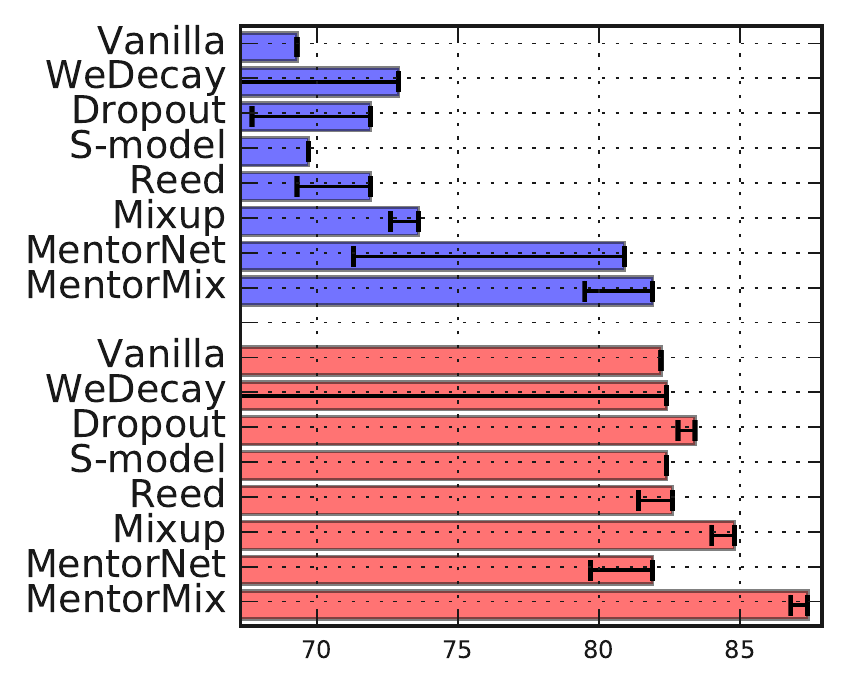}
    \caption{40\%}
\end{subfigure}
\begin{subfigure}[b]{0.19\textwidth}
    \includegraphics[width=\textwidth]{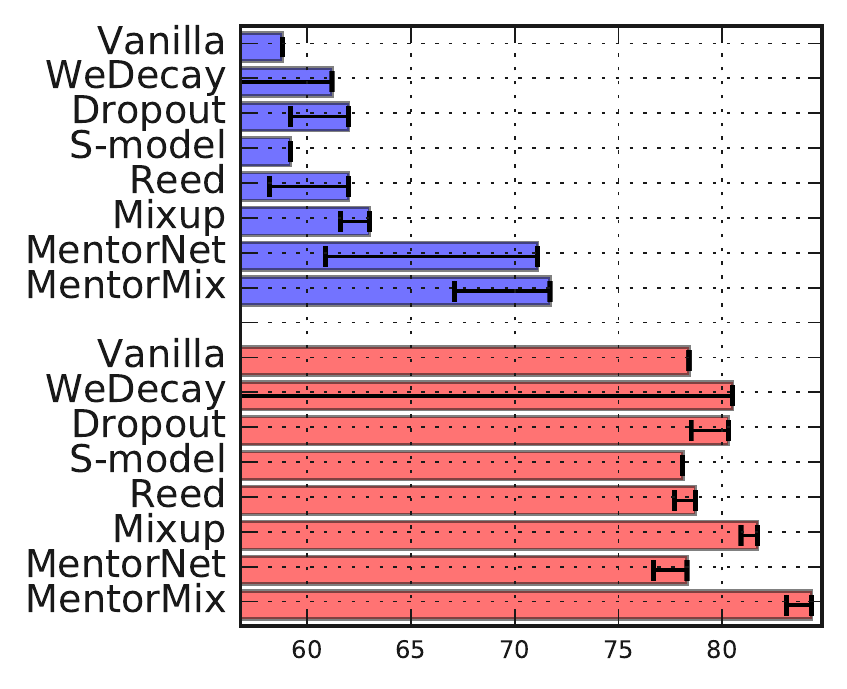}
    \caption{50\%}
\end{subfigure}
\begin{subfigure}[b]{0.19\textwidth}
    \includegraphics[width=\textwidth]{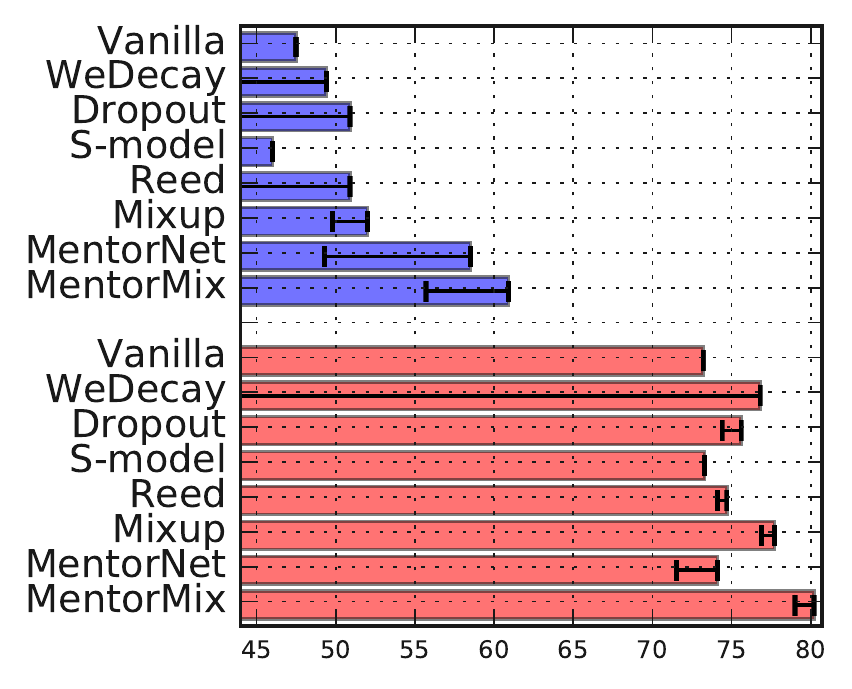}
    \caption{60\%}
\end{subfigure}
\begin{subfigure}[b]{0.19\textwidth}
    \includegraphics[width=\textwidth]{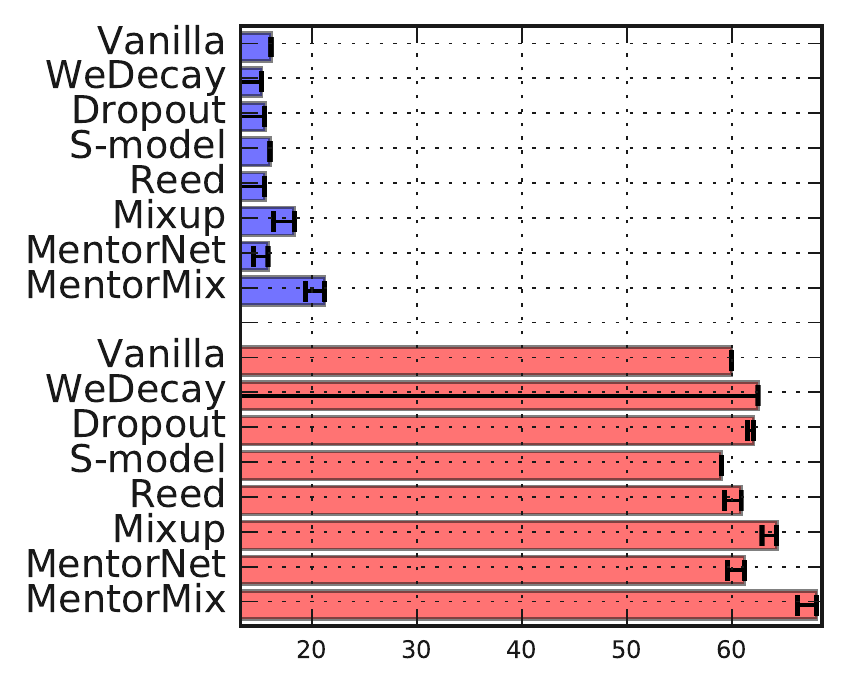}
    \caption{80\%}
\end{subfigure}
\vspace{-3mm}
\caption{\label{fig:comp_cars_ft}Comparison of the peak validation accuracy fine-tuned on Red and Blue Stanford Cars.}
\vspace{-3mm}
\end{figure}

This subsection presents detailed comparison results on our datasets. To be specific, we show the peak/final accuracy on the clean validation set in four tables: Table~\ref{tab:method_imagenet_ft}, Table~\ref{tab:method_imagenet_sc}, Table~\ref{tab:method_cars_ft}, and Table~\ref{tab:method_cars_sc}, where the best trial out of all searched hyperparameters is shown. The performance variance of all searched hyperparameters is shown in four figures: Fig.~\ref{fig:comp_imagenet_ft}, Fig.~\ref{fig:comp_imagenet_sc}, Fig.~\ref{fig:comp_cars_ft}, and Fig.~\ref{fig:comp_cars_sc}, where the black line shows the 95\% confidence interval. The results in all tables and figures show that the proposed method MentorMix consistently outperforms baseline methods and has comparable performance variance in the searched hyperparameter range.

\begin{table}[ht]
\vspace{-3mm}
\centering
\small
\caption{Best peak accuracy (\%) for baseline methods fine-tuned on Mini-ImageNet. The peak and final validation accuracies are shown in the format of XXX/YYY.}
\label{tab:method_imagenet_ft}
\begin{tabular}{@{}cccccccccc@{}}
\toprule
\multicolumn{1}{l}{\footnotesize \bf Type}& {\bf \footnotesize Noise Level}& {\bf Vanilla}& {\bf WeDecay}& {\bf Dropout} & {\bf S-Model} & {\bf Reed Soft} & {\bf Mixup} & {\bf MentorNet} & {\bf MentorMix} \\
\midrule
\multicolumn{1}{c}{\multirow{10}{*}{\rotatebox[origin=c]{90}{Blue Noise}}} 
&0&85.1/83.3&84.5/81.9&85.0/83.6&85.2/83.5&\bbf{85.6}/84.3&83.4/82.0&84.8/83.4&85.0/83.9\\
&5&84.5/82.0&84.2/79.7&84.5/80.8&84.1/81.5&84.7/81.9&83.5/82.0&\bbf{84.9}/84.5&84.8/84.0\\
&10&83.9/77.8&83.8/75.9&84.3/78.0&84.3/78.2&84.4/81.5&83.3/80.1&84.5/84.3&\bbf{85.2}/83.6\\
&15&83.6/75.8&83.1/73.8&83.6/74.7&83.7/76.1&\bbf{84.7}/80.4&84.0/79.5&84.5/81.2&84.6/83.2\\
&20&83.5/74.0&82.8/71.9&83.5/73.9&83.9/73.7&84.5/79.2&82.9/77.9&84.3/76.5&\bbf{84.5}/83.6\\
&30&82.9/66.6&82.8/64.1&82.4/67.5&82.4/66.0&83.0/76.7&83.3/74.6&83.2/74.0&\bbf{84.5}/82.8\\
&40&81.6/61.0&81.1/58.5&82.4/59.0&82.2/60.2&83.0/78.3&81.8/70.8&82.5/80.4&\bbf{84.3}/81.3\\
&50&80.9/50.5&80.5/47.9&82.0/68.1&80.0/50.8&81.6/53.1&79.9/64.6&81.8/71.0&\bbf{83.7}/79.0\\
&60&80.5/40.9&79.7/40.3&81.4/38.9&80.7/42.3&81.6/45.2&79.2/63.6&81.0/79.8&\bbf{83.4}/77.9\\
&80&76.1/19.7&76.5/17.6&78.9/24.7&76.9/20.2&77.8/22.2&76.1/60.0&77.5/32.3&\bbf{82.0}/73.3\\
\midrule
\multicolumn{1}{c}{\multirow{10}{*}{\rotatebox[origin=c]{90}{Red Noise}}} 
&0&84.7/82.6&83.6/81.9&84.3/83.0&\bbf{85.0}/83.3&84.8/82.5&83.0/81.8&84.6/83.5&84.4/83.4\\
&5&84.6/80.0&84.0/78.9&84.6/80.2&84.7/80.4&85.2/80.9&84.8/82.3&84.6/84.0&\bbf{85.3}/85.0\\
&10&83.9/78.1&83.9/78.1&83.9/78.1&84.5/78.4&84.3/78.6&84.1/80.8&84.3/84.1&\bbf{85.1}/85.0\\
&15&82.3/77.4&83.0/76.5&83.1/76.5&83.0/77.5&84.3/78.0&83.9/80.8&83.6/82.9&\bbf{85.5}/84.7\\
&20&82.9/76.1&82.9/75.6&82.7/75.5&82.5/76.9&84.3/77.4&83.2/79.4&83.5/83.4&\bbf{85.0}/84.6\\
&30&82.0/74.6&81.2/74.6&81.6/73.7&82.5/73.8&83.0/74.7&83.3/78.6&82.7/82.6&\bbf{83.9}/81.4\\
&40&80.7/74.0&81.2/71.7&81.2/72.6&81.4/73.9&82.3/73.1&82.4/77.6&81.9/80.2&\bbf{83.1}/81.5\\
&50&80.3/72.9&80.3/71.4&80.7/71.9&80.5/72.2&81.7/72.4&82.0/77.3&81.1/79.3&\bbf{82.2}/80.4\\
&60&78.6/70.3&79.2/68.8&80.1/69.4&78.8/70.1&80.7/69.5&80.4/75.5&80.5/73.4&\bbf{81.0}/77.1\\
&80&76.3/64.6&76.1/63.1&76.1/63.4&76.7/65.0&76.7/65.3&76.6/71.3&76.8/72.8&\bbf{77.2}/74.0\\
\bottomrule
\end{tabular}
\vspace{-3mm}
\end{table}

\begin{table}[ht]
\centering
\small
\vspace{-3mm}
\caption{Best peak accuracy (\%) for baseline methods trained from scratch on Mini-ImageNet. The peak and final validation accuracies are shown in the format of XXX/YYY. '-' denotes the method that has failed to converge. }
\label{tab:method_imagenet_sc}
\begin{tabular}{@{}cccccccccc@{}}
\toprule
\multicolumn{1}{l}{\footnotesize \bf Type}& {\bf \footnotesize Noise Level}& {\bf Vanilla}& {\bf WeDecay}& {\bf Dropout} & {\bf S-Model} & {\bf Reed Soft} & {\bf Mixup} & {\bf MentorNet} & {\bf MentorMix} \\
\midrule
\multicolumn{1}{c}{\multirow{10}{*}{\rotatebox[origin=c]{90}{Blue Noise}}} 
&0&73.1/72.8&-/-&73.1/67.9&73.8/71.8&73.8/71.2&73.7/73.0&73.2/71.5&\bbf{75.1}/74.4\\
&5&70.9/70.7&-/-&70.2/63.4&71.1/67.1&71.1/66.7&72.4/70.3&72.5/69.5&\bbf{75.0}/74.7\\
&10&69.0/63.9&-/-&68.5/60.5&68.1/63.6&69.2/63.4&70.2/67.2&70.2/67.9&\bbf{73.6}/73.3\\
&15&67.1/60.7&-/-&65.6/56.7&66.6/61.1&67.7/60.5&68.1/63.2&69.7/66.1&\bbf{73.6}/73.2\\
&20&63.0/58.0&-/-&65.1/53.4&63.5/57.7&65.2/57.6&66.5/60.6&67.4/65.8&\bbf{73.9}/73.3\\
&30&59.9/50.7&-/-&61.5/47.2&62.4/50.8&63.6/52.2&63.0/54.6&66.0/64.2&\bbf{72.3}/71.6\\
&40&56.9/43.5&-/-&58.1/40.2&57.3/45.0&60.2/44.5&60.3/46.9&62.5/62.1&\bbf{70.4}/69.4\\
&50&52.9/36.4&-/-&54.3/33.5&50.9/36.1&54.1/36.8&55.3/42.8&59.6/57.7&\bbf{69.2}/66.7\\
&60&44.6/26.0&-/-&48.2/23.3&46.7/27.1&47.2/27.2&48.9/47.4&52.0/47.2&\bbf{66.5}/62.9\\
&80&25.9/14.0&-/-&28.4/18.8&26.5/18.7&29.0/12.7&28.5/28.5&25.4/18.5&\bbf{59.9}/52.7\\
\midrule
\multicolumn{1}{c}{\multirow{10}{*}{\rotatebox[origin=c]{90}{Red Noise}}} 
&0&70.9/68.5&-/-&71.8/65.7&71.4/68.4&71.8/68.4&72.8/72.3&71.2/68.9&\bbf{74.3}/73.7\\
&5&70.9/66.6&-/-&71.8/62.8&71.2/67.0&71.9/66.7&71.8/69.4&71.5/67.4&\bbf{73.6}/73.4\\
&10&70.8/63.9&-/-&71.0/61.3&69.8/63.9&70.4/63.6&71.1/68.3&70.8/65.6&\bbf{73.0}/71.4\\
&15&69.8/62.2&-/-&69.3/60.0&69.0/60.5&69.3/62.2&69.9/65.9&69.7/66.3&\bbf{71.5}/70.8\\
&20&68.3/60.3&-/-&68.7/57.6&67.9/60.3&68.3/60.4&69.3/64.4&67.9/62.8&\bbf{70.1}/69.1\\
&30&66.1/56.5&-/-&66.6/55.0&65.2/56.3&66.6/56.7&66.8/61.8&66.2/64.0&\bbf{68.3}/67.2\\
&40&64.5/54.7&-/-&\bbf{66.1}/53.0&64.1/54.6&64.7/54.0&65.8/59.6&63.9/56.5&66.0/64.7\\
&50&60.9/51.7&-/-&62.1/50.1&61.3/51.3&62.6/52.5&63.2/58.4&61.7/58.0&\bbf{63.3}/61.8\\
&60&57.6/49.0&-/-&59.7/46.8&57.0/47.5&59.0/49.2&59.0/53.4&58.8/51.3&\bbf{60.0}/57.5\\
&80&48.8/39.8&-/-&49.5/37.6&49.0/40.6&50.1/40.1&\bbf{50.7}/45.5&49.3/43.4&50.2/48.4\\
\bottomrule
\end{tabular}
\vspace{-3mm}
\end{table}

\begin{table}[ht]
\centering
\vspace{-3mm}
\small
\caption{Best peak accuracy (\%) for baseline methods trained from scratch on Stanford Cars. The peak and final validation accuracies are shown in the format of XXX/YYY.}
\label{tab:method_cars_ft}
\begin{tabular}{@{}cccccccccc@{}}
\toprule
\multicolumn{1}{l}{\footnotesize \bf Type}& {\bf \footnotesize Noise Level}& {\bf Vanilla}& {\bf WeDecay}& {\bf Dropout} & {\bf S-Model} & {\bf Reed Soft} & {\bf Mixup} & {\bf MentorNet} & {\bf MentorMix} \\
\midrule
\multicolumn{1}{c}{\multirow{10}{*}{\rotatebox[origin=c]{90}{Blue Noise}}} 
&0&91.2/90.6&92.4/92.2&91.9/91.3&91.0/90.7&91.3/91.0&91.7/91.6&90.1/90.0&\bbf{92.9}/92.9\\
&5&88.8/87.7&90.8/90.5&89.5/88.4&88.8/88.3&88.8/88.8&90.3/90.0&90.3/89.8&\bbf{92.4}/92.2\\
&10&86.4/84.6&89.1/87.9&87.7/85.1&85.4/84.2&85.4/87.5&89.1/88.9&89.5/89.5&\bbf{91.8}/91.8\\
&15&83.6/81.7&87.5/86.4&85.6/81.7&83.9/81.4&83.9/86.4&87.7/87.2&89.1/89.1&\bbf{91.3}/91.2\\
&20&81.3/78.2&84.9/82.9&83.7/77.4&81.2/78.1&81.2/83.9&85.6/85.6&88.1/87.8&\bbf{90.5}/90.4\\
&30&76.7/68.2&79.1/75.3&78.6/67.0&75.7/68.5&75.7/79.8&79.8/76.4&85.3/85.2&\bbf{87.3}/86.3\\
&40&69.3/56.8&72.9/63.8&71.9/56.0&69.7/58.2&69.7/71.8&73.6/68.1&80.9/79.3&\bbf{81.9}/77.7\\
&50&58.8/44.1&61.2/48.7&62.0/44.2&59.2/45.1&59.2/60.0&63.0/56.2&71.1/66.7&\bbf{71.7}/65.1\\
&60&47.5/32.8&49.4/36.9&50.9/31.8&46.0/32.4&46.0/47.8&52.0/43.6&58.5/57.0&\bbf{60.9}/52.5\\
&80&16.1/10.8&15.2/10.0&15.5/10.1&16.0/10.6&16.0/15.9&18.3/17.2&15.8/13.8&\bbf{21.2}/19.2\\
\midrule
\multicolumn{1}{c}{\multirow{10}{*}{\rotatebox[origin=c]{90}{Red Noise}}} 
&0&91.0/90.7&92.3/92.1&91.8/91.2&90.9/90.7&91.2/90.8&92.3/92.3&91.2/91.1&\bbf{93.2}/93.2\\
&5&90.3/90.1&91.7/91.7&90.6/89.6&89.8/89.2&90.3/89.6&91.9/91.8&89.7/89.3&\bbf{92.2}/92.2\\
&10&89.2/88.5&90.7/90.5&90.0/89.1&89.7/89.1&89.4/88.9&90.7/90.5&89.1/88.7&\bbf{91.9}/91.9\\
&15&88.1/87.5&90.1/89.5&Mix/88.1&88.2/87.8&88.7/88.4&89.8/89.7&88.2/87.8&\bbf{91.4}/91.4\\
&20&86.9/86.2&89.5/89.0&88.4/87.0&87.3/86.8&87.4/86.1&89.2/89.0&87.7/86.7&\bbf{90.6}/90.5\\
&30&85.0/84.3&87.0/86.0&86.3/84.0&85.0/84.2&84.5/83.8&87.1/86.9&84.6/84.3&\bbf{89.3}/89.3\\
&40&82.2/81.4&82.4/81.3&83.4/81.7&82.4/80.9&82.6/81.6&84.8/84.3&81.9/81.0&\bbf{87.4}/87.4\\
&50&78.4/76.7&80.5/80.0&80.3/77.2&78.1/76.6&78.7/76.7&81.7/81.6&78.3/76.8&\bbf{84.3}/83.9\\
&60&73.2/71.4&76.8/75.0&75.6/72.1&73.3/71.6&74.7/72.1&77.7/77.4&74.1/72.9&\bbf{80.2}/80.1\\
&80&60.0/57.7&62.5/61.2&62.1/57.3&59.0/56.8&60.9/58.1&64.3/63.0&61.2/57.2&\bbf{68.1}/67.9\\
\bottomrule
\end{tabular}
\vspace{-3mm}
\end{table}

\begin{table}[ht]
\vspace{-3mm}
\centering
\small
\caption{Best peak accuracy (\%) for baseline methods trained from scratch on Stanford Cars. The peak and final validation accuracies are shown in the format of XXX/YYY. '-' denotes the method that has failed to converge.}
\label{tab:method_cars_sc}
\begin{tabular}{@{}cccccccccc@{}}
\toprule
\multicolumn{1}{l}{\footnotesize \bf Type}& {\bf \footnotesize Noise Level}& {\bf Vanilla}& {\bf WeDecay}& {\bf Dropout} & {\bf S-Model} & {\bf Reed Soft} & {\bf Mixup} & {\bf MentorNet} & {\bf MentorMix} \\
\midrule
\multicolumn{1}{c}{\multirow{10}{*}{\rotatebox[origin=c]{90}{Blue Noise}}}
&0&90.5/89.9&-/-&\bbf{92.4}/92.4&89.6/89.5&91.2/91.2&91.5/91.3&90.6/90.4&92.1/92.0\\
&5&86.7/86.3&-/-&89.7/89.6&87.7/87.6&86.4/86.2&90.4/90.4&87.6/87.6&\bbf{91.8}/91.6\\
&10&82.4/81.7&-/-&87.9/87.6&81.5/81.4&83.5/83.1&87.5/87.4&84.0/83.9&\bbf{90.4}/90.2\\
&15&78.3/77.9&-/-&84.8/84.5&75.4/75.2&79.4/79.1&84.1/84.0&79.5/79.3&\bbf{88.9}/88.9\\
&20&69.9/69.0&-/-&82.3/82.2&73.0/72.8&73.0/72.7&81.7/81.6&75.4/75.2&\bbf{88.1}/87.9\\
&30&62.7/62.6&-/-&75.8/71.1&56.9/56.4&65.2/64.7&71.2/71.1&58.9/57.7&\bbf{80.9}/80.7\\
&40&40.4/37.2&-/-&59.0/58.7&37.0/37.0&41.1/40.7&60.1/60.0&47.6/47.5&\bbf{66.2}/66.2\\
&50&14.6/14.0&-/-&34.7/33.2&26.9/26.6&21.9/21.9&44.2/42.9&27.5/27.5&\bbf{54.2}/53.5\\
&60&09.1/06.8&-/-&18.5/18.0&8.5/8.5&11.8/11.3&\bbf{27.5}/25.4&14.5/14.2&21.2/17.2\\
&80&03.1/01.3&-/-&02.4/02.3&02.8/02.8&02.8/02.7&\bbf{03.3}/03.0&02.8/02.7&02.9/02.8\\
\midrule
\multicolumn{1}{c}{\multirow{10}{*}{\rotatebox[origin=c]{90}{Red Noise}}} 
&0&90.8/90.8&-/-&\bbf{92.2}/92.2&90.1/90.1&90.3/90.0&91.9/91.9&90.2/90.1&91.8/91.6\\
&5&89.2/89.2&-/-&91.2/90.8&89.0/88.9&88.9/88.8&90.3/90.2&88.8/88.6&\bbf{91.4}/91.3\\
&10&88.3/88.3&-/-&90.2/90.2&87.8/87.8&87.9/87.7&89.9/89.9&88.3/88.3&\bbf{91.2}/90.9\\
&15&86.3/86.3&-/-&89.6/89.6&87.0/86.9&87.2/87.2&89.4/89.1&86.1/59.9&\bbf{89.7}/89.5\\
&20&84.9/84.7&-/-&\bbf{88.9}/88.9&83.7/83.6&85.8/85.7&87.8/87.6&85.0/84.8&88.7/88.6\\
&30&80.4/80.2&-/-&87.6/87.6&82.2/81.9&83.4/83.0&85.6/85.2&81.1/80.9&\bbf{87.8}/87.7\\
&40&77.4/76.9&-/-&\bbf{84.0}/84.0&78.0/77.8&78.2/77.8&82.8/82.5&80.2/76.9&81.0/80.4\\
&50&70.6/70.3&-/-&79.3/79.2&70.1/70.1&73.6/73.5&79.1/78.9&72.0/72.0&\bbf{80.4}/79.8\\
&60&66.2/66.2&-/-&\bbf{76.3}/75.9&61.8/61.4&66.8/66.6&72.5/72.1&66.7/66.6&75.0/74.9\\
&80&43.3/43.0&-/-&\bbf{61.8}/61.8&46.4/46.4&47.4/46.7&55.7/55.4&51.0/50.9&58.6/58.6\\
\bottomrule
\end{tabular}
\vspace{-3mm}
\end{table}

\clearpage

\eat{
\section{Results on more network architectures}
\begin{table}[]
\centering
\caption{Comparison of the standard deviation of the final accuracies across 10 noise levels and network architectures. A higher standard deviation suggests a poorer generalization performance of DNNs trained on noisy labels. To better show the difference, we compute ``relative diff $\Delta$''as the ratio between the blue and red standard deviation. Our finding in the main paper, \ie~DNNs generalize much better on the red label noise, seems to be consistent on different network architectures.}
\label{tab:architecture_finding1}
\begin{tabular}{l|l|cc|cc}
\toprule
{\multirow{2}{*}{Net Architecture}} & {\multirow{2}{*}{Method}} & \multicolumn{2}{c}{Mini-ImageNet} & \multicolumn{2}{c}{Stanford Cars} \\
{} & {} & Fine-tuned & Trained from scratch &  Fine-tuned  & Trained from scratch \\
\midrule
{\multirow{3}{*}{Inception-ResNet}} & {Blue noise} & 0.205 & 0.195 & 0.268 & 0.347 \\
{} & {Red noise} & 0.051 & 0.088 & 0.104 & 0.146 \\
{} & {Relative diff $\Delta$} & \hl{402\%} & \hl{222\%} & \hl{258\%} & \hl{238\%}\\
\hline
\multirow{3}{*}{EfficientNet} & {Blue noise} &0.200 &0.183 &0.251 & 0.322\\
 & {Red noise} &0.064 &0.070 &0.079 & 0.174\\
 {} & {Relative diff $\Delta$} & \hl{313\%}  & \hl{261\%}  & \hl{318\%} & \hl{185\%} \\
\hline
\multirow{3}{*}{MobileNet} & {Blue noise} & 0.184& 0.189&0.274 &0.306 \\
 & {Red noise} & 0.070&0.088 & 0.113&0.172 \\
 {} & {Relative diff $\Delta$} & \hl{263\%} & \hl{215\%} & \hl{242\%} & \hl{178\%} \\
\hline
\multirow{3}{*}{ResNet-50} & {Blue noise} &0.191 & 0.190& 0.271&0.299 \\
 & {Red noise} &0.067 &0.088 &0.119 &0.156 \\
 {} & {relative diff $\Delta$} & \hl{285\%} & \hl{216\%} & \hl{228\%} & \hl{192\%}\\
\hline
\multirow{3}{*}{ResNet-101} & {Blue noise} & 0.193& 0.192& 0.279&0.285 \\
 & {Red noise} &0.067 &0.090 & 0.120&0.162 \\
 {} & {Relative diff $\Delta$} & \hl{288\%} & \hl{213\%} & \hl{233\%} & \hl{176\%} \\
\bottomrule
\end{tabular}
\end{table}
}

\bibliography{cr_paper}
\bibliographystyle{icml2020}